%% file: main.tex
\RequirePackage{xparse}
\documentclass{article} %
\usepackage{preprint,times}

\input{math_commands.tex}

\usepackage[bookmarks=true,colorlinks=true,citecolor=blue,linkcolor=blue,urlcolor=blue]{hyperref}
\usepackage{url}
\usepackage{graphicx}
\usepackage[labelformat=simple]{subcaption}

\usepackage{xspace}
\usepackage{bm}

\usepackage{microtype}
\usepackage{flushend}
\usepackage{algorithm}
\usepackage{algorithmic}
\usepackage{booktabs}
\usepackage{multirow}

\usepackage{amsmath}
\usepackage{amssymb}
\usepackage{mathtools}
\usepackage{amsthm}
\usepackage{pdflscape}
\usepackage{pifont}

\makeatletter
\renewcommand{\paragraph}{%
  \@startsection{paragraph}{4}%
  {\z@}{0.1ex \@plus .1ex}{-1em}%
  {\normalfont\normalsize\bfseries}%
}
\makeatother

\newif\ifshowcomments
\ifshowcomments
    \usepackage[textsize=scriptsize,textwidth=3.25cm]{todonotes}
\else
    \usepackage[disable,textsize=scriptsize,textwidth=3.2cm]{todonotes}
\fi

\newcommand{\todomw}[1]{}
\newcommand{\todoinmw}[1]{}

\newcommand{\todokc}[1]{}

\newcommand{\xmark}{\ding{55}}

\newcommand{\alg}{CLAW\xspace}

\title{CLAW: 
Continuous Latent Action World \\Models 
via Adversarial Regularization}

\author{%
\makebox[\dimexpr\textwidth-2\tabcolsep\relax][c]{%
\begin{tabular}[t]{c}
\bfseries
\mbox{Tewodros W.~Ayalew\textsuperscript{1}},
\mbox{Samuel Wheeler\textsuperscript{3}},
\mbox{Matthew Jeung\textsuperscript{1}},
\mbox{Xiao Zhang\textsuperscript{1}}, \\
\bfseries
\mbox{Andre de la Cruz Arce\textsuperscript{1}},
\mbox{Kaylene Stocking\textsuperscript{2}},
\mbox{Michael Maire\textsuperscript{1}},
\mbox{Matthew R.~Walter\textsuperscript{2}}
\\[8pt]
\mdseries
\mbox{\textsuperscript{1}University of Chicago} \quad
\mbox{\textsuperscript{2}Toyota Technological Institute at Chicago} \\
\mdseries
\mbox{\textsuperscript{3}Argonne National Laboratory}
\\[2pt]
\mdseries Project Page: \url{https://ripl.github.io/claw/}
\end{tabular}}}

\begin{document}

\maketitle
\input{sections/00_abstract}
\input{sections/01_intro}

\input{sections/02_related_work}

\input{sections/03_method}

\input{sections/04_experiments}

\input{sections/06_discussion}

\input{sections/05_conclusion}

\bibliography{references}
\bibliographystyle{bibliography}

\clearpage
\appendix
\section*{Appendix}
\input{supp}

\input{extended_sup}

\end{document}

%% file: math_commands.tex
\usepackage{amsmath,amsfonts,bm}

\def\eqref#1{equation~\ref{#1}}

\def\1{\bm{1}}

\DeclareMathAlphabet{\mathsfit}{\encodingdefault}{\sfdefault}{m}{sl}
\SetMathAlphabet{\mathsfit}{bold}{\encodingdefault}{\sfdefault}{bx}{n}

%% file: sections/00_abstract.tex
\begin{abstract}
We introduce \alg, a fully end-to-end self-supervised framework for learning a world model jointly with continuous latent action representations directly from action-free videos. 
Our approach leverages adversarial latent regularization to prevent the latent from encoding future observations, so that it captures the structure of the action. The diffusion-based world model then models environment dynamics as a distribution over future observations rather than as a single deterministic rollout. Both components are learned from observation sequences alone without any action annotations.
By simultaneously training the Latent Action Model and world model, \alg learns to reason about how inferred actions induce environment transitions from visual observations alone. We show that the resulting latent action world model supports both imitation learning from observation and goal-directed planning. In imitation learning, latent actions extracted from raw videos enable behavior cloning. For planning, \alg generates sequences of latent actions and maps them to executable actions to reach desired goals. 
Extensive experiments across diverse tasks and embodiments demonstrate that \alg produces semantically meaningful latent action representations, supports effective action transfer, and enables planning and imitation from observation, outperforming existing methods.
\end{abstract}

%% file: sections/01_intro.tex
\section{Introduction}

Models that predict how actions affect the state of an environment have long been fundamental to planning~\citep{bertsekas2012dynamic}, control~\citep{tassa2012synthesis}, and learning~\citep{sutton1990integrated, sutton1991dyna, atkeson1997comparison, Deisenroth2011PILCOAM}. By modeling the consequences of actions, agents can evaluate candidate behaviors, optimize policies through imagined experience, and adapt to new tasks with improved data efficiency. Inspired by human cognition, renewed attention has focused on ``world models''~\citep{conant1970every, richalet1978model, lozano1983robot, bryson2018applied, ha2018worldmodel} as forward-predictive models that capture how an environment evolves in response to the behavior of an interacting agent~\citep{clark2013predictivebrains, bubic2010cognitionprediction, hafner2019learninglatentdynamicsplanning, hafner2020dreamcontrollearningbehaviors, schrittwieser2020mastering, hafner2024masteringdiversedomainsworld}. Contemporary world models often condition their predictions on explicit action labels~\citep{hafner2019learninglatentdynamicsplanning, hu2023gaia1generativeworldmodel, bar2025navigationworldmodels}. However, action representations are often specific to a particular agent or action space, limiting transfer across agents with different embodiments or action spaces~\citep{gao2025adaworld, garrido2026learninglatentactionworld, huang2026crossembodiment}. %
Moreover, requiring action labels prevents  learning at scale from the vast amounts of unlabeled video data available online~\citep{ye2025latentactionpretrainingvideos}.

A world model that instead reasons over learned latent action representations is not tied to a particular agent's action space, allowing it to be trained and applied across tasks and embodiments without explicit action labels~\citep{bu2025univlalearningacttaskcentric}. Recent methods train such models from visual observations alone~\citep{bruce2024genie, schmidt2024learningactactions, chen2024igorimagegoalrepresentationsatomic, gao2025adaworld,ye2025latentactionpretrainingvideos}. However, it remains difficult to learn latent actions that are semantically meaningful, i.e., structured so that transitions with similar effects map to nearby latents across contexts and embodiments. This structure is what ultimately allows the latent actions to be grounded to a robot's executable actions from only a small amount of supervision~\citep{zhang2026latent}. Another challenge is the tendency for the latent to either \emph{leak}~\citep{lee2026latent} content from the future observation, letting the world model take a shortcut that bypasses learning action semantics, or \emph{collapse}~\citep{zhang2026latent, wang2025co} to an uninformative representation~\citep{bruce2024genie, nikulin2025latentactionlearningrequires}.

To address these challenges, we introduce \alg, a fully self-supervised framework that jointly learns continuous latent action representations and a predictive world model directly from raw, action-free video. Our approach couples a Latent Action Model (LAM) with the world model through reciprocal supervision: the LAM infers compact representations of observed transitions that condition future prediction, while the world model provides the learning signal that shapes these representations. We further introduce adversarial latent regularization to prevent the latent actions from encoding future observations directly, encouraging them instead to capture the underlying structure of actions. Unlike prior continuous latent action methods, which use future prediction primarily to shape the latent representation and discard or freeze the predictive model afterward, \alg retains the world model as a component of the learned system.

Once trained, the learned latent action world model supports two complementary forms of agent learning and decision making. First, the learned latent action representations provide supervisory signals for policy learning from action-free videos, enabling both imitation learning from observation (ILfO)~\citep{torabi2019recent} and latent action pretraining for downstream policy adaptation. Second, the learned world model enables goal-directed planning by predicting the outcomes of candidate latent action sequences, allowing an agent to optimize its behavior toward specified visual goals~\citep{mendonca2023structured, ebert2018visual}.

We evaluate our method across diverse environments, tasks, and agent embodiments, spanning simulated and real-world manipulation, open-world gameplay, and procedurally generated environments. We assess the learned latent action representations through ILfO, latent action policy pretraining, nearest-neighbor action retrieval, and action transfer~\citep{gao2025adaworld, bruce2024genie}. We further evaluate the learned dynamics through goal-directed visual planning and controllability under latent action conditioning. These evaluations show that \alg learns structured and transferable latent actions that support effective policy learning and planning, outperforming existing approaches across a broad range of downstream tasks.

%% file: sections/02_related_work.tex
\section{Related Work}

\paragraph{Model-based Learning and World Models} 

World models encompass a broad class of models that predict the consequences of an agent's actions~\citep{ha2018worldmodel}. Such models are used extensively in model-based reinforcement learning ~\citep{williams2017informationlearning, nagabandi2017neuralnetworkdynamicsmodelbased, hafner2020dreamcontrollearningbehaviors, hafner2022masteringataridiscreteworld, hafner2024masteringdiversedomainsworld, janner2021trustmodelmodelbasedpolicy} and planning~\citep{hafner2019learninglatentdynamicsplanning,sekar2020planningexploreselfsupervisedworld,micheli2023transformerssampleefficientworldmodels, schrittwieser2020mastering}.
World models learn dynamics from image-based inputs, which capture fine details and don't require hand-designed state representations~\citep{alonso2024diffusionworldmodelingvisual, bai2025wholebodyconditionedegocentricvideo, assran2025vjepa2selfsupervisedvideo, bar2025navigationworldmodels, wu2024ivideogptinteractivevideogptsscalable, zhou2025dinowmworldmodelspretrained}. Typically, these models are trained with ground-truth actions, conditioning future predictions on known control inputs to model environment dynamics and enable downstream planning or policy learning~\citep{bar2025navigationworldmodels, zhou2025dinowmworldmodelspretrained, alonso2024diffusionworldmodelingvisual, wang2026interactive}.

\paragraph{Continuous Latent Action Learning}

Recent work has explored continuous latent action representations as an alternative to discrete latent action spaces. CLAM~\citep{liang2025clam} trains a latent inverse dynamics model by deterministic next-observation reconstruction, but relies on an action decoder trained on action-labeled play data to keep the latent decodable. UniSkill~\citep{kim2025uniskill} learns skill representations from future-frame prediction without action labels, but imposes an information bottleneck and builds on pretrained components, including InstructPix2Pix initialization and a frozen DepthAnythingV2 depth estimator. CoMo~\citep{yang2025comolearningcontinuouslatent} combines three mechanisms: early temporal feature differencing, a temporal contrastive objective focused on foreground motion, and a constrained latent dimensionality. 
Each of these methods restricts what the latent can represent or supervises it with action labels, and trains the representation separately from the policy that consumes it. \alg instead removes future-observation information adversarially while retaining a continuous latent space without an explicit information bottleneck, and trains the inverse and forward models simultaneously.

\paragraph{Latent Action World Models} 

Latent action world models infer actions from observation sequences alone, enabling training with unlabeled video and promising transferability across embodiments and contexts~\citep{gao2025adaworld, bruce2024genie, nikulin2025latentactionlearningrequires, ye2025latentactionpretrainingvideos}. Latent action world models typically use %
an inverse dynamics model that encodes a latent action representation based on observation sequences, and a forward dynamics model (FDM) that predicts the future state conditioned on the latent action and past observations~\citep{schmidt2024learningactactions, bu2025univlalearningacttaskcentric, cui2024dynamoindomaindynamicspretraining}. Often, latent action world models involve pretraining an action-agnostic FDM and only then introduce inferred actions in a second fine-tuning stage~\citep{wu2024ivideogptinteractivevideogptsscalable, mendonca2023structured, nvidia2025cosmosworldfoundationmodel, che2024gamegenxinteractiveopenworldgame, he2025pretrainedvideogenerativemodels}. Instead, we build upon research on `action-aware pretraining' that learns the world model and action representation simultaneously, which has been shown to improve downstream controllability and transfer~\citep{bruce2024genie}. 

A core challenge in training latent action world models is that the action representation will often embed information about future observations instead of encoding information about the action alone, especially with image-based observations~\citep{nikulin2025latentactionlearningrequires}.
Prior approaches typically regulate information leakage through explicit bottlenecks, most commonly via quantization \citep{bruce2024genie, ye2025latentactionpretrainingvideos, bu2025univlalearningacttaskcentric}, variational constraints~\citep{gao2025adaworld}, or capacity-limited continuous latents~\citep{yang2025comolearningcontinuouslatent}. 
However, recent work has shown that continuous actions facilitate finer-grained control and often stronger performance than discretization~\citep{intelligence2025pi05visionlanguageactionmodelopenworld, yang2025comolearningcontinuouslatent, kim2025finetuningvisionlanguageactionmodelsoptimizing}. Alternatively, other works introduce supervision through a small amount of action-labeled data~\citep{nikulin2025latentactionlearningrequires, yang2025comolearningcontinuouslatent}, which reduces leakage but requires some action annotations. In contrast, our approach applies adversarial latent regularization to discourage encoding of future visual information. By jointly training the LAM and world model, we learn representations that capture both action structure and environment dynamics without relying on quantization or action supervision.

%% file: sections/03_method.tex
\section{Latent Action World Model} %
\begin{figure*}
    \centering
    \includegraphics[width=0.9\linewidth]{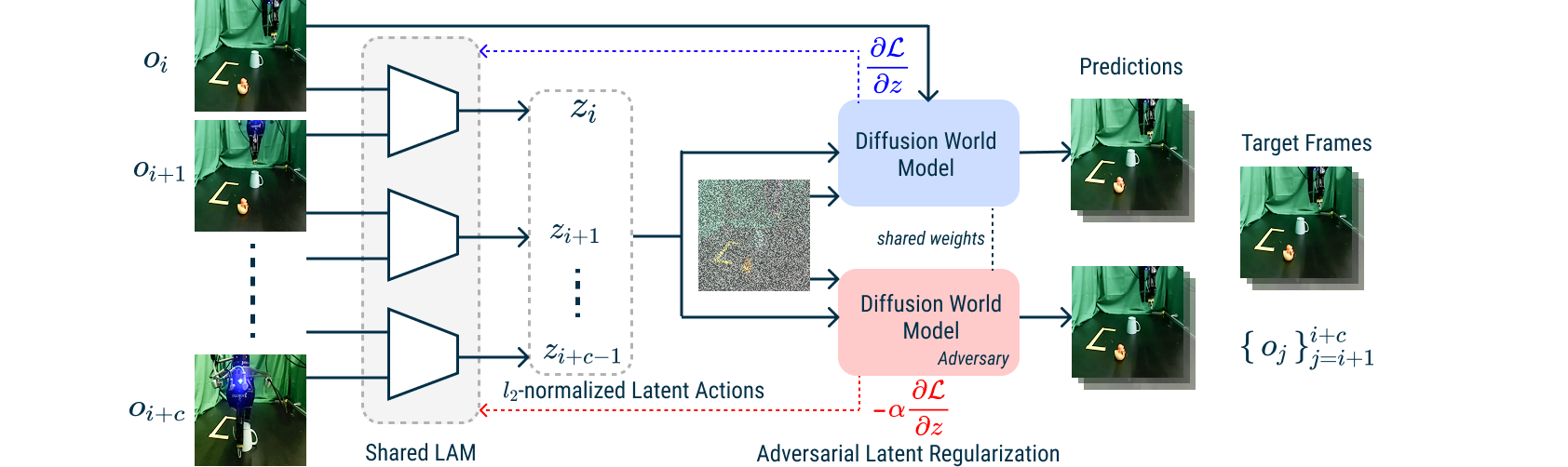}
    \caption{Overview of the \alg framework. The LAM encodes compact, normalized action representations (left). The adversarial latent regularization employs two diffusion world models with shared weights, with only the first receiving the initial frame alongside the action (right).}
    \label{fig:claw}
\end{figure*}

An action-conditioned world model (from now on, ``world model'') approximates observation transition dynamics $P(o_{i+1} \mid o_i, a_i)$, predicting future observations $o$ from past observations and actions $a$. Unlike standard world model formulations, latent action world models do not assume access to action annotations or the ability to sample from $\mathcal{A}$ during training. Instead, we introduce a learned latent variable $z_i$
that captures the effect of the underlying action on the observation dynamics. We then model these dynamics as $P(o_{i+1} \mid o_i, z_i)$ and learn the world model directly from observation sequences. The training data is an action-free video dataset $\mathcal{V} = \{ o_{1:T} \}^N$, where each video \mbox{$o_{1:T} = \{ o_1, \ldots, o_T \}$} is a sequence of RGB frames $o_i \in \mathbb{R}^{W \times H \times 3}$. Videos vary in length ($T$) and comprise diverse, unlabeled trajectories.
These videos may come, for example, from gameplay recordings or videos of robotic play~\citep{lynch2020learning}.

Figure~\ref{fig:claw} provides an overview of our proposed architecture and training setup. Similar to conventional inverse dynamics models (IDMs)~\citep{edwards2019imitating, schmidt2024learningactactions}, the Latent Action Model (LAM) is designed to infer the action responsible for the transition between consecutive observations. 
We use a ViT backbone~\citep{dosovitskiy2020image} to map consecutive observations, ${o_i, o_{i+1}}$, into a continuous latent action representation $z_i \in \mathbb{R}^n$ (Fig.~\ref{fig:action_encoder}): $z_i = \textrm{LAM}_{\theta}(o_{i}, o_{i+1})$. The predicted latent actions are constrained to lie on the unit sphere through $\ell_2$ normalization.

\alg subsequently projects the LAM output corresponding to the action token through an MLP to yield the continuous latent action representation, which conditions the world model and captures the action responsible for the observed transition. 
The LAM is trained end-to-end with the forward dynamics world model, described next.

\subsection{Diffusion-based World Model}
We represent the dynamics as a diffusion-based world model, implemented with a UViT backbone~\citep{bao2023all}. Given an initial observation frame $o_i$ and the sequence of latent actions $\{z_j\}_{j=i}^{i+c}$ inferred by the LAM, it predicts the corresponding future frames $\{o_j\}_{j=i+1}^{i+c+1}$, where $c$ defines the horizon. By conditioning on both the initial frame and inferred latent actions, the world model learns to use the inverse dynamics learned by the LAM to produce more accurate predictions.

The forward diffusion process follows a variance-preserving stochastic differential equation that progressively adds Gaussian noise to the target frames. The diffusion backbone operates on image patches within a transformer architecture. We structure the input  as $\left[ t_\tau \; \{z_j\}_{j=i}^c \; o_{i} \; \{x_j\}_{j=i}^c \right]$, 
where $t_\tau$ is the diffusion timestep embedding, \mbox{$\{z_j\}_{j=i}^c \in \mathbb{R}^{c \times d}$} are latent action tokens for $c$ transitions from the LAM, $o_i$ is the set of conditioning frame patches, and $x_j$ are patches from the noised observations $o_j$ with $i < j \le i+c$ generated by the forward diffusion process $q(o^{j}_t \mid o^{j}_0)$. Latent action tokens are projected to match the transformer embedding dimension. To maintain causal consistency and prevent information leakage, we use a modified causal attention mask that blocks noisy target tokens from attending to future action tokens or observations beyond their timestep (depicted in Figure~\ref{fig:worldmodel}).

Training optimizes a score network $s_\phi(x^t,t,z^t,o_{i})$ to predict the noise $\epsilon$ added during the forward diffusion process.
\begin{equation}
    \label{eqn:diffusion_loss}
    \mathcal{L} =
    \mathbb{E}_{t \sim \mathcal{U}(0,1),\, \epsilon \sim \mathcal{N}(0,\mathbf{I})}
    \left[
    \left\|
    \epsilon - s_\phi(x^t,t,z^t,o_{i})
    \right\|^2
    \right],
\end{equation}
where the noisy samples are $x^t = \sqrt{\alpha(t)}\,x^0 + \sqrt{1-\alpha(t)}\,\epsilon$.
At each step, the model conditions on the latent action representation $z^t$ and the reference frame $o_{i}$ to iteratively denoise the sample.

\subsection{Adversarial Information Regularization}
\label{section:adversarial_bottleneck}

The LAM and world model are trained together end-to-end to predict future frames. Unfortunately, this can lead to two kinds of degenerate solutions. The latent can collapse onto $z_i=o_{i+1}$, in which case the world model predicts accurately but learns no action semantics; or it can become uninformative, in which case the world model ignores it and predicts from $o_i$ alone. Adversarial latent regularization targets the first, and its gradual schedule avoids inducing the second.

In our approach, the LAM is coupled to the world model through two prediction pathways that share the same latent action representation $z_i$, as shown in Figure~\ref{fig:claw}. The first conditions the world model on both $o_i$ and $z_i$, encouraging $z_i$ to encode prediction-relevant information. The second provides only $z_i$. Since accurate predictions of future observations in this second branch can only occur if information about $o_{i+1}$ leaks into $z_i$, we \emph{penalize} accurate predictions by inserting a gradient reversal layer (GRL)~\citep{ganin2016domain} between $z_i$ and the world model. The GRL scales and reverses the gradient during backpropagation as $\frac{\partial \mathcal{L}}{\partial z} \leftarrow -\delta \frac{\partial \mathcal{L}}{\partial z}$; this \emph{adversarial latent regularization} encourages the LAM to avoid encoding information about future observations into the latent action $z_i$.

During training, the strength of the adversarial signal is controlled by a gradually increasing weight $\delta$, following a sigmoidal schedule
\begin{equation}
    \delta = \min\left(\frac{2}{1 + \exp(-\gamma p)} - 1, 0.999\right) \quad \textrm{where} \quad p = 
    \frac{\text{step}}{n_{\text{steps}}}
\end{equation}
where $\gamma = 10$ for all experiments as in~\citet{ganin2016domain}, and $n_{\text{steps}}$ is the total number of training steps. The schedule lets the model first learn to predict accurately, then progressively increases the pressure on the LAM to remove future-observation information, with $\delta$ capped at $0.999$. Starting from $\delta \approx 0$ also guards against the second failure mode, since the world model learns to depend on the latent before any adversarial pressure is applied.

%% file: sections/04_experiments.tex
\section{Experiments}

We evaluate \alg{} in terms of (1) the quality and structure of the learned latent action space, and (2) goal-directed planning using the latent action world model. Following prior work~\citep{gao2025adaworld, bruce2024genie}, we train all latent action world models solely on task-agnostic interaction data, collected by an agent exploring the environment without explicit objectives (Table~\ref{tab:evaluationoverview}).

\subsection{Baselines}

We compare \alg against AdaWorld~\citep{gao2025adaworld}, LAPO~\citep{schmidt2024learningactactions}, and CoMo~\citep{yang2025comolearningcontinuouslatent}, all of which learn latent actions from visual observations without ground-truth action labels. These baselines span continuous variational latent actions (AdaWorld), discrete vector-quantized latent actions (LAPO), and continuous contrastively learned latent motion (CoMo). We exclude CLAM~\citep{liang2025clam}, whose latent representation is shaped by action-labeled play data, and UniSkill~\citep{kim2025uniskill}, which relies on pretrained perceptual components. Genie~\citep{bruce2024genie} is not publicly available, so LAPO serves as the discrete vector-quantized method. LAPA~\citep{ye2025latentactionpretrainingvideos} is a larger-scale variant of a similar latent action learning framework that also uses vector quantization to learn discrete latent actions and supports downstream policy finetuning with action-labeled data. In addition to these baselines, we report \alg{} w/o Adv, an ablation of our method trained with the adversarial weight set to $\delta = 0$, so that the second prediction pathway continues to train the world model but provides no gradient signal to the LAM.

\subsection{Benchmarks and Environments}
\label{sec:benchmarks}
We assess \alg across 14 tasks in 5 simulated environments and 2 real-world robotic tasks.
\textbf{VP$^2$}~\citep{tian2023vp2} is a  benchmark for visual planning in tabletop manipulation based on the Robosuite~\citep{zhu2020robosuite} and RoboDesk~\citep{kannan2021robodesk} environments.
\textbf{OGBench}~\citep{park2025ogbench} is a benchmark in which a UR5e arm manipulates articulated objects; we evaluate \texttt{Open Drawer}, \texttt{Close Drawer}, and \texttt{Push Button}.
\textbf{Crafter}~\citep{hafner2021crafter} is a 2D open-world survival environment with procedurally generated worlds, where we evaluate a long-horizon task (\texttt{Greedy Wood Collection}) and a shorter goal-conditioned one (\texttt{Single Wood Collection}).
\textbf{Procgen}~\citep{cobbe2019procgen} is a suite of procedurally generated environments; we evaluate visual planning on Climber, Jumper, and Ninja using MPC-CEM~\citep{deboer2005cem}. Environment configurations, data generation, and task definitions are in Appendix~\ref{section:benchmark_appendix}.

\textbf{Real-world tabletop manipulation.} We evaluate latent action pretraining on a 6-DOF UR5 robot, training the latent action world model on Something-Something V2~\citep{goyal2017something}, a dataset of humans interacting with household items, without access to robot actions or demonstrations. In \texttt{Recycle Bottle}, a randomly positioned bottle must be placed in the correctly colored bin. In \texttt{Push Ramen}, a ramen package and distractor objects are placed at randomized positions. 
Performance uses a staged progress metric~\citep{kress2024robot} normalized to $[0,1]$ and defined in Appendix~\ref{section:benchmark_appendix_realworld}.

\subsection{Evaluating the Learned Latent Action Representation}
\subsubsection{Latent Action Policy Learning}
\label{sec:latent_policy}
First, we evaluate our learned latent action representations in the Imitation Learning from Observations (ILfO) setting~\cite{bruce2024genie}, where the agent learns to perform downstream tasks from expert demonstrations without access to the expert's actions. To enable imitation from action-free videos, we extract latent actions $z_i$ from expert videos using the trained LAM. 
These latent actions serve as targets for training a behavior cloning policy $\pi(z_i \mid o_i)$.
During inference, predicted latent actions are mapped to executable actions by nearest-neighbor retrieval against a small set of action-annotated reference videos. 

\begin{table*} [t]
\centering
\scriptsize
\setlength{\tabcolsep}{4pt}
\renewcommand{\arraystretch}{1.0}

\caption{Results from latent action policy learning evaluations on Crafter and OGBench. Each policy is evaluated over 200 random
environment seeds per task.}
\label{tab:crafter_bc_eval}
\begin{tabular}{lcccccc}
    \toprule
    & \multicolumn{3}{c}{\textbf{Crafter}} & \multicolumn{3}{c}{\textbf{OGBench}} \\
    \cmidrule(lr){2-4} \cmidrule(lr){5-7}
    Method & Wood Harvest (\%) & Avg.\ Wood & Wood Acquire (\%) & Open Drawer (\%) & Close Drawer (\%) & Push Button (\%) \\
    \midrule
    Random   & $25.0$      & $0.28$      & $27.5$      & $\hphantom{1}2.5$       & $14.5$      & $\hphantom{1}8.0$ \\
    LAPO     & $35.5$      & $1.12$      & $38.0$      & $\hphantom{1}4.5$       & $39.5$      & $\hphantom{1}9.5$ \\
    AdaWorld & $43.5$      & $1.60$      & $46.5$      & $\hphantom{1}7.5$       & $29.0$      & $\bm{16.5}$ \\
    CoMo & $80.5$ & $3.01$ & $79.5$ & $\hphantom{1}9.0$  & $34.0$ & $11.5$ \\
    \alg     & $\bm{86.0}$ & $\bm{4.46}$ & $\bm{84.0}$ & $\bm{13.5}$             & $\bm{42.0}$ & $15.0$ \\
    \bottomrule
\end{tabular}
\end{table*}

\begin{table*}[b]
\centering
\scriptsize
\begin{minipage}[t]{0.48\textwidth}
\centering
\caption{Visual planning on Crafter and Procgen.}
\label{tab:visual_planning_procgen}
\renewcommand{\arraystretch}{1.1}
\setlength{\tabcolsep}{5pt}
\begin{tabular}{lcccc}
\toprule
Method & Crafter & Jumper & Climber & Ninja \\
\midrule
LAPO           & $46.67\%$ & $58.33\%$ & $53.33\%$ & $48.33\%$ \\
AdaWorld       & $53.33\%$ & $56.67\%$ & $\mathbf{66.67\%}$ & $46.67\%$ \\
CoMo           & $61.67\%$ & $38.33\%$ & $45.00\%$ & $31.66\%$ \\
\alg{} w/o Adv & $26.67\%$ & $35.00\%$  & $38.33\%$ & $25.00\%$ \\
\alg{}         & $\mathbf{68.33\%}$ & $\mathbf{63.33\%}$ & $60.00\%$ & $\mathbf{56.67\%}$ \\
\bottomrule
\end{tabular}
\end{minipage}%
\hfill
\begin{minipage}[t]{0.48\textwidth}
\centering
\caption{Real-world manipulation results.}
\label{tab:real_world_results}
\setlength{\tabcolsep}{4pt}
\begin{tabular}{lcc}
\toprule
Method & Push Ramen & Recycle Bottle \\
\midrule
Scratch & $0.65$ & $0.85$ \\
\midrule
LAPA        & $0.73$ & $0.83$ \\
AdaWorld    & $0.50$ & $0.80$ \\
CoMo        & $0.48$ & $0.53$ \\
\alg{}      & $\mathbf{0.85}$ & $\mathbf{0.88}$ \\
\bottomrule
\end{tabular}
\end{minipage}
\end{table*}

\paragraph{Results}
As shown in Table~\ref{tab:crafter_bc_eval}, the \alg-based latent action policy outperforms all baselines across both Crafter tasks, achieving higher success rates and collecting substantially more wood on average in the \texttt{Greedy Wood Collection} task. On OGBench, \alg achieves the highest success rate on \texttt{Open Drawer} and \texttt{Close Drawer}, while AdaWorld performs slightly better on \texttt{Push Button} ($16.5\%$ vs.\ $15.0\%$ for \alg, $n{=}200$). Notably, despite being pretrained only on general survival videos without explicit supervision for wood harvesting, the learned latent actions still enable effective wood collection in Crafter, indicating disentanglement of action representations from task-specific rewards while preserving meaningful signals for imitation.

\subsubsection{Real-World Latent Action Pretraining}
\label{sec:realworld_pretraining}
Following the experimental protocol of \citet{ye2025latentactionpretrainingvideos}, we evaluate whether latent actions inferred by the LAM can substitute for ground-truth actions as a pretraining signal for real-world robot policies. We pretrain an ACT policy~\citep{zhao2023learning} to predict our inferred latent actions from visual observations, then freeze its vision encoder and finetune the remaining components on ground-truth actions for each task. We compare against the same architecture trained on ground-truth actions alone following the standard setup of~\citet{zhao2023learning} (\texttt{Scratch}). Task definitions, the staged progress metric, and the demonstration protocol are in Appendix~\ref{section:benchmark_appendix_realworld}.

\paragraph{Results}

Table~\ref{tab:real_world_results} reports mean normalized progress over 20 trials per task. \alg achieves the highest score on both tasks, with the clearest margin on the harder \texttt{Push Ramen} task, where it improves over \texttt{Scratch} by $0.20$. AdaWorld and CoMo do not exceed \texttt{Scratch} on either task,
suggesting that effective latent-action pretraining depends on learning useful action structure

\subsubsection{Latent Action Structure}
\label{sec:latent_structure}
\begin{figure*}[t]
    \centering
    \includegraphics[width=\textwidth]{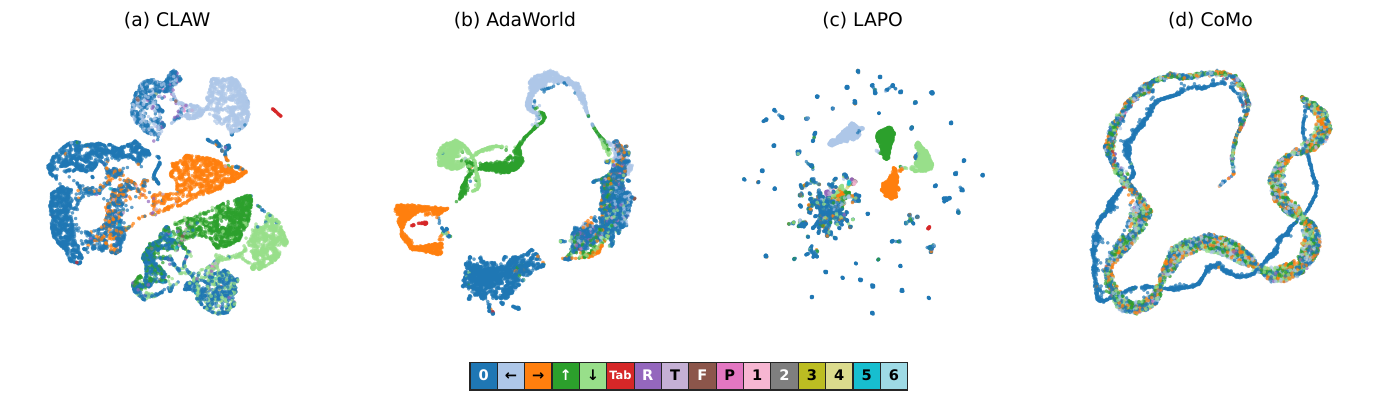}
    \caption{Latent action structure on Crafter. Latent actions from matched observation transitions are projected independently with UMAP for each method and colored by ground-truth Crafter actions, which are not used during training. Legend chips show the corresponding Crafter keyboard controls (arrows for movement; \texttt{0} denotes no-op).}
    \label{fig:umap_comparison}
\end{figure*}

We evaluate whether latent representations learned without action supervision recover structure aligned with the underlying action space. Using the same set of Crafter transitions for all methods, we extract latent actions with each pretrained model and independently project the resulting representations into two dimensions using UMAP. Ground-truth Crafter actions are used only to color points for visualization and are never provided during latent action learning.

As shown in Figure~\ref{fig:umap_comparison}, the methods exhibit qualitatively different latent geometries. \alg organizes transitions into locally action-consistent regions, with several behaviors, including movement directions, forming distinguishable regions in the embedding. 
Some overlap between action labels is expected, because the recorded Crafter action is the command the player issued, not the transition it produced. A blocked movement or a failed craft carries a non-noop label but produces almost no visual change, while a \texttt{Noop} can coincide with visible change and land far from the centroid of the no-op cluster.
Discrete labels and observable transition effects are therefore related but not equivalent, and we quantify both in Appendix~\ref{app:pseudo-noop}.

\subsubsection{Closest Action Retrieval}
\label{sec:closest_action_retrieval}

\begin{figure}[!t]
    \centering
    \begin{minipage}[t]{0.43\textwidth}
        \vspace{0pt}
        \centering
        \includegraphics[width=0.45\linewidth]{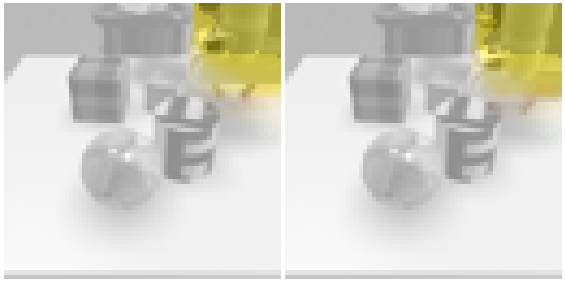}
        \vspace{0.1cm}
        \vspace{0.1cm}
        \par
        \begin{subfigure}[t]{0.24\linewidth}
            \centering
            \includegraphics[width=\linewidth]{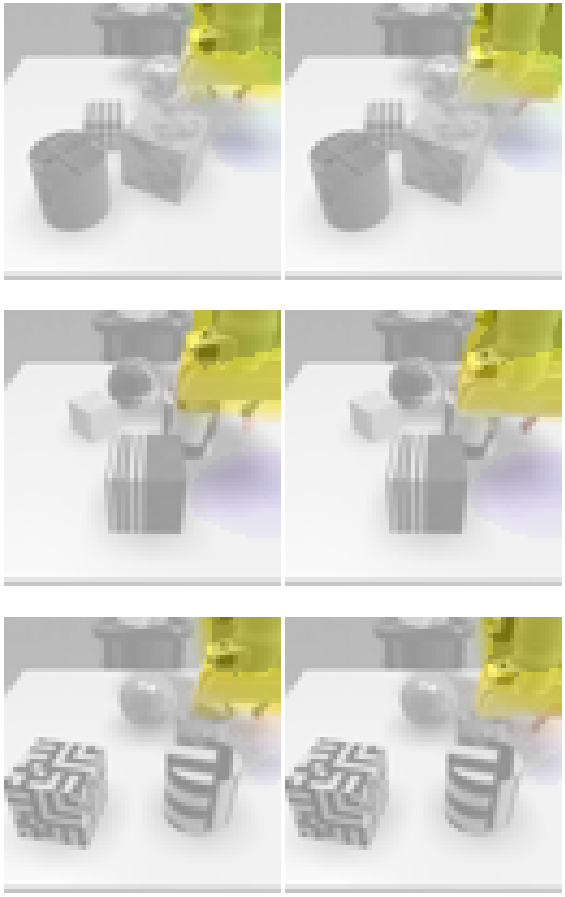}
            \caption{\tiny \alg}
            \label{fig:bottom1}
        \end{subfigure}%
        \hspace*{\fill}%
        \begin{subfigure}[t]{0.24\linewidth}
            \centering
            \includegraphics[width=\linewidth]{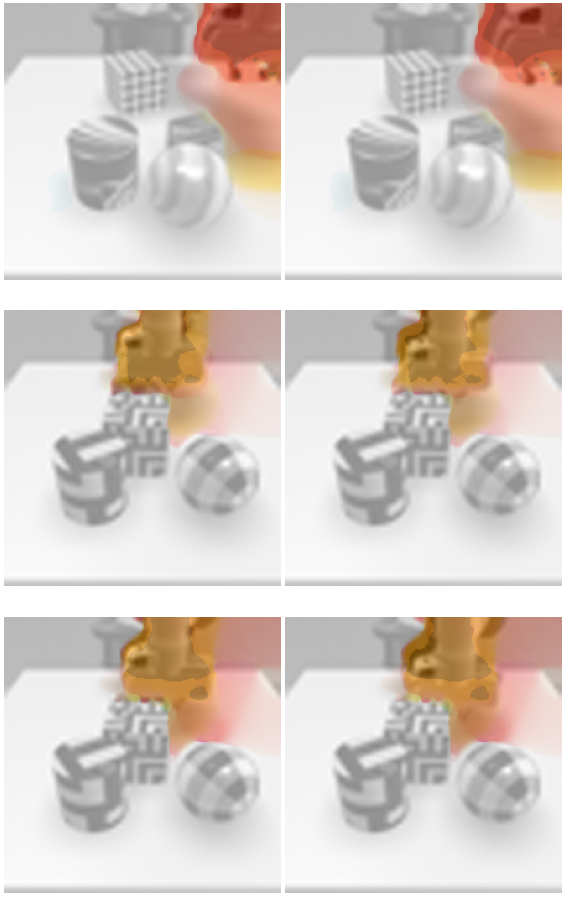}
            \caption{\tiny AdaWorld}
            \label{fig:bottom2}
        \end{subfigure}%
        \hspace*{\fill}%
        \begin{subfigure}[t]{0.24\linewidth}
            \centering
            \includegraphics[width=\linewidth]{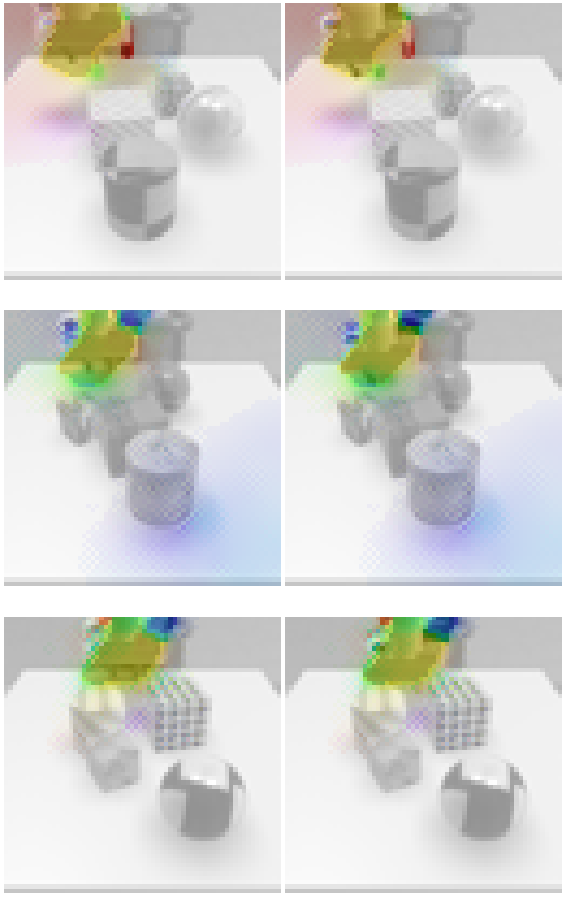}
            \caption{\tiny LAPO}
            \label{fig:bottom3}
        \end{subfigure}%
        \hspace*{\fill}%
        \begin{subfigure}[t]{0.24\linewidth}
            \centering
            \includegraphics[width=\linewidth]{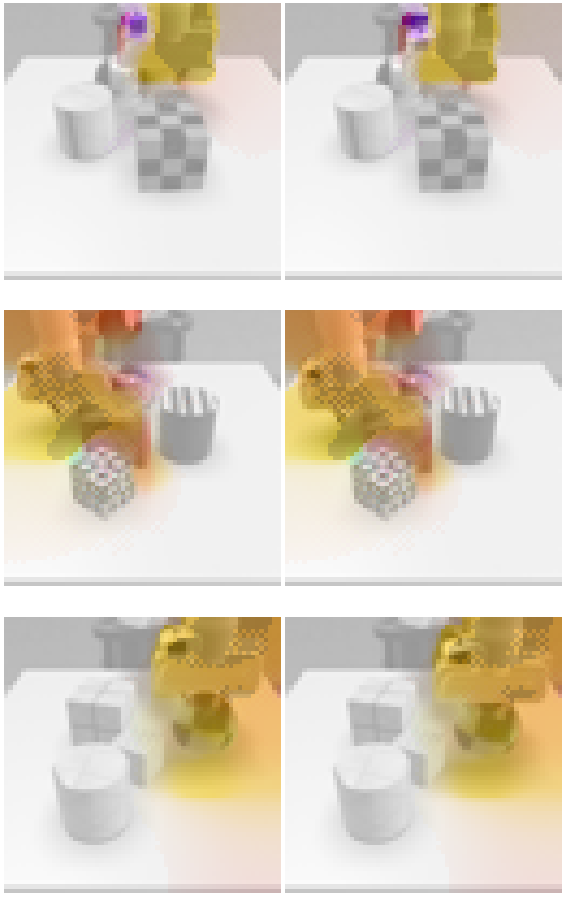}
            \caption{\tiny CoMo}
            \label{fig:bottom4}
        \end{subfigure}%
        \caption{
            \textbf{Closest Latent Action retrieval via $L_2$ nearest-neighbor:}
            (Top) Query transition.
            (Bottom) Top-3 neighbors retrieved via $L_2$ search in the latent space of \alg, AdaWorld, LAPO, and CoMo. All frames are shown as grayscale with the optical flow ($t\!\to\!t{+}1$) overlaid using the Middlebury color wheel.
        }
        \label{fig:action_retrieval}
    \end{minipage}%
    \hfill
    \begin{minipage}[t]{0.53\textwidth}
        \vspace{0pt}
        \centering
        \includegraphics[width=\linewidth]{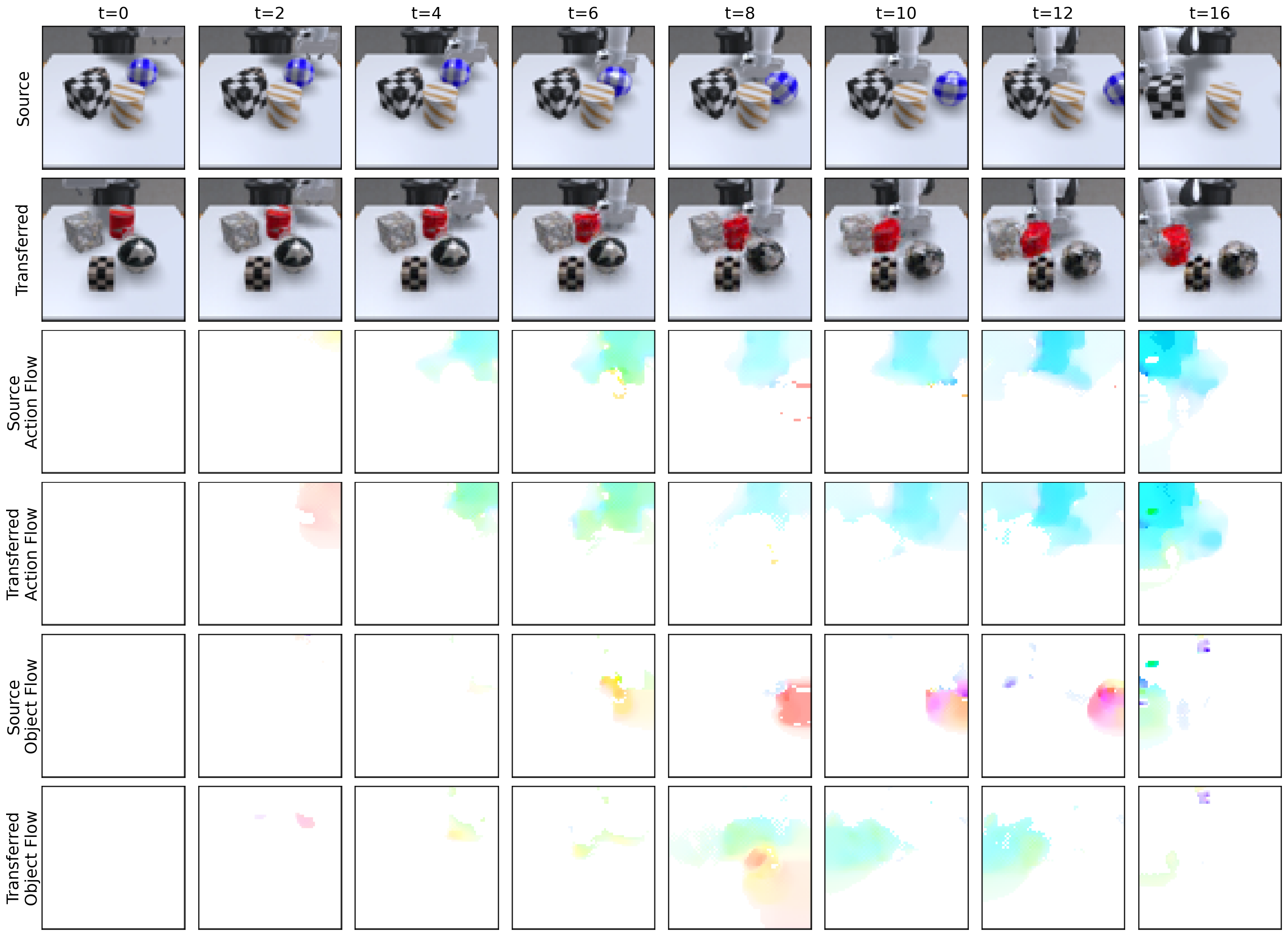}
        \caption{\textbf{Latent action transfer via action swap with optical-flow validation.} Row 1: source trajectory. Row 2: transferred rollout from a random target initial frame using source latent actions. Rows 3–4: source and transferred action (robot's gripper) optical flow. Rows 5–6: source and transferred object flow. 
        }
        \label{fig:action_swap}
    \end{minipage}
\end{figure}

Finally, we qualitatively evaluate the structure of the learned latent action space via $L_2$ 
nearest-neighbor retrieval. For each query, we sample a pair of consecutive frames 
representing an action transition and encode it using the LAM. We then retrieve the $n$ nearest latent actions from the dataset under the $L_2$ metric and visualize the corresponding frame transitions. Qualitative inspection of the retrieved neighbors allows us to assess whether the latent space organizes semantically similar actions into compact, consistent neighborhoods, reflecting the degree to which the learned representations capture meaningful behavioral structure rather than low-level visual similarity.

\paragraph{Results} As shown in  Figure~\ref{fig:action_retrieval}, across queries, the nearest neighbors in CLAW's learned latent action space exhibit similar end-effector motion, whereas non-action information like object identity varies substantially. This is in contrast to baseline methods, where nearest-neighbor retrieval reveals that the inferred latent actions confuse end-effector motion with other non-action scene properties. See Appendix~\ref{sec:retrieval_appendix} for further explanation and more examples in Robosuite and other environments.

\subsection{Evaluating the Dynamics Model}
\subsubsection{Visual Planning}
\label{sec:planning}
\begin{table}[t]
    \centering
    \caption{Visual planning evaluation results on Robosuite and RoboDesk tasks (mean $\pm$ standard error of the mean (SEM) over three control runs).
    }
    \label{tab:visual_planning_robosuite}
    \scriptsize
    \setlength{\tabcolsep}{3.5pt}
     \resizebox{\textwidth}{!}{%
    \begin{tabular}{l|c|ccccc}
        \toprule
        Method & Push & Open Slide & Blue Button & Green Button & Red Button & Upright Block\\
        \midrule
        Random & $\hphantom{0}0.00\pm0.00\%$  & $\hphantom{0}0.00\pm0.00\%$ & $10.0\pm2.72\%$ & $1.11\pm1.57\%$ & $1.11\pm1.57\%$ & $0.00\pm0.00\%$\\
        LAPO & $18.33\pm1.89\%$ & $11.67\pm3.40\%$ & $8.89\pm2.40\%$ & $6.67\pm1.83\%$ & $3.33\pm2.27\%$ & $15.56\pm4.54\%$ \\
        AdaWorld* & $\mathbf{64.60\pm1.52\%}$ & $\hphantom{0}5.83\pm2.85\%$ & $\mathbf{29.17\pm2.50\%}$ & $\mathbf{10.83\pm2.50\%}$ & $10.00\pm2.36\%$ & $5.00\pm0.96\%$ \\
        CoMo & $33.43\pm2.25\%$ & $11.10\pm0.91\%$ & $16.67\pm3.14\%$ & $3.33\pm1.92\%$ & $11.10\pm0.91\%$ &$\mathbf{38.89\pm0.91\%}$\\
        CLAW w/o Adv& $16.98\pm1.35\%$ & $\hphantom{0}6.37\pm2.27\%$ & $8.33\pm1.44\%$ & $6.67\pm1.57\%$ & $3.33\pm1.57\%$ & $2.22\pm1.81\%$ \\
        CLAW & $\mathbf{63.70\pm1.71\%}$ & $\mathbf{23.33\pm5.67\%}$ & $12.22\pm2.40\%$ & $\mathbf{11.11\pm0.91\%}$ & $\mathbf{25.56\pm0.91\%}$ & $\mathbf{43.33\pm4.71\%}$ \\
        \bottomrule
    \end{tabular}
    }
    \vspace{1mm}
    \raggedright{\footnotesize * Results reported from the original paper~\citet{gao2025adaworld}.}
\end{table}

Here, we use standard model predictive control (MPC) to plan sequences of latent actions through the world model toward a goal observation. Each predicted latent action is mapped to a true action by retrieving its nearest neighbor from a small set of annotated reference videos. We evaluate on goal-reaching success when executing the plan via the nearest-neighbor actions.

\paragraph{Results}
Table~\ref{tab:visual_planning_robosuite} reports visual planning success rates on the VP$^2$ benchmark across Robosuite and RoboDesk tasks. \alg achieves the highest success rate on four of the six tasks.
However, CLAW has lower success rates on the Green and Blue Button tasks. We hypothesize that these tasks are inherently challenging for latent action planning because the goal region is partially occluded and the fixed camera viewpoint makes it difficult to estimate vertical end-effector motion. 

Table~\ref{tab:visual_planning_procgen} presents visual planning success rates on Crafter and three Procgen tasks using the MPC-CEM planning procedure. \alg achieves the best performance on three of the four tasks and attains the highest average success rate overall. The consistent improvements over LAPO highlight the advantage of conditioning the world model on expressive continuous latent actions. %
Additional visual planning trajectories are in Appendix~\ref{section:visual_planning_supp}.

\subsubsection{Action Transfer}
\label{sec:action_transfer}
We qualitatively evaluate the ability of \alg to transfer actions across videos using an action swap task, following the  protocol from~\citet{gao2025adaworld}. In this experiment, a sequence of latent actions is extracted from a source video and applied to an initial observation from a different target video. 
To assess the quality of action transfer, we compare three sequences: the original source and target videos, and the new video with the source video actions applied to the target video initial observation. This comparison provides insight into whether the transferred actions induce the intended dynamics while preserving consistency with the target scene, highlighting the generalizability and robustness of the learned latent action representation.

\paragraph{Results}
As shown in Figure~\ref{fig:action_swap}, by $t=4$ the transferred end-effector motion closely matches the source behavior (rows three and four). The end-effector tracks the source trajectory from a different initial frame, matching what an absolute action space would produce. The objects, by contrast, move differently, consistent with their different initialization in the new scene (rows five and six). \alg therefore separates the action from the scene configuration it was inferred from, and learns plausible dynamics conditioned on it. Additional action transfer experiments in Procgen are in Appendix~\ref{section:action_transfer_supp}.

\subsection{Controllability and Leakage}
\label{sec:controllability}

\begin{table}[t]
    \centering
    \caption{Latent action controllability and future-observation leakage. Higher $\Delta_t$PSNR indicates better controllability, while lower leakage indicates that inferred actions produce consistent effects across different contexts. Bold marks the best $\Delta_t$PSNR; leakage is not bolded, as it is interpretable only jointly with $\Delta_t$PSNR. \textit{Future-copy} and \textit{Oracle} are anchors whose $\Delta_t$PSNR diverges.}
    \label{tab:controllability}
    \resizebox{\textwidth}{!}{%
    {\scriptsize
    \setlength{\tabcolsep}{5pt}
    \begin{tabular}{ll ccccc | cc}
        \toprule
        Environment & Metric
        & LAPO
        & AdaWorld
        & CoMo
        & \alg w/o Adv
        & \alg
        & Future-copy
        & Oracle \\
        \midrule

        \multirow{2}{*}{Crafter}
        & $\Delta_t$PSNR$\uparrow$
        & $\bm{9.14}$ & $0.15$ & $7.43$ & $2.47$ & $8.31$ & $\infty$ & $\infty$ \\
        & Leakage$\downarrow$
        & $0.42 \pm 0.15$ & $0.01 \pm 0.02$ & $0.44 \pm 0.14$
        & $0.05 \pm 0.06$ & $0.28 \pm 0.12$
        & $1.00 \pm 0.00$ & $0.00 \pm 0.00$ \\
        \midrule

        \multirow{2}{*}{Climber}
        & $\Delta_t$PSNR$\uparrow$
        & $4.90$ & $1.86$ & $5.59$ & $4.30$ & $\bm{5.83}$ & $\infty$ & $\infty$ \\
        & Leakage$\downarrow$
        & $0.68 \pm 0.09$ & $0.17 \pm 0.06$ & $0.80 \pm 0.08$
        & $0.42 \pm 0.12$ & $0.51 \pm 0.11$
        & $1.00 \pm 0.00$ & $0.00 \pm 0.00$ \\
        \midrule

        \multirow{2}{*}{Jumper}
        & $\Delta_t$PSNR$\uparrow$
        & $5.91$ & $1.84$ & $6.91$ & $4.18$ & $\bm{9.07}$ & $\infty$ & $\infty$ \\
        & Leakage$\downarrow$
        & $0.50 \pm 0.09$ & $0.17 \pm 0.06$ & $0.86 \pm 0.05$
        & $0.31 \pm 0.13$ & $0.48 \pm 0.12$
        & $1.00 \pm 0.00$ & $0.00 \pm 0.00$ \\
        \midrule

        \multirow{2}{*}{Ninja}
        & $\Delta_t$PSNR$\uparrow$
        & $5.27$ & $2.56$ & $7.32$ & $4.34$ & $\bm{9.02}$ & $\infty$ & $\infty$ \\
        & Leakage$\downarrow$
        & $0.35 \pm 0.10$ & $0.23 \pm 0.09$ & $0.86 \pm 0.06$
        & $0.24 \pm 0.15$ & $0.17 \pm 0.11$
        & $1.00 \pm 0.00$ & $0.00 \pm 0.00$ \\

        \bottomrule
    \end{tabular}}}
\end{table}

\paragraph{Controllability}
We evaluate controllability using $\Delta_t$PSNR~\citep{bruce2024genie}, which compares predictions conditioned on latents inferred from ground-truth transitions against predictions conditioned on latents drawn at random from the empirical latent distribution. Higher $\Delta_t$PSNR indicates stronger dependence of the predicted future on the latent action.

\paragraph{Leakage}
We introduce a \emph{pairwise leakage score} measuring whether latent actions encode context-specific information beyond the underlying semantic action. Within groups of transitions sharing the same ground-truth action, we transfer each latent to the other contexts and compare its cross-context prediction error against its matched-context error. The score lies in $[-1,1]$: values near zero indicate cross-context consistency, positive values indicate source-context dependence. Table~\ref{tab:controllability} includes two reference setups. \textit{Oracle} predicts each target context's true next frame and has zero leakage; \textit{Future-copy} simulates a latent action that encodes only the future frame and attains the maximum leakage score of $1$.
Full definitions, the evaluation protocol, and spatial leakage maps are in Appendix~\ref{app:leakage_metric}.

\paragraph{Results}
A latent action representation can fail in two ways. In the first, the latent retains information about the context it was inferred from, so the world model makes good predictions in that context but inferred actions transfer poorly, resulting in high $\Delta_t$PSNR but high leakage. As seen in Table~\ref{tab:controllability}, CoMo shows this pattern most clearly. In the second, the world model makes little use of the latent, resulting in low $\Delta_t$PSNR and driving leakage toward zero for reasons unrelated to representation quality. \alg w/o Adv reports lower leakage than \alg{} on three environments, but this is only because its world model predictions depend only weakly on the inferred latent actions, as indicated by its low $\Delta_t$PSNR and consistent with its low planning success on Crafter and Robosuite Push (Tables~\ref{tab:visual_planning_procgen},~\ref{tab:visual_planning_robosuite}). \alg avoids both failure modes, attaining the highest $\Delta_t$PSNR on Climber, Jumper, and Ninja while keeping leakage below LAPO and CoMo on most environments.

%% file: sections/06_discussion.tex
\section{Discussion}
\label{sec:discussion}

\alg's latent space is organized by action rather than by observation. Nearby latents correspond to similar transitions across visually unrelated scenes (Sec.~\ref{sec:latent_structure}, Sec.~\ref{sec:closest_action_retrieval}), and perturbing a latent degrades downstream performance gradually rather than abruptly, suggesting action information is spread across the space rather than concentrated in isolated points (Appendix~\ref{sec:more_analysis_latent}). This matters because it allows a small set of annotated transitions to ground the inferred latents to executable actions, which our policy-learning and planning evaluations rely on. \citet{schmidt2024learningactactions} established that discrete latent actions recover this structure; our results show it also emerges in a continuous space shaped by an adversarial objective, without quantization or a codebook.

Successful planning depends on latents that transfer: the world model must predict the same behavior from a given latent action regardless of the initial observation. \alg's low scores on our proposed leakage metric and high $\Delta_t$PSNR indicate that it learns an action representation that consistently controls the world model across contexts.
Sec.~\ref{sec:action_transfer} shows this qualitatively, with latents extracted from one trajectory producing the same behavior in a different context. Ablations confirm that adversarial latent regularization is the cause: \alg w/o Adv reports even lower leakage in Table~\ref{tab:controllability}, but only because its predictions barely depend on the latent, and its planning success falls accordingly.

%% file: sections/05_conclusion.tex
\section{Conclusion}
\label{section:conclusion}

We introduce an end-to-end self-supervised method for jointly learning a world model and continuous latent action representations from unlabeled video. Adversarial regularization addresses future-observation leakage without vector quantization or action supervision, and across imitation learning from observation, visual planning, and real-world policy pretraining, \alg achieves the best overall performance among evaluated latent action methods. Its main limitation, shared with video world models generally, is that the model learns what transitions look like rather than which ones are permitted, so it can generate futures that the environment would never produce. Addressing this likely requires interaction rather than observation alone; we discuss it and other limitations in Appendix~\ref{sec:limitations}.

%% file: supp.tex
\section{Extended Related Work}
\subsection{Comparison with Continuous Latent Action Methods}
\label{sec:continuous_lam_comparison}

Table~\ref{tab:continuous_lam_comparison} summarizes the main supervision, pretraining, and downstream-use differences between \alg{} and recent continuous latent action methods. We focus the table on high-level properties and discuss the specific learning mechanisms below.

\begin{table}[H]
    \centering
    \caption{\textbf{Comparison of continuous latent action learning methods.}
    A checkmark indicates that the property holds for the method.}
    \label{tab:continuous_lam_comparison}
    {\scriptsize
    \setlength{\tabcolsep}{5pt}
    \begin{tabular}{lcccc}
        \toprule
        Property & \alg & CLAM & UniSkill & CoMo \\
        \midrule

        Latent space learned from action-free video alone
        & \checkmark & \xmark & \checkmark & \checkmark \\

        No pretrained perceptual components
        & \checkmark & \checkmark$^{*}$ & \xmark & \xmark \\

        Single-stage training of representation and dynamics
        & \checkmark & \xmark & \xmark & \xmark \\

        Dynamics model used for planning
        & \checkmark & \xmark & \xmark$^{\dagger}$ & \xmark \\

        \bottomrule
    \end{tabular}}

    \vspace{0.3em}
    \raggedright
    \scriptsize
    $^{*}$CLAM learns its latent representation without pretrained perception
    in its primary setting; see text.
    $^{\dagger}$UniSkill uses its forward skill dynamics model to generate
    sub-goal images for policy conditioning, but does not optimize over it as a
    dynamics model for planning.
\end{table}

CLAM~\citep{liang2025clam} jointly trains a latent inverse dynamics model and a
forward dynamics model through next-observation reconstruction. Its latent
representation is additionally shaped by an action decoder trained on
action-labeled play data, with supervised gradients propagating into the latent
inverse dynamics model; this joint training is what keeps the latent space
decodable. Although CLAM's inverse dynamics model is trained on unlabeled
observations, its latent action space is therefore not learned from action-free
video alone. After pretraining, the forward dynamics model is discarded and the
latent representation is used to relabel observation-only demonstrations for a
separately trained policy.

UniSkill~\citep{kim2025uniskill} jointly trains an inverse skill dynamics model
and a diffusion-based forward skill dynamics model, with future-frame
prediction supervising the learned skill representation and an explicit
information bottleneck preventing trivial solutions. It requires no
ground-truth action labels, but builds on pretrained components including
InstructPix2Pix initialization for the forward model and a frozen
DepthAnythingV2 depth estimator, whose depth maps the inverse model consumes in
place of raw RGB. The skill representation is subsequently used to train a
skill-conditioned policy.

CoMo~\citep{yang2025comolearningcontinuouslatent} learns continuous latent
motion from action-free Internet videos using an MAE-pretrained ViT as its
visual feature extractor. It combines three mechanisms to prevent shortcut
learning: early temporal feature differencing, a temporal contrastive objective
that emphasizes foreground motion and temporal direction, and a constrained
latent dimensionality following the information bottleneck principle. The
forward dynamics decoder is used only to supervise the inverse model during
this first stage; the resulting latent motion then serves as pseudo-action
supervision for a separately trained policy.

In contrast, \alg discourages the inferred latent from encoding future visual information adversarially rather than through discrete quantization, an explicit information bottleneck, or action supervision, while retaining a continuous latent representation. The
predictive model is retained after training and used directly for controllable
prediction and goal-directed planning, rather than serving only as a
representation-learning objective.

\section{\alg Implementation Details}
\label{sec:claw_implementation}
\begin{figure}[h]
    \centering
    \begin{subfigure}{0.4\textwidth}
        \centering
        \includegraphics[width=\linewidth]{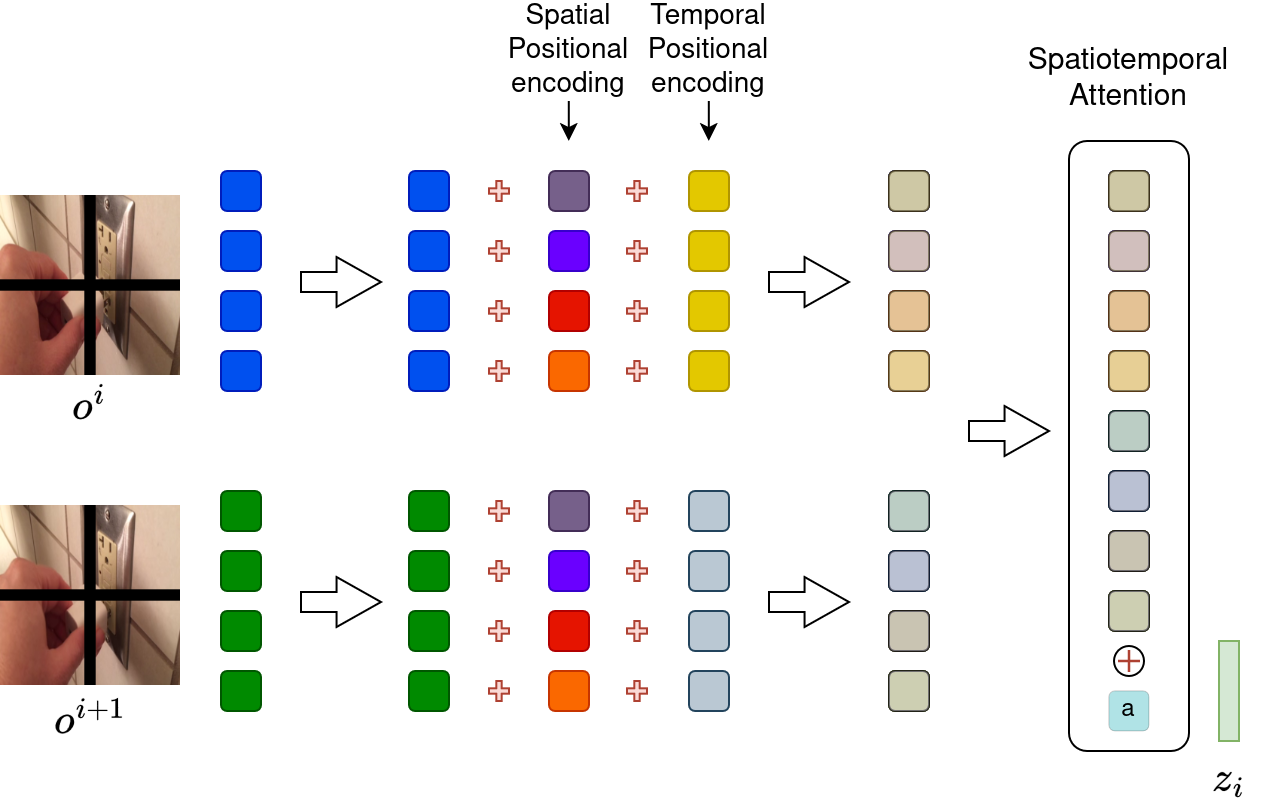}
        \caption{Latent Action Model}
        \label{fig:action_encoder}
    \end{subfigure}
    \rule{0.5pt}{0.32\textwidth}
    \hspace{0.02\textwidth}
    \begin{subfigure}{0.54\textwidth}
        \centering
        \includegraphics[width=\linewidth]{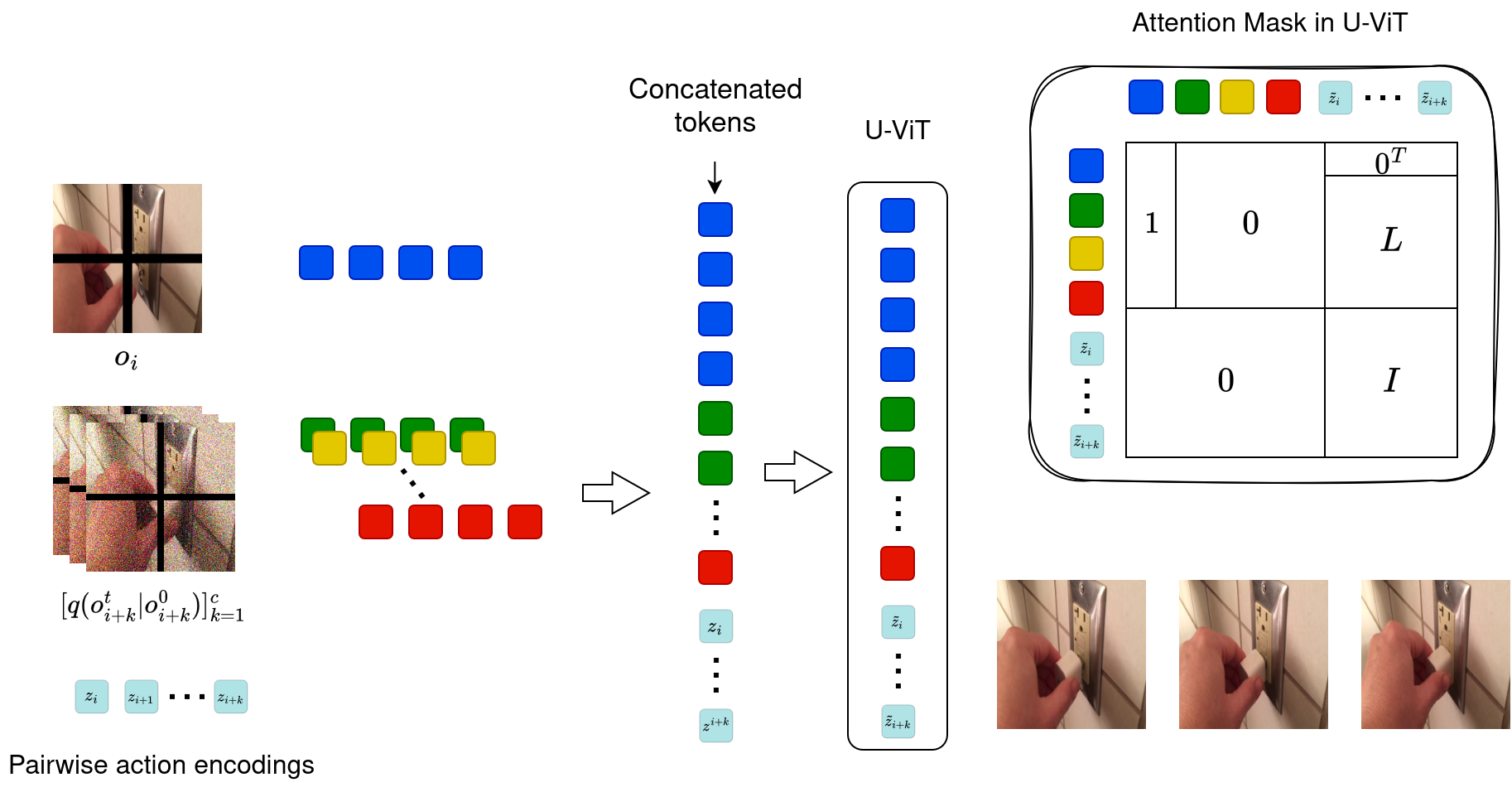}
        \caption{Diffusion-based World Model}
        \label{fig:worldmodel}
    \end{subfigure}
    \vspace{1.0em}
    \caption{\alg consists of two main components: a ViT based latent action model (LAM) and a U-ViT based diffusion world model. The LAM is designed to infer compact latent actions directly from visual state transitions, while the diffusion model predicts future observations conditioned on these latent actions. Under the causal attention mask, where $L$ denotes a lower-triangular block and $I$ an identity block, each noised target token attends only to the other tokens in the same frame, to the initial frame tokens, and to the action tokens corresponding to transitions from the initial frame up to the timestep of the noised target frame. This ensures that the model cannot access future information while learning to denoise based on valid action-conditioned dynamics. }
\end{figure}

\subsection{Latent Action Model}
 \alg's LAM uses a ViT backbone~\citep{dosovitskiy2020image, he2022masked} as a transition model that maps consecutive observations to a continuous action code. Given a pair of frames $\{o_{i}, o_{i+1}\}$, the encoder produces a latent action $z_i \in \mathbb{R}^n$ as illustrated in Figure~\ref{fig:action_encoder}: $z_i = \textrm{LAM}_{\theta}(o_{i}, o_{i+1})$. 

 These patches are projected into the model's hidden dimension and augmented with both spatial and temporal positional encodings. The spatial positional encoding is shared across corresponding spatial locations over time, enabling the model to preserve spatial correspondence between frames, while the temporal positional encoding identifies the timestep associated with each frame. Together, these encodings produce a structured spatiotemporal token representation that captures both appearance and temporal evolution.
 
 A learnable action token is concatenated with the spatiotemporal patch tokens and processed using a spatiotemporal Vision Transformer encoder. The full sequence is processed by a ViT-style transformer encoder, which outputs the latent action that summarizes the transition between the two input observations. The final representation of this token is projected into a compact latent action vector, which is subsequently $\ell_2$ normalized to produce the latent action embedding. This latent representation serves as a compressed description of the transition between observations and forms the core of the latent action model.

\subsection{Diffusion-based World Model}
\label{sec:claw_diffusion_details}
Our world model is a diffusion-based video predictor built on a U-ViT~\citep{bao2023all} backbone. It learns to denoise target frames conditioned on latent actions extracted from observed transitions, so rollouts proceed by applying a sequence of inferred actions while iteratively refining predictions from an initial scene. Because generation is autoregressive in time, the model must respect temporal causality: predictions at step $t$ may use the starting observation and actions up to $t$, but must not depend on future frames or future actions. We therefore train with a modified causal attention mask that enforces this structure in the token sequence rather than relying on the network to learn it implicitly.

The mask is designed to block information leakage at the token level. Noisy target tokens cannot attend to future action tokens, nor to visual tokens from timesteps beyond their own. As illustrated in Figure~\ref{fig:worldmodel}, each noised target token is allowed to attend only to (i)~other tokens within the same frame, (ii)~tokens from the initial conditioning frame, and (iii)~action tokens associated with transitions from the initial frame through the timestep of the noised target. Intuitively, a token being denoised at time $t$ may aggregate spatial context within its frame, retain access to the rollout's starting state, and read the action history that explains how the scene should evolve up to $t$, but it cannot ``peek'' at later actions or later observations. This restricted attention pattern keeps denoising aligned with valid action-conditioned dynamics and makes sequential transfer reliable: each generated frame is produced from causal action context, not from future information in the training or inference sequence.

We model frame generation with a variance-preserving diffusion process. Given a clean target frame $o_0$ (and conditioning on a query frame and latent action), the forward noising distribution is
\begin{equation}
  q(o_t \mid o_0) = \mathcal{N}\!\big(\sqrt{{\alpha}_t}\, o_0,\; {\beta}_t\, I\big),
  \qquad
  x_t = \sqrt{{\alpha}_t}\, o_0 + \sqrt{{\beta}_t}\,\epsilon,
  \quad \epsilon \sim \mathcal{N}(0,I),
\end{equation}
where $t \in [0,1]$ indexes noise level, ${\beta}_t = 1 - {\alpha}_t$.

\subsection{\alg Configuration}
Training and hyperparameter details for \alg are provided in Table~\ref{tab:config}.

\begin{table}[t]
\caption{Training and model configuration for the latent action model and diffusion world model.}
\centering
\small
\begin{tabular}{ll}
\toprule
\textbf{Component} & \textbf{Configuration} \\
\midrule

Training Steps & 100,000 \\
Batch Size & 512 \\

\midrule
Optimizer & AdamW \\
Learning Rate & $3 \times 10^{-4}$ \\
Weight Decay & 0.03 \\
Betas & $(0.99, 0.99)$ \\

\midrule
Warmup Steps & 5,000 \\

\midrule
\multicolumn{2}{l}{\textbf{Latent Action Model (LAM)}} \\
Spatiotemporal Encoder & ViT \\
Input Resolution & $64 \times 64$ \\
Patch Size & $4 \times 4$ \\
Embedding Dimension & 512 \\
Depth & 8 \\
Attention Heads & 8 \\
Latent Action Dimension & 32 \\
Latent Normalization & $\ell_2$ normalization \\

\midrule
\multicolumn{2}{l}{\textbf{Diffusion World Model}} \\
Backbone & U-ViT \\
Embedding Dimension & 512 \\
Depth & 12 \\
Attention Heads & 8 \\
Patch Size & $4 \times 4$ \\
Diffusion Process & Variance-preserving SDE \\
$\beta_0$ & 0.1 \\
$\beta_1$ & 20 \\
Conditioning & Query frame + noised target + latent actions \\
Attention Mask & Structured causal masking \\
Sampler & Probability-flow ODE \\
Sampling Steps & 50 \\

\bottomrule
\end{tabular}
\label{tab:config}
\end{table}

\subsection{Latent Action Policy Training Pipeline}
\label{app:training_pipeline}

\subsubsection{Latent Action Extraction}
For each partition, latent actions are computed from consecutive pairs of visual observations using the Latent Action Model (LAM). Given two successive frames $(o_t, o_{t+1})$, a short temporal sequence is formed (i.e., the action chunk) and passed through LAM.
The LAM produces a latent action embedding $z_t = LAM(o_t, o_{t+1})$. We use a latent vector size of 32 for all our experiments. These latent vectors represent the transformation between consecutive observations.
For each video, we extract latent actions and save them together with the corresponding video frames: $\mathcal{D} = \{(o^i, z_i)\}_{i=1}^{T}.$

\subsubsection{Latent Action Policy Network}

The policy network is implemented as a convolutional neural network based on a ResNet-18~\citep{he2016deep} backbone. The final fully connected layer is replaced with a linear projection to the latent action dimension 32 to match our latent action dimension. Given an observation $o_t$, the policy predicts a latent action $\hat{z}_t = \pi_\theta(o_t)$
where $\pi_\theta$ denotes the policy network with parameters $\theta$. The output vector is normalized to unit norm to match the geometry of the latent action space learned by the encoder.

\subsubsection{Training Objective}

The policy is trained in a supervised regression setting using the extracted latent actions as targets. For a batch of observations and latent labels $(o_t, z_t)$, the training objective minimizes the L1 distance between predicted and target latent actions:
\[
\mathcal{L}(\theta)=\|\pi_\theta(o_t)-z_t\|_1.
\]
The Adam optimizer is used for training with a learning rate of $10^{-4}$ and a batch size of 32. Training is performed for 100 epochs.

\subsection{Visual Planning}
\label{sec:visual_planning_appendix}

We perform planning using Model Predictive Control combined with the Cross-Entropy Method (MPC-CEM)~\citep{deboer2005cem}. At each planning step, the planner searches over sequences of latent actions using the learned world model. CEM samples candidate latent action sequences $\tilde{z}_{1:H} = \{\tilde{z}_1, \tilde{z}_2, \ldots, \tilde{z}_H\}$, where $\tilde{z}_i \in \mathbb{R}^d$. Because the world model is trained on latent actions produced by the LAM, arbitrary samples from the CEM proposal distribution may lie outside the latent distribution encountered during training. We therefore project each sampled latent action onto the empirical support of the learned latent space by retrieving its nearest neighbor from a reference latent set $\mathcal{Z}_{\mathrm{ref}}$:
\begin{equation}
    z_i = \operatorname{NN}\left(\tilde{z}_i,\mathcal{Z}_{\mathrm{ref}}\right).
\end{equation}
The resulting sequence $z_{1:H}=\{z_1,z_2,\ldots,z_H\}$ is used to condition the world model. This constrains planning queries to latent actions produced by the LAM and prevents the planner from exploiting out-of-distribution latent vectors on which the world model was not trained.

For each candidate sequence, the world model $f_\theta$ predicts a future rollout
\[
\hat{o}_{t+1:t+H}=f_\theta(o_t,z_{1:H}).
\]
The planner minimizes a visual goal-reaching objective defined on the terminal predicted observation,
\[
J(z_{1:H})=d\!\left(\hat{o}_{t+H},o_g\right),
\]
where $d(\cdot,\cdot)$ is a task-dependent visual distance to the goal. For Procgen, $d$ is an $\ell_2$ distance between frames. For Crafter, $d$ is a patch-based distance that measures local disagreement with the goal image. The same distance is used to determine task termination.

We use a CEM-style search procedure with elite retention and fixed isotropic exploration scale. At the first iteration, $N$ candidate latent-action sequences are sampled from $\mathcal{N}(\mu,\sigma_0^2 I)$, where $\mu$ is initialized from the unused suffix of the previous MPC plan when available and otherwise to zero. After projection and evaluation, the top $K$ projected sequences are retained and $\mu$ is set to the single lowest-cost projected sequence. In subsequent iterations, the retained elites are carried forward unchanged, while the remaining candidates are formed by adding those elites to fresh samples from $\mathcal{N}(\mu,\sigma_0^2 I)$. Thus, the exploration scale $\sigma_0$ remains fixed throughout optimization, and the proposal mean is determined by the current best projected sequence rather than by averaging the elite population.

After optimization, the first $k$ latent actions of the best projected sequence are grounded to executable environment actions using the paired reference set $\mathcal{R}=\{(\mathbf{z}^{(m)},\mathbf{a}^{(m)})\}_{m=1}^{M}$ and executed before replanning. The unused suffix of the selected plan is used to initialize the next planning step. Thus, nearest-neighbor retrieval serves two distinct roles: projection onto the empirical latent support before world model evaluation, and grounding of the selected latent actions to executable environment actions.

For the Procgen and Crafter planning experiments, we run five CEM updates per MPC iteration, sample $N=64$ candidate sequences with planning horizon $H=5$, and retain the top $K=32$. After optimization, the first $k=5$ actions of the best sequence are executed, so no suffix remains for warm starting. Planning continues for at most 20 MPC iterations or until $d(o_t,o_g)<\epsilon$. World-model rollouts use 50 diffusion denoising steps.

For VP$^2$, we use the benchmark's provided planner, which scores sampled action sequences using a cost that combines pixel-wise MSE to the goal image with a learned task-success classifier. We use the benchmark's recommended weights $\lambda_1=0.5$ and $\lambda_2=10$.

\begin{algorithm}
\caption{Model-Based Visual Planning with MPC-CEM}
\label{alg:MPC-CEM}
\small
\begin{algorithmic}[1]
\REQUIRE World model $f_\theta$, environment $\mathcal{E}$, reference sets $\mathcal{R}$ and $\mathcal{Z}_{\mathrm{ref}}$
\REQUIRE Initial observation $o_0$, goal $o_g$, horizon $H$, execution horizon $k$
\REQUIRE CEM iterations $I$, samples $N$, elites $K=N/2$, noise scale $\sigma_0$, threshold $\epsilon$
\ENSURE Executed actions $\mathcal{A}_{\mathrm{exec}}$
\STATE $o_t \leftarrow o_0$, $\mathcal{A}_{\mathrm{exec}} \leftarrow \emptyset$, $t \leftarrow 0$
\STATE $\mathbf{z}^{\mathrm{warm}} \leftarrow \emptyset$
\WHILE{$d(o_t,o_g)\geq\epsilon$ and $t<t_{\max}$}
    \IF{$\mathbf{z}^{\mathrm{warm}}=\emptyset$}
        \STATE $\mu \leftarrow \mathbf{0}$
    \ELSE
        \STATE $\mu \leftarrow \mathbf{z}^{\mathrm{warm}}$, zero-padded to length $H$
    \ENDIF
    \STATE $\mathcal{Z}_{\mathrm{elite}} \leftarrow \emptyset$
    \FOR{$j=1$ to $I$}
        \IF{$j=1$}
            \STATE Sample $N$ sequences $\tilde{\mathbf{z}}_{1:H}^{(i)} \sim \mathcal{N}(\mu,\sigma_0^2 I)$
        \ELSE
            \STATE Keep the $K$ sequences in $\mathcal{Z}_{\mathrm{elite}}$
            \STATE Sample $\boldsymbol{\delta}^{(\ell)} \sim \mathcal{N}(\mu,\sigma_0^2 I)$, $\ell=1,\ldots,K$
            \STATE Form $K$ additional candidates
            \[
            \tilde{\mathbf{z}}_{1:H}^{(K+\ell)}
            \leftarrow
            \mathbf{z}_{1:H}^{\mathrm{elite},(\ell)}
            +
            \boldsymbol{\delta}^{(\ell)},
            \qquad
            \ell=1,\ldots,K
            \]
        \ENDIF
        \FOR{$i=1$ to $N$}
            \FOR{$h=1$ to $H$}
                \STATE $z_h^{(i)} \leftarrow \arg\min_{z\in\mathcal{Z}_{\mathrm{ref}}}\|\tilde{z}_h^{(i)}-z\|_2$
            \ENDFOR
            \STATE $\hat{o}_{t+1:t+H}^{(i)} \leftarrow f_\theta(o_t,\mathbf{z}_{1:H}^{(i)})$
            \STATE $J_i \leftarrow d(\hat{o}_{t+H}^{(i)},o_g)$
        \ENDFOR
        \STATE $\mathcal{Z}_{\mathrm{elite}} \leftarrow$ $K$ lowest-cost projected sequences
        \STATE $i^\star \leftarrow \arg\min_i J_i$
        \STATE $\mathbf{z}_{1:H}^{*} \leftarrow \mathbf{z}_{1:H}^{(i^\star)}$
        \STATE $\mu \leftarrow \mathbf{z}_{1:H}^{*}$
    \ENDFOR
    \FOR{$h=1$ to $k$}
        \STATE $(\cdot,a_h^*) \leftarrow \operatorname{NN}_{\mathcal{R}}(z_h^*)$
    \ENDFOR
    \STATE $o_{t+k} \leftarrow \mathcal{E}(o_t,\mathbf{a}_{1:k}^{*})$
    \STATE $\mathcal{A}_{\mathrm{exec}} \leftarrow \mathcal{A}_{\mathrm{exec}} \cup \mathbf{a}_{1:k}^{*}$
    \STATE $\mathbf{z}^{\mathrm{warm}} \leftarrow \mathbf{z}_{k+1:H}^{*}$
    \STATE $o_t \leftarrow o_{t+k}$, $t \leftarrow t+k$
\ENDWHILE
\STATE \textbf{return} $\mathcal{A}_{\mathrm{exec}}$
\end{algorithmic}
\end{algorithm}

\section{Benchmark and Dataset}
\label{section:benchmark_appendix}
\begin{figure}[!htbp]
    \centering
    \includegraphics[width=0.75\linewidth]{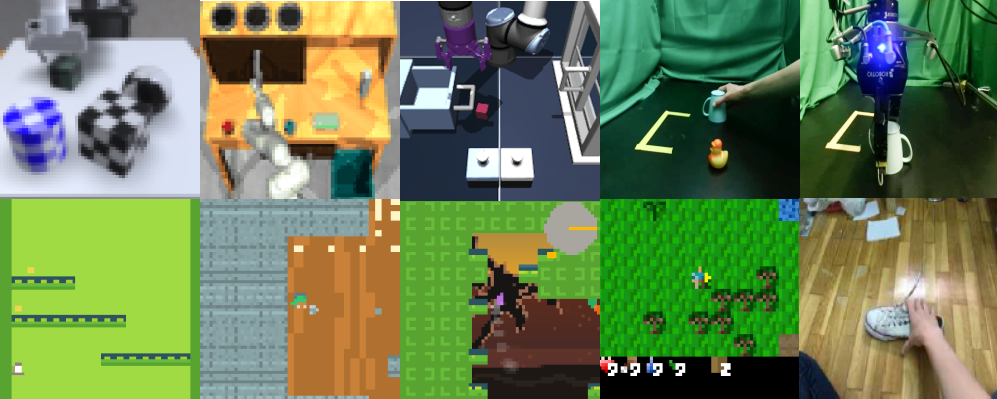}
    \caption{Benchmarks and environments that we evaluate \alg on. Top row (left-to-right): VP2 Robosuite, VP2 RoboDesk, OGBench Scene, Real-world human, Real-world robot. Bottom row (left-to-right): Procgen Climber, Procgen Ninja, Procgen Jumper, Crafter, Something-Something v2.}
    \label{fig:environmentsused}
\end{figure}

In this Appendix section, we give further details about each benchmark experiment and the corresponding tasks. See Table \ref{tab:evaluationoverview} for an overview and Figure~\ref{fig:environmentsused} for visuals.

\begin{table}[h]
\caption{Breakdown of CLAW benchmarks: environments, latent action world model (LAWM) training details, task adaptation details, and evaluation details. For evaluation methods, we abbreviate Closest Action Retrieval (AR), Action Transfer (AT), Visual Planning (VP), Latent Action Policy Learning (PL), Latent Action Pretraining (LAP), and Controllability (C).}
\centering
\small
\begin{tabular}{l|l|r|l|r|l|r}
\hline
\textbf{Env} & \multicolumn{2}{c|}{\textbf{LAWM Training}} & \multicolumn{2}{c|}{\textbf{Task Adaptation}} & \multicolumn{2}{c}{\textbf{Evaluation}}\\
\cline{2-3} \cline{4-5} \cline{6-7}
 & \textbf{Dataset} & \textbf{Training Data} & \textbf{Method} & \textbf{\# Trajs} & \textbf{Methods} & \textbf{\# Tasks} \\
\hline
\multirow{2}{*}{VP$^2$} 
 & Robosuite & 175k frames & Retrieval & 40 & AR, AT, VP & 1\\
 & RoboDesk & 13.9k videos & Retrieval & 40/task & AR, AT, VP & 5\\
\hline
OGBench 
 & Oracle & 826k frames & Retrieval & 25/task & AR, AT, VP, PL & 3\\
\hline
Crafter 
 & Human Expert & 62k frames & Retrieval & 20 & AR, AT, VP, PL, C & 2\\
\hline
\multirow{3}{*}{Procgen} 
 & Climber & 70k frames & Retrieval & 40 & AR, AT, VP & 1 \\
 & Jumper & 70k frames & Retrieval & 40 & AR, AT, VP & 1 \\
 & Ninja & 70k frames & Retrieval & 40 & AR, AT, VP & 1 \\
\hline
Real-World 
 & Something-Something V2 & 168k videos & Fine-tuning & 50/task & LAP & 2\\
\hline
\end{tabular}
\label{tab:evaluationoverview}
\end{table}

\subsection{VP2}
For the VP$^2$~\citep{tian2023vp2} benchmark, we train \alg and the baselines on the benchmark datasets and evaluate their performance on the provided planning tasks without additional task-specific supervision. For Robosuite, we use the 5k video dataset and evaluate on 100 visual planning problems for the Push task derived from~\citet{zhu2020robosuite}. For RoboDesk~\citep{kannan2021robodesk}, we use 13.9k training videos and evaluate on five tabletop tasks.

\subsection{OGBench}
For the OGBench ~\citep{park2025ogbench} benchmark, we collected 750 videos using the environment's built-in oracles to train our world models, then evaluated latent action policy learning on three downstream tasks: Open Drawer, Close Drawer, and Push Button.

\subsubsection{OGBench Data Generation}
\label{section:ogbench_data}
We generated data within OGBench's~\citep{park2025ogbench} \texttt{Scene} environment, a manipulation setup with a single robot arm, sliding handle drawer, sliding window, a cube, and two buttons. We employed OGBench's default \texttt{PlanOracle}, a scripted, non-Markovian expert policy, to collect 750 total trajectories with \texttt{noise} set to zero, filtering the tasks to just `drawer' and `button'. For each trajectory, we set a maximum episode length of 1001 (\texttt{max\_episode\_length}=1001), which terminates early upon task completion. For the pixel-based observations, we use the untransparent robotic arm setting (\texttt{pixel\_transparent\_arm}=False). 

\subsection{Crafter}
In the Crafter~\citep{hafner2021crafter} environment, we evaluate the latent action models on two tasks. The \texttt{Greedy Wood Collection} task performance is measured by two distinct metrics, the first being success rate, the percentage of trajectories in which the agent collects at least one piece of wood. The second measures the average number of wood units collected, reflecting the agent’s ability to maximize cumulative reward over a long horizon. The \texttt{Single Wood Collection} is also measured in success rate. All policy evaluations are performed by deploying the learned model in the environment and measuring its performance over $200$ random seeds over the environment state. 

\subsubsection{Crafter Benchmark Data Generation}
\label{section:crafterdata}
For our imitation learning, we generated expert video samples in the Crafter game environment~\citep{hafner2021crafter} using a PPO agent based on Achievement Distillation~\citep{moon2023discovering}. We trained separate agents on two distinct tasks, \texttt{Single Wood Collection} and \texttt{Greedy Wood Collection}.

\paragraph{Agent Setup and Training Details} We use PPO\_AD (Achievement Distillation)~\citep{moon2023discovering} implemented from the open-sourced codebase, using the default \texttt{PPOADModel} / \texttt{PPOADAlgorithm} architecture and losses as defined in the original paper and repository. Unless otherwise noted, we follow the repository defaults, with the following key settings: 8 parallel workers (\texttt{nproc}=8), 512-step rollouts (\texttt{nstep}=512), discount $\gamma=0.95$, and GAE $\lambda=0.65$. PPO optimization uses 3 epochs per update with 8 minibatches (\texttt{nepoch}=3, \texttt{nbatch}=8), clipping $\epsilon=0.2$, value loss coefficient $0.5$, entropy coefficient $0.01$, learning rate $3\times 10^{-4}$, and gradient clipping $0.5$. We terminated training after 80 epochs.

\paragraph{Crafter Environment and Reward}
We train in Crafter, which provides an environment reward at each timestep, as the sum of two components. First, an \emph{achievement} reward gives $+1$ only when an achievement is unlocked for the first time within the current episode (and $0$ otherwise). The only achievement relevant to our benchmarks is \texttt{collect\_wood}, which is given upon the first wood collected by the agent. Second, a \emph{health-shaping} reward provides a small dense signal equal to $-0.1$ for each health point lost and $+0.1$ for each health point regenerated. Beyond these standard rewards, we make the following edits for the two specific tasks:

\texttt{Single Wood Collection}: The only achievement that yields a $+1$ achievement reward is \texttt{collect\_wood}, with all other achievement rewards removed.

\texttt{Greedy Wood Collection}: In addition to removing all non wood-collecting achievement rewards, we add a dense reward of $+1$ for each additional wood collected beyond the first.
 
\subsection{Procgen Benchmark}

For the Procgen~\citep{cobbe2019procgen} benchmark, we followed the data collection procedure of AdaWorld (outlined in Appendix ~\ref{section:procgen}). We generated 500 videos for each task (Climber, Jumper, and Ninja) as training data. We evaluate the trained model on downstream visual planning tasks. For each environment, planning problems are constructed by rolling out random action sequences of length 15 and using the initial frame as the starting observation for the world model and the frame at ($t = 15$) as the target. Planning is performed using a standard MPC-CEM procedure, as described in Appendix~\ref{sec:visual_planning_appendix}.

\subsubsection{Procgen Benchmark Data Generation}
\label{section:procgen}

We followed the same method as AdaWorld~\citep{gao2025adaworld} to collect frame transitions from the following Procgen benchmark tasks: Climber, Jumper, and Ninja. Within each task, we collected 500 videos of an agent acting in the environment, which takes a random action at each step. Each video started from a random scene, and terminated after 1000 environment steps or when the environment returned done. Following AdaWorld, we replaced uniform action sampling with a biased schedule. For a short window we increased the probability of selecting one action, then switched to another. Cycling the favored action this way encouraged each video to cover a broader range of scenes.

\subsection{Real-World Task Details}
\label{section:benchmark_appendix_realworld}

Consistent with prior work on long-horizon manipulation evaluation and recent recommendations for more nuanced robot policy evaluation beyond binary success metrics~\citep{kress2024robot}, we measure performance using a staged progress-based completion metric rather than binary success. A policy receives credit for each sequential stage of a task that it completes, and we report the average progress score across trials.

\begin{figure}[h]
\centering
\begin{subfigure}[b]{0.48\textwidth}
\includegraphics[width=\linewidth]{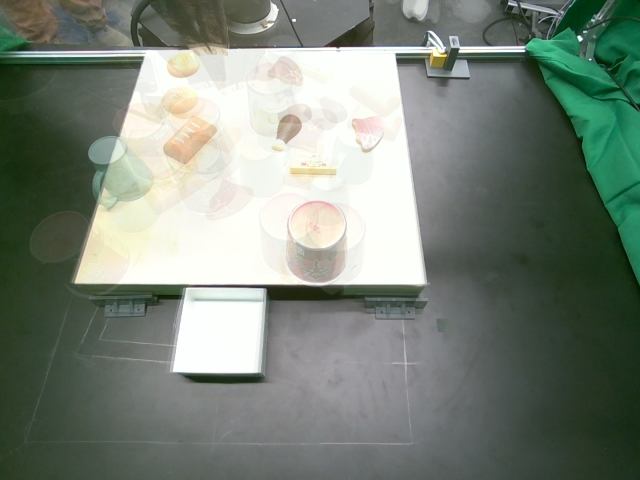}
\caption{Push Ramen}
\label{fig:initial-states-ramen}
\end{subfigure}
\hfill
\begin{subfigure}[b]{0.48\textwidth}
\includegraphics[width=\linewidth]{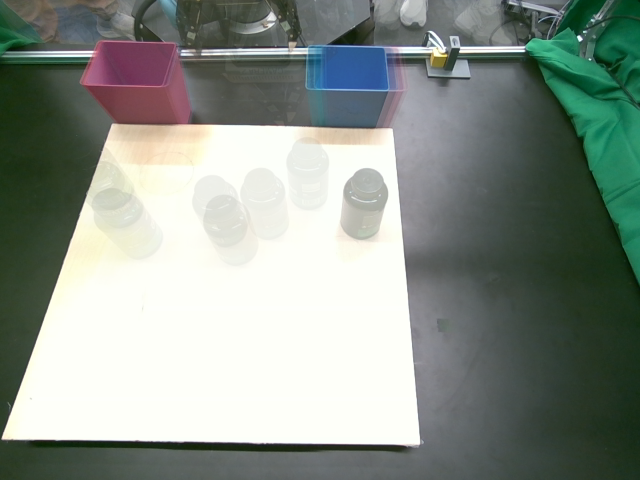}
\caption{Recycle Bottle}
\label{fig:initial-states-bottle}
\end{subfigure}
\caption{Initial states of the real-robot training demonstrations, viewed from \texttt{cam5}. Each panel overlays the first frame of eight randomly sampled successful trajectories, so variation in object and end-effector placement is visible in a single image.}
\label{fig:initial-states}
\end{figure}

\paragraph{Tasks} \texttt{Recycle Bottle} is decomposed into two sequential stages: (1) reaching and grasping the bottle, and (2) lifting it and releasing it into the recycling bin. \texttt{Push Ramen} is likewise decomposed into (1) reaching and making contact with the ramen package, and (2) pushing it to the target box. Scores therefore range from 0 to 2 and are normalized to $[0,1]$ for reporting, so a policy that consistently reaches the object but fails to complete the task scores $0.5$. Each task is evaluated over 20 independent trials. Object positions are randomized across trials as described in Sec.~\ref{sec:benchmarks}. Figure~\ref{fig:initial-states} shows the spread of initial object and end-effector configurations across training demonstrations for both tasks.

\section{Additional Experiments and Results}

\subsection{Analysis of Learned Latent Action}
\label{sec:more_analysis_latent}
\subsubsection{Clustering}
\label{sec:latent_clustering}
\begin{figure}[t]
    \centering
    \includegraphics[width=\linewidth]
    {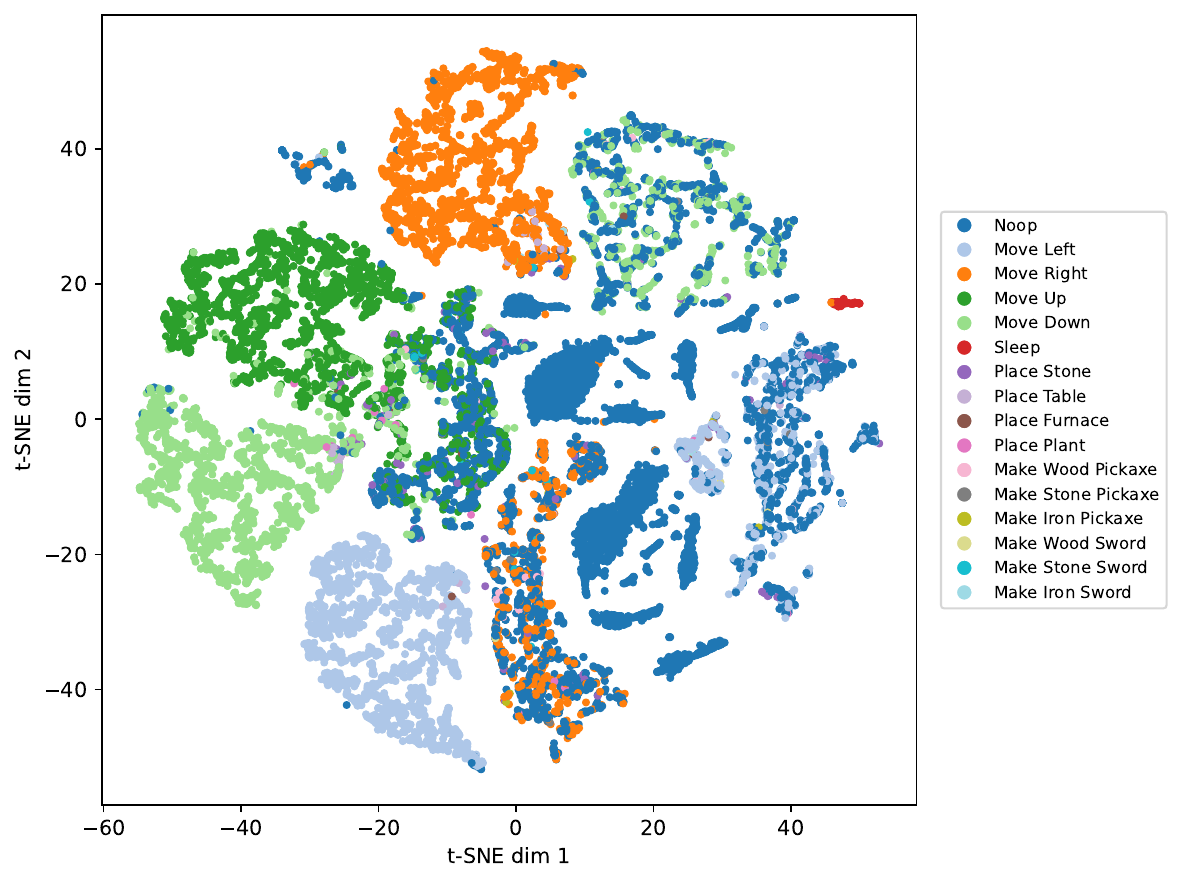}
    \caption{\textbf{Detailed structure of CLAW's continuous latent action
    space.}
    Latent actions extracted from Crafter transitions are projected into two
    dimensions using t-SNE. Each point represents an observation transition
    and is colored by its ground-truth action label for visualization only.
    Although CLAW is trained without action supervision or a discrete codebook,
    related behaviors form coherent regions in the learned continuous latent
    space.}
    \label{fig:cluster}
\end{figure}

To further examine the organization of \alg's continuous latent action space
beyond the comparative UMAP analysis in Sec.~\ref{sec:latent_structure}, we
visualize its structure using an alternative nonlinear projection. We encode
consecutive Crafter observations as
$z_i = \mathrm{LAM}_{\theta}(o_i,o_{i+1})$
and project the resulting latent actions into two dimensions using t-SNE.
Ground-truth Crafter action labels are used only to color the visualization and
are never provided to the latent action model during training.

Figure~\ref{fig:cluster} provides a more detailed view of the structure that
emerges within \alg's latent space. Actions corresponding to related behaviors
occupy coherent regions despite the absence of action supervision or a discrete
codebook. In particular, locomotion actions such as Move Left, Move Right,
Move Up, and Move Down form distinguishable regions, while several object
interaction and placement actions also exhibit coherent organization. The
emergence of action-aligned neighborhoods under a second visualization method
is consistent with the comparative results in Sec.~\ref{sec:latent_structure}
and further suggests that the learned representation captures behaviorally
relevant transition structure.

\paragraph{Legend for the Main-Paper UMAP Figure.}
The shared legend in Figure~\ref{fig:umap_comparison} of the main paper applies to all UMAP panels in that figure, with each colored square indicating the ground-truth Crafter action associated with points of the same color. Labels follow the Crafter GUI key bindings, ordered from left to right: \texttt{0} denotes no-op (no dedicated key); $\leftarrow$/$\rightarrow$/$\uparrow$/$\downarrow$ denote movement left/right/up/down (WASD in the GUI); \texttt{Tab} denotes sleep; \texttt{R}/\texttt{T}/\texttt{F}/\texttt{P} denote placing stone, table, furnace, and plant; and \texttt{1}--\texttt{6} denote crafting wood, stone, and iron pickaxes followed by wood, stone, and iron swords. The \texttt{Do} action is omitted from this visualization. Ground-truth action labels are used only for coloring the UMAP embeddings and are never provided to the models during latent-action learning.

\subsubsection{Pseudo-Noop Transitions and Off-Centroid Ground-Truth Noops}
\label{app:pseudo-noop}

The action-colored latent-space visualizations in
Sec.~\ref{sec:latent_structure} and Sec.~\ref{sec:latent_clustering} use
ground-truth Crafter actions as an external reference for interpreting the
learned representation. However, these labels record the command issued by the
policy and need not correspond one-to-one with observable transitions. We
therefore examine two complementary sources of action--observation ambiguity
within the dominant confusion class: non-noop commands that produce negligible
visual change, and ground-truth no-op commands whose resulting transitions
exhibit substantial visual change.

\paragraph{Notation.}
Let a trajectory contain observations $\{x_t\}_{t=0}^{T-1}$ and discrete
actions $\{a_t\}_{t=0}^{T-2}$, where $a_t$ is applied between $x_t$ and
$x_{t+1}$. We denote the environment's dedicated no-op action by $a^\star$.
For the Crafter analyses below, we use the same action vocabulary as the
clustering visualizations.

CLAW encodes each observation transition as
\[
    z_t = f_\theta(x_t,x_{t+1}) \in \mathbb{S}^{d-1},
\]
where $d=32$ and the latent representation is normalized to the unit sphere.

\paragraph{Pseudo-noop transitions.}
We call a transition $(x_t,x_{t+1},a_t)$ a \emph{pseudo-noop} when the
ground-truth action is non-noop but the resulting visual change is negligible:
\begin{equation}
    a_t \neq a^\star
    \qquad\text{and}\qquad
    \Delta_{\mathrm{pix}}(x_t,x_{t+1}) \leq \varepsilon,
\end{equation}
where
\begin{equation}
    \Delta_{\mathrm{pix}}(x_t,x_{t+1})
    =
    \frac{1}{CHW}
    \|x_{t+1}-x_t\|_1.
\end{equation}
We normalize frames to $[0,1]$ and use $\varepsilon=10^{-3}$.

Pseudo-noops are not necessarily annotation errors. The recorded action
correctly specifies the command issued by the human demonstrator, while the
resulting observation transition may nevertheless be effectively idle. In
Crafter, these cases most commonly arise from blocked locomotion, with
unsuccessful place and craft commands occurring less frequently.

\paragraph{Prevalence in Crafter.}
We analyze approximately 34k transitions from a Crafter training subset using
the same action vocabulary as the clustering analysis. Among transitions
satisfying $\Delta_{\mathrm{pix}}\leq10^{-3}$, approximately 78\% are
ground-truth \texttt{Noop}, while approximately 22\% carry a non-noop command
and therefore constitute pseudo-noops. Thus, although visually static
transitions are strongly enriched for true no-ops, a non-negligible fraction
correspond to other issued actions.

\begin{figure}[t]
    \centering
    \includegraphics[width=0.92\linewidth]
    {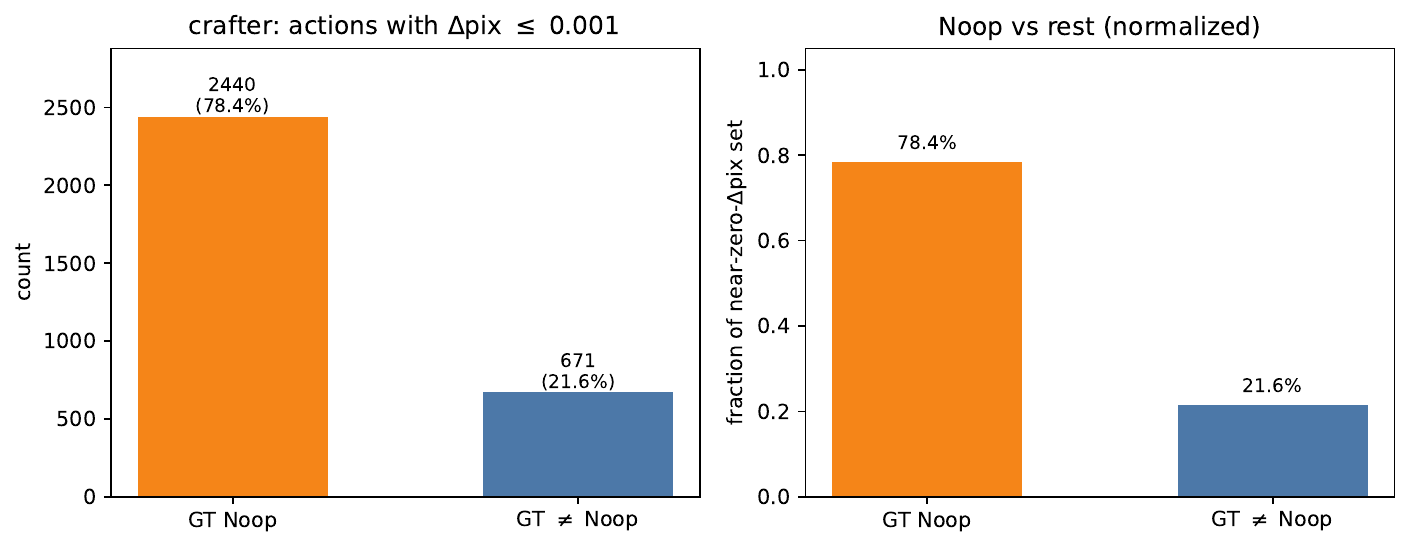}
    \caption{\textbf{True no-op and pseudo-noop transitions in Crafter.}
    Distribution of ground-truth \texttt{Noop} and non-noop actions among
    transitions with near-zero visual change
    ($\Delta_{\mathrm{pix}}\leq10^{-3}$). The analysis uses the same action
    vocabulary as the latent-space clustering visualizations, with
    \texttt{Do} excluded.}
    \label{fig:static-noop-vs-rest}
\end{figure}

Figure~\ref{fig:static-noop-vs-rest} summarizes this distribution.

\paragraph{Off-centroid ground-truth noops.}
The complementary case also occurs: a transition can carry a ground-truth
\texttt{Noop} label while exhibiting non-negligible visual change. To examine
the structure of these transitions in CLAW's latent space, we define an
empirical no-op centroid. Let $\mathcal{D}^{\star}$ denote the set of CLAW
latent representations associated with ground-truth \texttt{Noop}
transitions. We compute
\begin{equation}
    \bar{z}^{\star}
    =
    \frac{1}{|\mathcal{D}^{\star}|}
    \sum_{z \in \mathcal{D}^{\star}} z,
    \qquad
    c^{\star}
    =
    \frac{\bar{z}^{\star}}
    {\|\bar{z}^{\star}\|_2},
\end{equation}
and measure the distance of a latent transition to this reference using
\begin{equation}
    d(z,c^{\star})
    =
    \|z-c^{\star}\|_2.
\end{equation}

We refer to ground-truth \texttt{Noop} transitions with large distance from
$c^{\star}$ as \emph{off-centroid ground-truth noops}. These examples probe
heterogeneity within the labeled idle class rather than disagreement with the
ground-truth action label itself. The no-op centroid should therefore be
interpreted as a prototype of common no-op transitions, rather than as a
representation of every transition carrying the discrete no-op command.

\begin{figure*}[t]
    \centering
    \setlength{\tabcolsep}{8pt}
    \begin{tabular}{@{}cc@{}}
        \multicolumn{2}{c}{\textbf{Pseudo-noop transitions}}
        \\[0.4em]

        \includegraphics[width=0.46\textwidth]
        {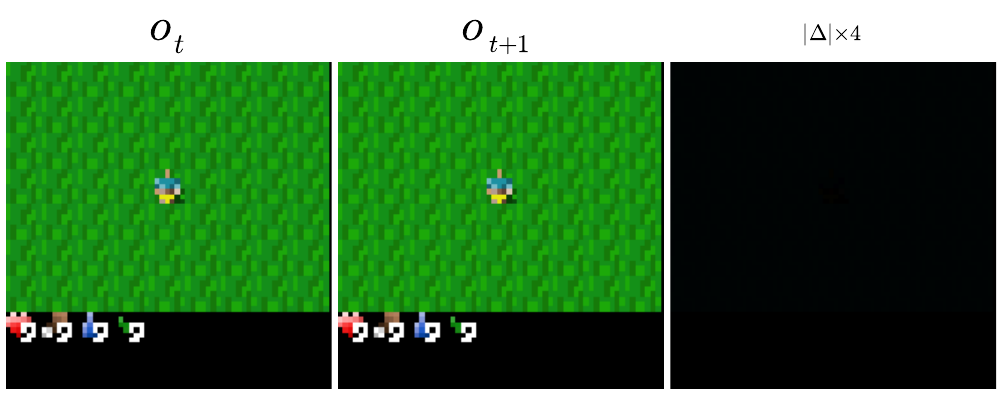}
        &
        \includegraphics[width=0.46\textwidth]
        {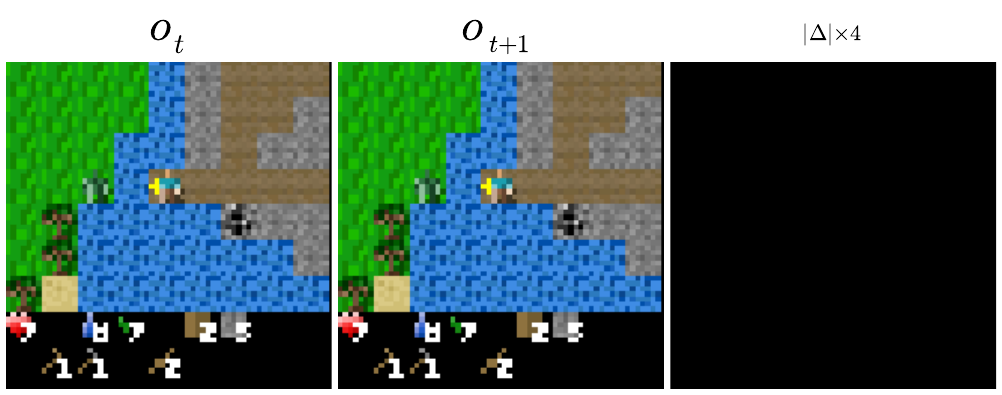}
        \\[0.9em]

        \multicolumn{2}{c}{\textbf{Off-centroid ground-truth Noop transitions}}
        \\[0.4em]

        \includegraphics[width=0.46\textwidth]
        {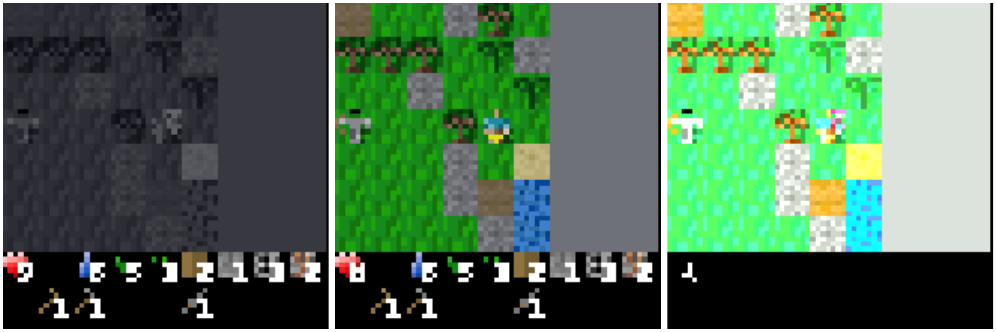}
        &
        \includegraphics[width=0.46\textwidth]
        {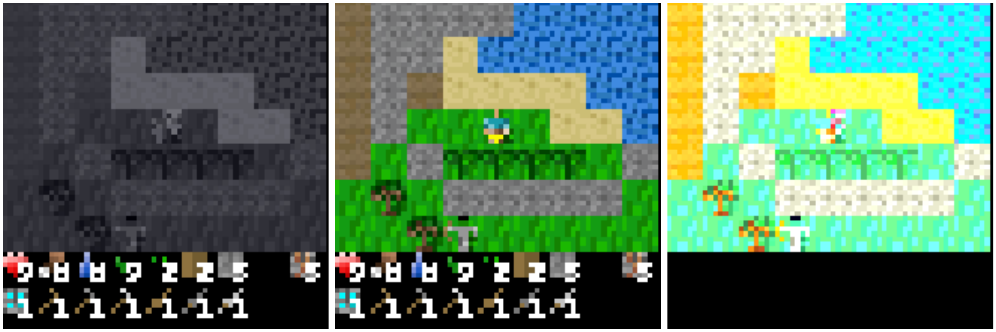}
    \end{tabular}

    \caption{\textbf{Two forms of action--observation mismatch in Crafter.}
    Top: pseudo-noop examples corresponding to \texttt{Move Down} and
    \texttt{Move Left}, where non-noop movement commands produce little or no
    observable change. Bottom: two ground-truth \texttt{Noop} transitions whose
    CLAW representations lie far from the empirical no-op centroid and exhibit
    appreciable visual change despite the idle command. Each example shows
    consecutive observations together with their visual difference.}
    \label{fig:pseudo-noop-examples}
\end{figure*}

Qualitatively, ground-truth \texttt{Noop} transitions far from the centroid
often exhibit appreciable visual changes, including changes in the
on-screen status display or other scene dynamics despite the idle command.
Conversely, pseudo-noop transitions can exhibit almost no visible change even
though the issued command is non-noop.
Figure~\ref{fig:pseudo-noop-examples} illustrates these complementary cases.

\paragraph{Implications for latent-action evaluation.}
These observations highlight an important limitation of interpreting observation-based latent actions solely through discrete ground-truth action labels. Pseudo-noops show that a non-noop command can induce an observation transition that is effectively indistinguishable from idleness, while off-centroid ground-truth noops show that an idle command can coexist with substantial visual change. Thus, the discrete action label and the observed transition effect do not define identical partitions of the data.

This distinction is particularly relevant when interpreting action-colored clustering visualizations. Ground-truth actions remain a useful external reference for assessing whether action-related structure emerges, but they should not be treated as a perfect semantic partition of visual transitions. Some apparent mixing can therefore reflect genuine ambiguity in the action--observation relationship rather than representational failure.

We accordingly interpret the clustering analyses together with nearest-neighbor retrieval, action transfer, and latent perturbation experiments. These complementary evaluations probe whether the learned representations capture consistent behavioral effects across visual contexts and whether their local geometry remains useful for downstream policy learning and planning.

\subsubsection{Latent Action Policy Robustness to Latent Perturbations}
\label{sec:latent_perturbation}

\begin{figure}
    \centering
    \includegraphics[width=\linewidth]{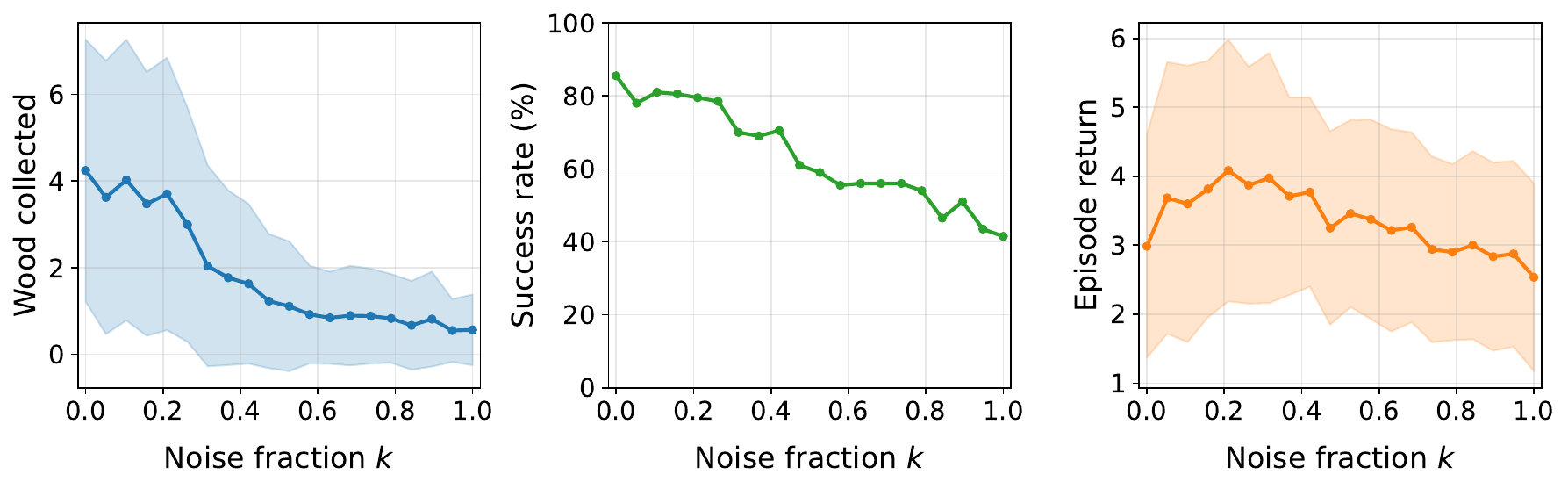}
    \caption{
\textbf{Robustness of latent-action policies to latent perturbations.}
Performance of the \texttt{Greedy Wood Collection} policy as Gaussian noise is interpolated into the predicted latent action (Eq.~\ref{eq:latent-noise}). Left: average wood collected. Center: success rate. Right: cumulative environment return. Shaded regions indicate one standard deviation over 200 evaluation episodes. Performance decreases smoothly as latent actions move away from the learned action, suggesting that neighboring latent actions correspond to similar behaviors and that policy performance depends on the semantic structure of the learned latent space.
}
    \label{fig:noised-latent-eval}
\end{figure}

A central premise of our method is that its learned latent action space captures meaningful and controllable environment dynamics. To evaluate whether downstream policies rely on this learned structure, we study the effect of perturbing latent actions during execution. Specifically, we consider the \texttt{Greedy Wood Collection} policy trained using latent action supervision and progressively corrupt its predicted latent actions with Gaussian noise.

At each environment step, the policy network maps the current observation to a latent action embedding $\mathbf{z}_{\pi}\in\mathbb{R}^{32}$. We generate a perturbed latent action by interpolating between the policy prediction and a randomly sampled latent vector:
\begin{equation}
    \mathbf{z}=(1-k)\mathbf{z}_{\pi}+k\boldsymbol{\epsilon},
    \qquad
    \boldsymbol{\epsilon}\sim\mathcal{N}(\mathbf{0},\mathbf{I}),
    \label{eq:latent-noise}
\end{equation}
where $k\in[0,1]$ controls the perturbation magnitude. When $k=0$, execution uses the policy's predicted latent action directly, while $k=1$ corresponds to entirely random latent actions.

Following the policy evaluation protocol described in Section~\ref{sec:latent_policy}, perturbed latent actions are converted to executable Crafter actions via nearest-neighbor retrieval in the latent action dataset. We evaluate policies in \texttt{CrafterReward-v1} across 200 episodes for each noise level, sweeping $k$ uniformly between $0$ and $1$. We report three metrics: (i) the average amount of wood collected per episode, (ii) the success rate, defined as the fraction of episodes that collect at least one unit of wood, and (iii) the cumulative environment return. Shaded regions denote one standard deviation across evaluation episodes.

Figure~\ref{fig:noised-latent-eval} shows that performance degrades smoothly as latent actions are increasingly corrupted. The policy maintains strong performance under small perturbations, but progressively loses effectiveness as latent actions move away from the policy predictions. At the extreme case of pure noise ($k=1$), both task success and reward approach random-behavior levels. These results suggest that successful control depends on the semantic structure encoded within \alg's latent action space rather than on the retrieval mechanism alone. The smooth degradation in performance suggests that useful task information is distributed continuously throughout the latent space rather than being concentrated in a small set of isolated latent codes.

\subsubsection{Structure of the Learned Latent Actions}
The clustering analyses in Sec.~\ref{sec:latent_structure} and Appendix~\ref{sec:more_analysis_latent} use ground-truth actions as an external reference, but that reference is imperfect. In Crafter the recorded action is the command the demonstrator issued, not the transition it produced. Different commands can induce nearly indistinguishable transitions, and \texttt{Noop} transitions can themselves contain appreciable visual change. Appendix~\ref{app:pseudo-noop} quantifies both cases. Action labels are therefore a useful reference for latent structure without defining a perfect partition of observable transition effects, and some overlap between action categories reflects ambiguity in the action--observation relationship rather than representational failure.

The remaining analyses avoid this dependence by probing behavioral consistency directly. Nearest-neighbor retrieval (Sec.~\ref{sec:closest_action_retrieval}, with further examples in Appendix~\ref{sec:retrieval_appendix}) shows that latent proximity tracks similar transitions across visually unrelated scenes. Perturbation (Appendix~\ref{sec:latent_perturbation}) shows performance degrading smoothly as the latent is corrupted, so task-relevant information is distributed across the space rather than concentrated in isolated codes. Action transfer (Sec.~\ref{sec:action_transfer}, Appendix~\ref{section:action_transfer_supp}) shows latents extracted from one trajectory inducing corresponding behavior in a different visual context. None of these compares against a discrete label.

\subsection{Retrieval Task}
\label{sec:retrieval_appendix}
\begin{figure}[h]
    \centering
    \includegraphics[width=0.8\linewidth]{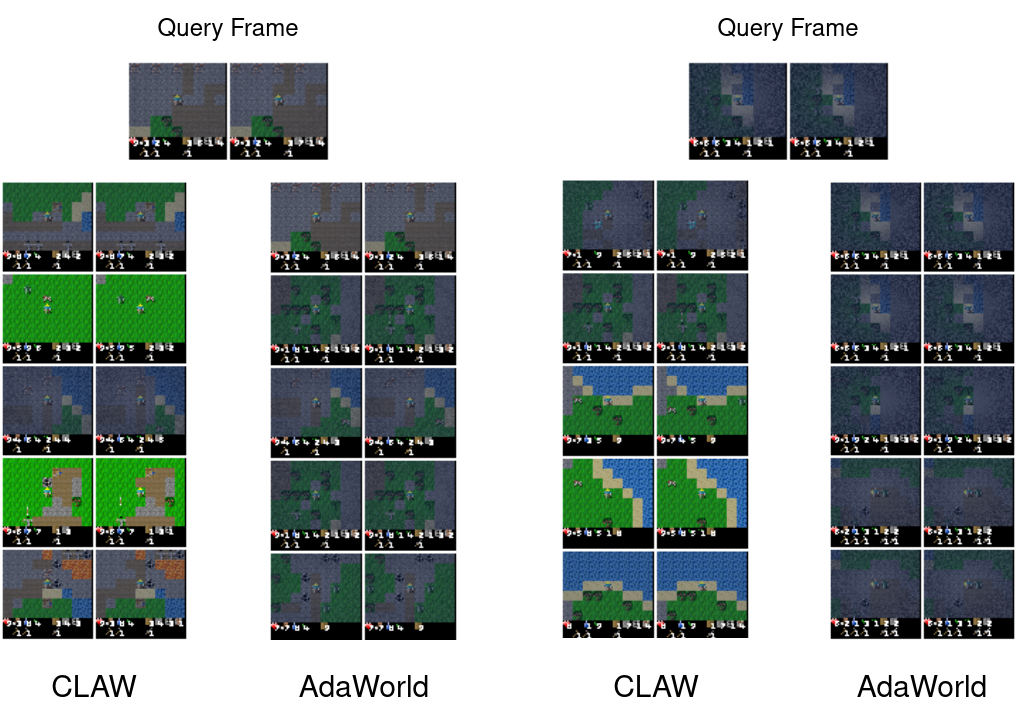}
    \caption{Nearest action retrieval comparison between \alg and AdaWorld. For each query transition, we retrieve the closest latent actions from the latent action embeddings of the Crafter dataset and visualize their corresponding frame pairs. \alg retrieves a more diverse set of transitions that accurately reflect the underlying action semantics, with reduced sensitivity to color and texture. In contrast, AdaWorld predominantly retrieves transitions from visually similar scenes, suggesting a stronger dependence on appearance cues rather than action-specific dynamics.}
    \label{fig:action_retrieval_crafter}
\end{figure}

As shown in Figure~\ref{fig:action_retrieval_crafter}, our model retrieves a more diverse set of transitions that correspond to the correct actions, while exhibiting reduced reliance on superficial visual attributes such as color or texture. In contrast, AdaWorld’s Latent Action Model frequently retrieves transitions from visually similar scenes, revealing a bias toward scene-specific appearance cues rather than action semantics.

In Figure~\ref{fig:action_retrieval_crafter}, the query corresponds to a \texttt{DO} action where the agent interacts with an object in the environment. \alg retrieves transitions that capture similar object interactions, whereas AdaWorld mainly retrieves transitions from visually similar scenes with limited or no interaction. On the right side of the figure, both \alg and AdaWorld recover the correct action category; however, \alg shows substantially less reliance on scene color and retrieves transitions from a more diverse set of environments.

\begin{figure}[htbp]
    \centering
    \begin{subfigure}{0.3\textwidth}
        \centering
        \includegraphics[width=\linewidth]{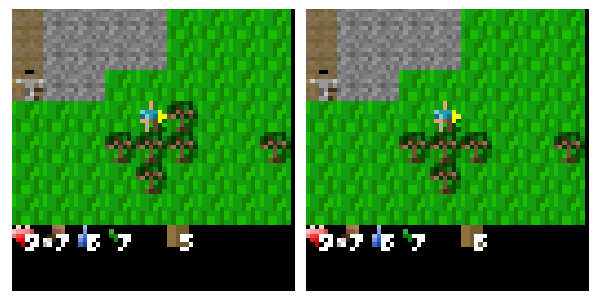}
    \end{subfigure}
    \\[0.1cm]
    \begin{subfigure}{0.2\textwidth}
        \centering
        \includegraphics[width=\linewidth]{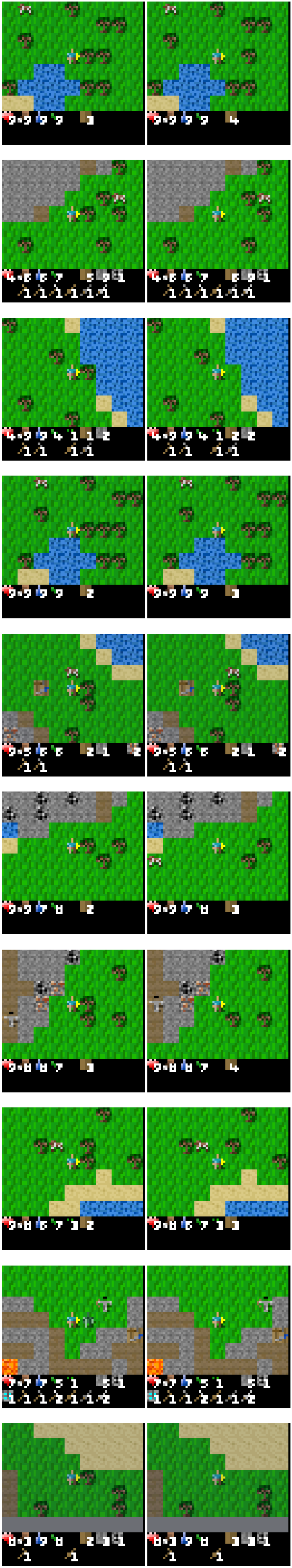}
        \caption{\alg}
    \end{subfigure}
    \hspace{0.02\textwidth}
    \begin{subfigure}{0.2\textwidth}
        \centering
        \includegraphics[width=\linewidth]{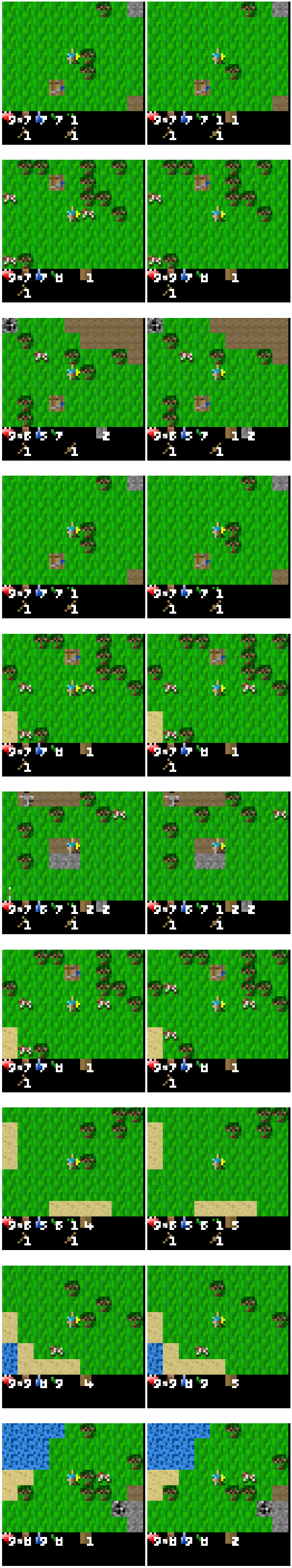}
        \caption{AdaWorld}
    \end{subfigure}
    \hspace{0.02\textwidth}
    \begin{subfigure}{0.2\textwidth}
        \centering
        \includegraphics[width=\linewidth]{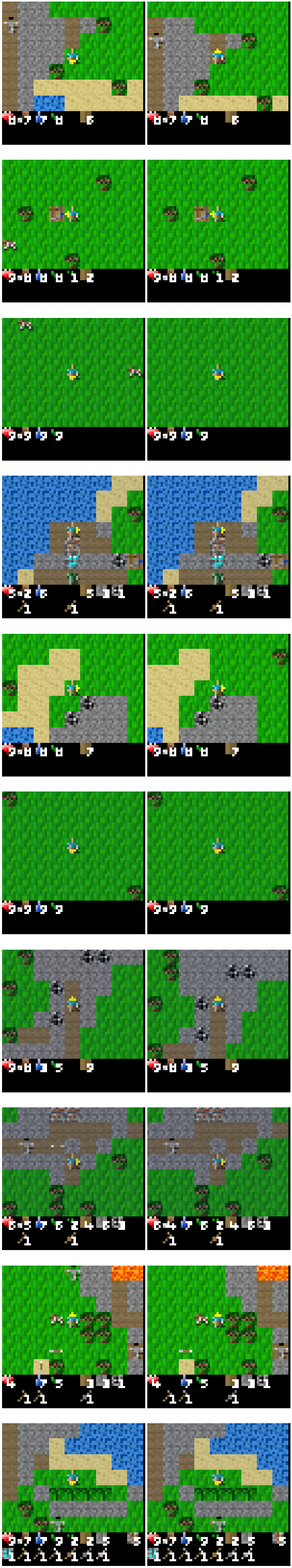}
        \caption{LAPO}
    \end{subfigure}
    \hspace{0.02\textwidth}
    \begin{subfigure}{0.2\textwidth}
        \centering
        \includegraphics[width=\linewidth]{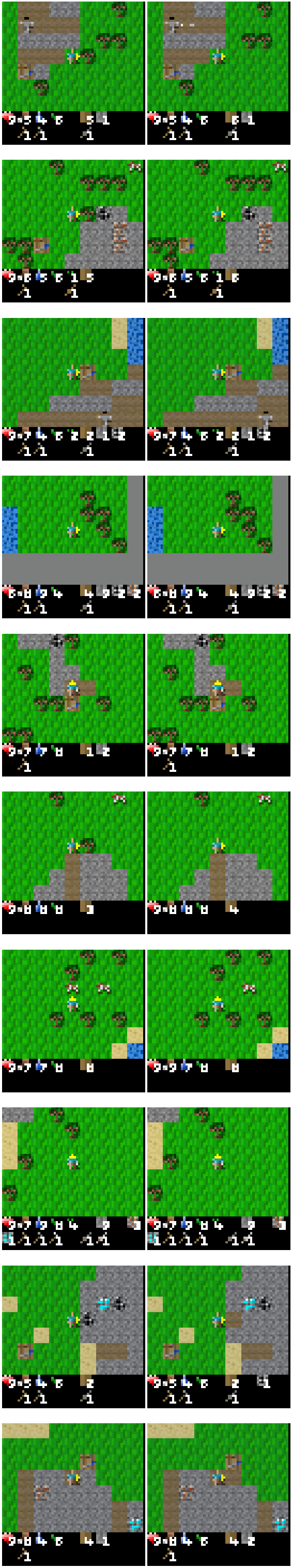}
        \caption{CoMo}
    \end{subfigure}
    \caption{
        \textbf{Closest Latent Action retrieval via $L_2$ nearest-neighbor:}
        (Top) query transition.
        (Bottom) top-10 neighbors retrieved via $L_2$ search in each method's latent space (\alg, AdaWorld, LAPO, CoMo). Retrieval is independent per encoder.
    }
    \label{fig:action_retrieval_crafter_supp}
\end{figure}

Additionally, Figure~\ref{fig:action_retrieval_crafter_supp} presents a query transition corresponding to a \texttt{chopping tree} action. The nearest neighbors retrieved by \alg consistently correspond to the same chopping behavior. Although AdaWorld also retrieves several chopping tree examples, it additionally includes multiple \texttt{no-op} actions. In contrast, LAPO retrieves a heterogeneous mixture of actions involving different motions and behaviors. These results indicate that \alg learns action representations with stronger semantic clustering compared to the baselines.

\begin{figure}[htbp]
    \centering
    \begin{subfigure}{0.3\textwidth}
        \centering
        \includegraphics[width=\linewidth]{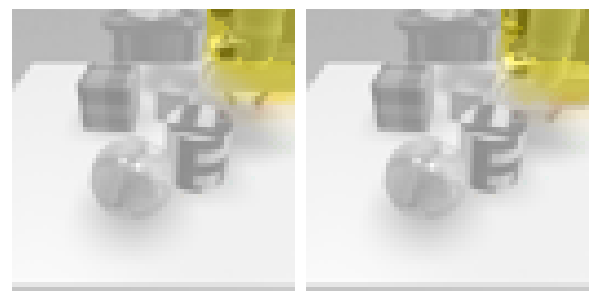}
    \end{subfigure}
    \\[0.1cm]
    \begin{subfigure}{0.2\textwidth}
        \centering
        \includegraphics[width=\linewidth]{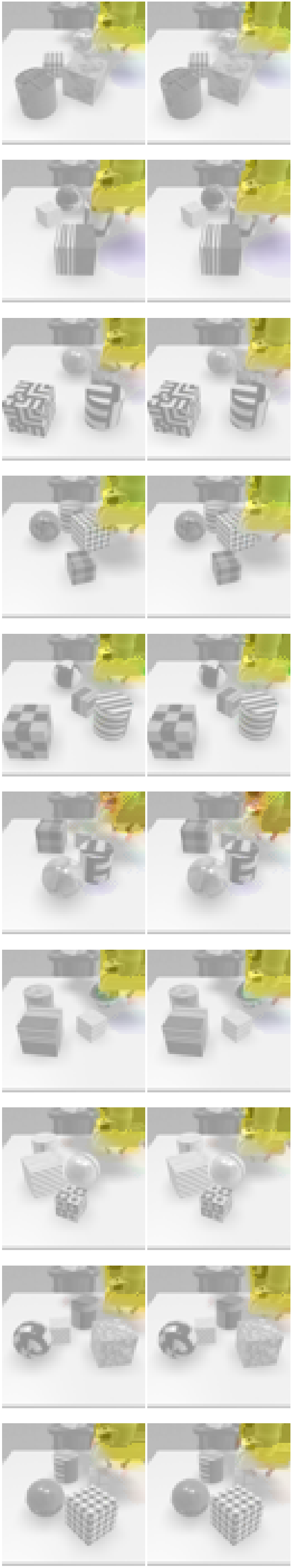}
        \caption{CLAW}
    \end{subfigure}
    \hspace{0.05\textwidth}
    \begin{subfigure}{0.2\textwidth}
        \centering
        \includegraphics[width=\linewidth]{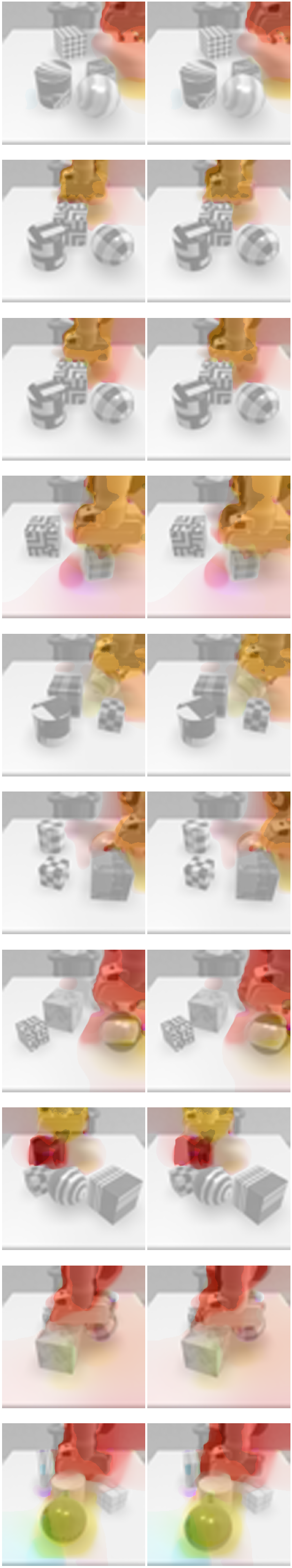}
        \caption{AdaWorld}
    \end{subfigure}
    \hspace{0.05\textwidth}
    \begin{subfigure}{0.2\textwidth}
        \centering
        \includegraphics[width=\linewidth]{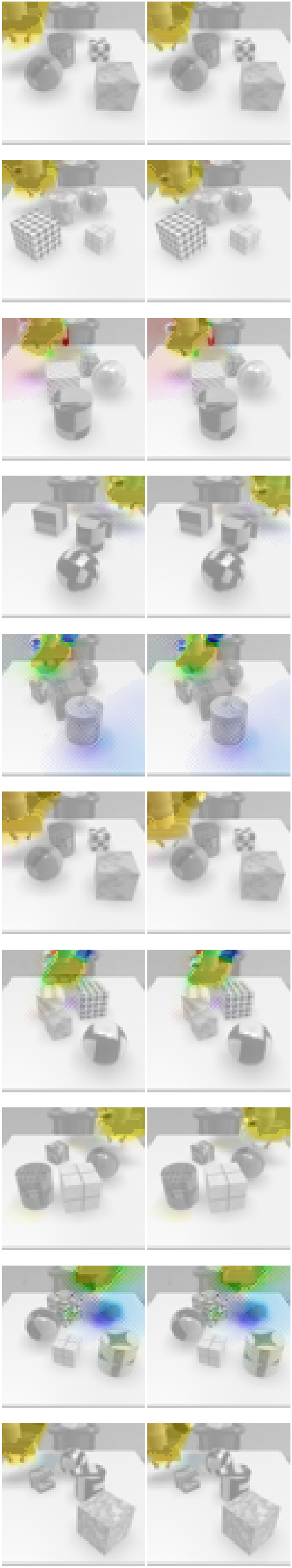}
        \caption{LAPO}
    \end{subfigure}
    \hspace{0.05\textwidth}
    \begin{subfigure}{0.2\textwidth}
        \centering
        \includegraphics[width=\linewidth]{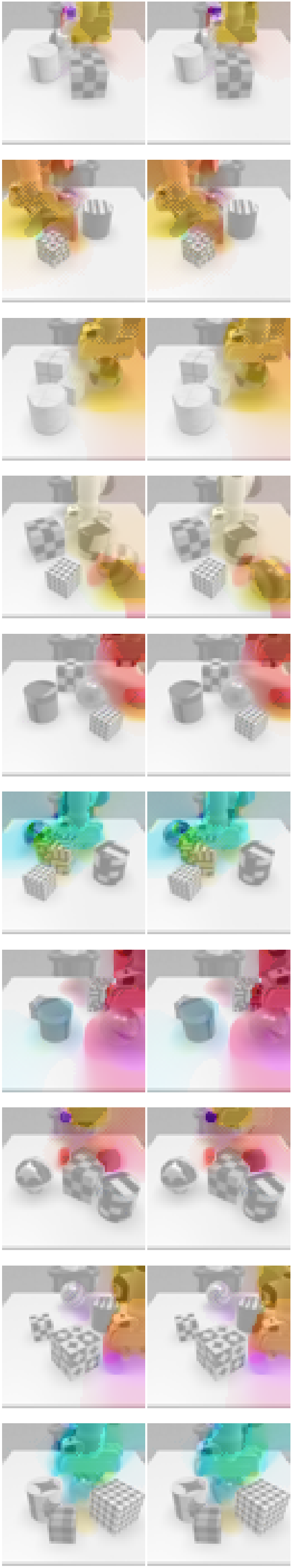}
        \caption{CoMo}
    \end{subfigure}
    \caption{
        \textbf{Closest Latent Action retrieval via $L_2$ nearest-neighbor:}
        (Top) query transition.
        (Bottom) top-10 neighbors retrieved via $L_2$ search in each method's latent space (\alg, AdaWorld, LAPO, CoMo). Retrieval is independent per encoder. All frames shown as grayscale with optical flow ($t\!\to\!t{+}1$) overlaid with Middlebury color wheel (Fig.~\ref{fig:middleburycolorwheel}).
    }
    \label{fig:action_retrieval_robosuite_supp}
\end{figure}

\begin{figure}[htbp]
    \centering
    \begin{subfigure}{0.3\textwidth}
        \centering
        \includegraphics[width=\linewidth]{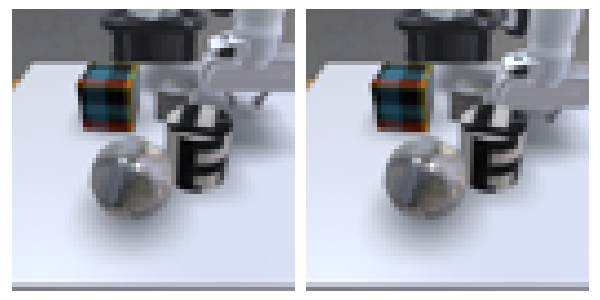}
    \end{subfigure}
    \\[0.1cm]
    \begin{subfigure}{0.2\textwidth}
        \centering
        \includegraphics[width=\linewidth]{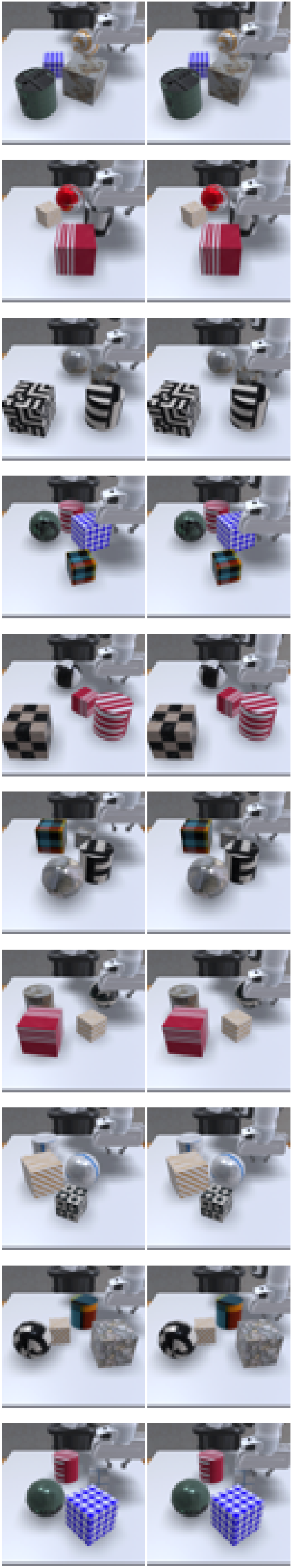}
        \caption{CLAW}
    \end{subfigure}
    \hspace{0.02\textwidth}
    \begin{subfigure}{0.2\textwidth}
        \centering
        \includegraphics[width=\linewidth]{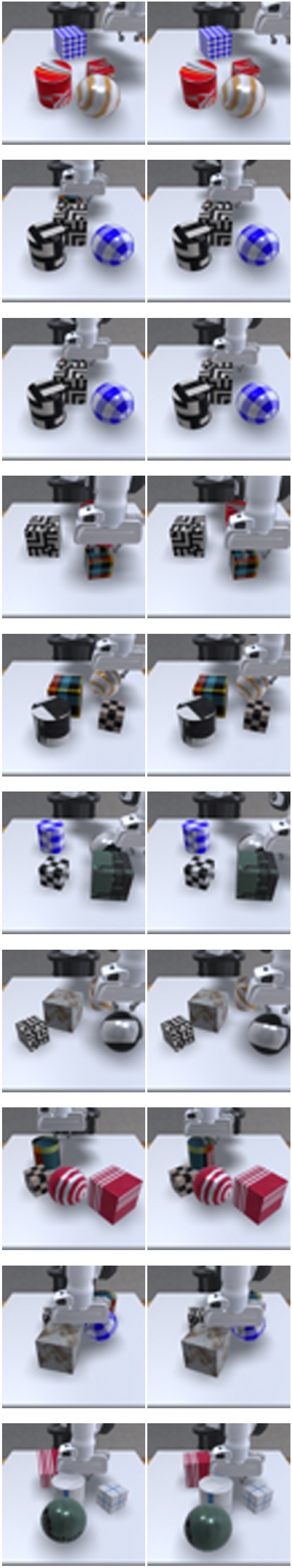}
        \caption{AdaWorld}
    \end{subfigure}
    \hspace{0.02\textwidth}
    \begin{subfigure}{0.2\textwidth}
        \centering
        \includegraphics[width=\linewidth]{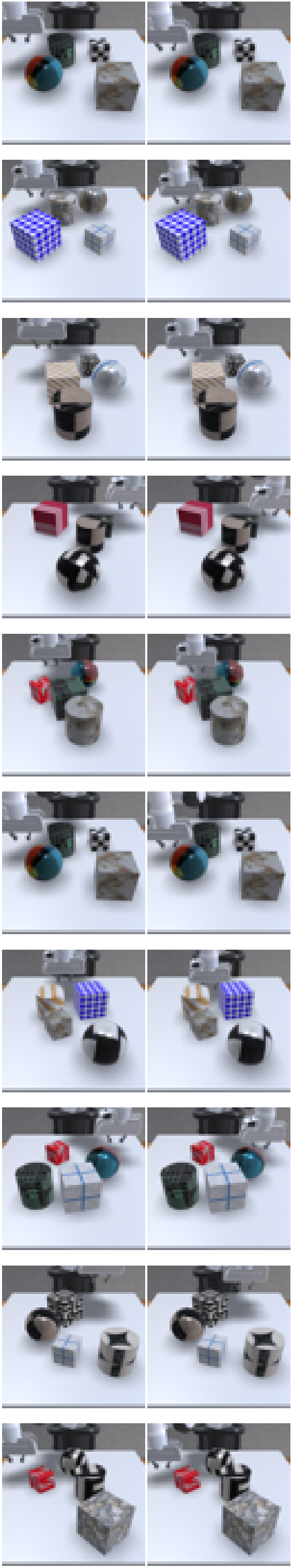}
        \caption{LAPO}
    \end{subfigure}
    \hspace{0.02\textwidth}
    \begin{subfigure}{0.2\textwidth}
        \centering
        \includegraphics[width=\linewidth]{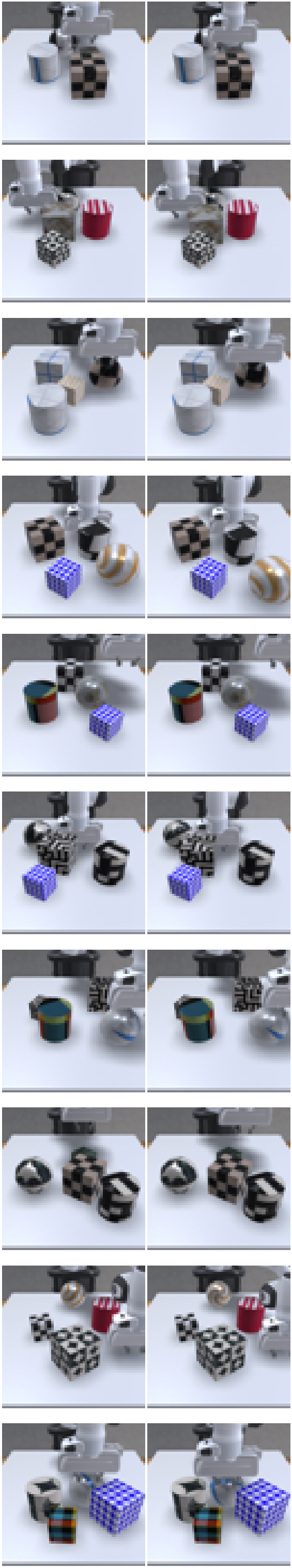}
        \caption{CoMo}
    \end{subfigure}
    
    \caption{
        \textbf{RGB Visualization of Closest Latent Action retrieval via $L_2$ nearest-neighbor:}
        (Top) query transition.
        (Bottom) top-10 neighbors retrieved via $L_2$ search in each method's latent space (\alg, AdaWorld, LAPO, CoMo). Retrieval is independent per encoder. All frames shown as RGB ($t\!\to\!t{+}1$).
    }
    \label{fig:action_retrieval_robosuite_2_supp}
\end{figure}

\begin{figure}[htbp]
    \centering
    \begin{subfigure}{0.3\textwidth}
        \centering
        \includegraphics[width=\linewidth]{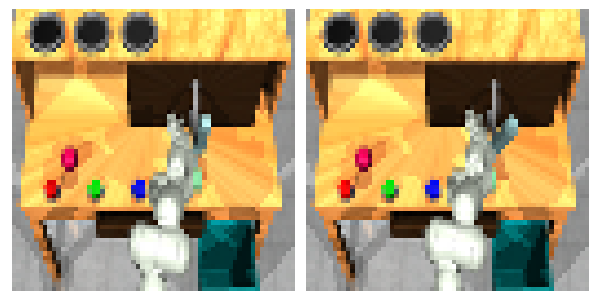}
    \end{subfigure}
    \\[0.1cm]
    \begin{subfigure}{0.2\textwidth}
        \centering
        \includegraphics[width=\linewidth]{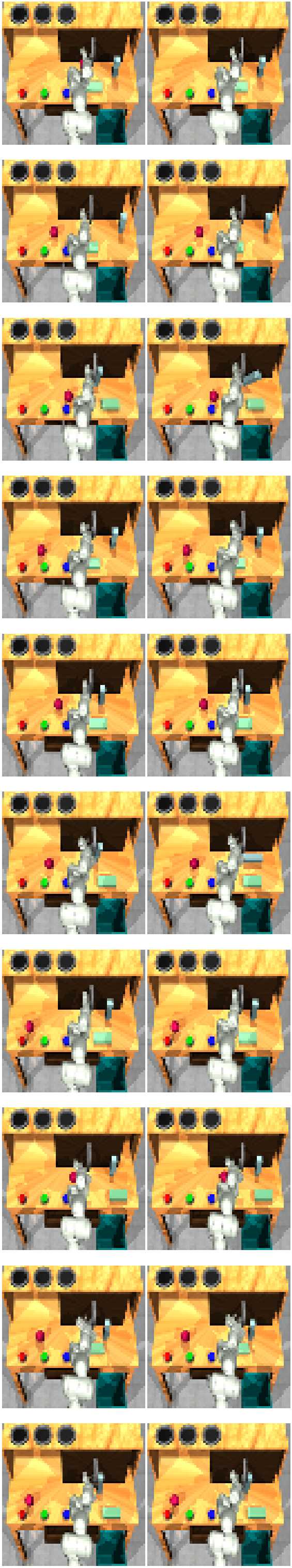}
        \caption{\alg}
    \end{subfigure}
    \hspace{0.02\textwidth}
    \begin{subfigure}{0.2\textwidth}
        \centering
        \includegraphics[width=\linewidth]{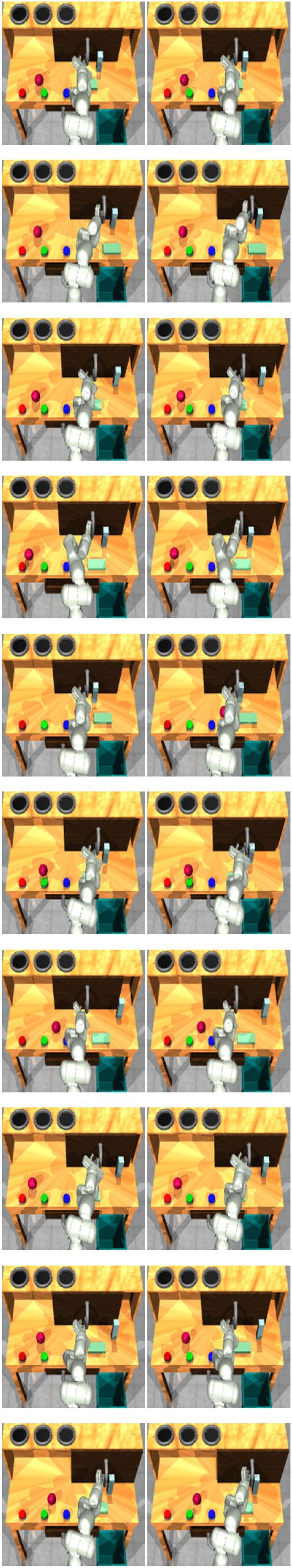}
        \caption{AdaWorld}
    \end{subfigure}
    \hspace{0.02\textwidth}
    \begin{subfigure}{0.2\textwidth}
        \centering
        \includegraphics[width=\linewidth]{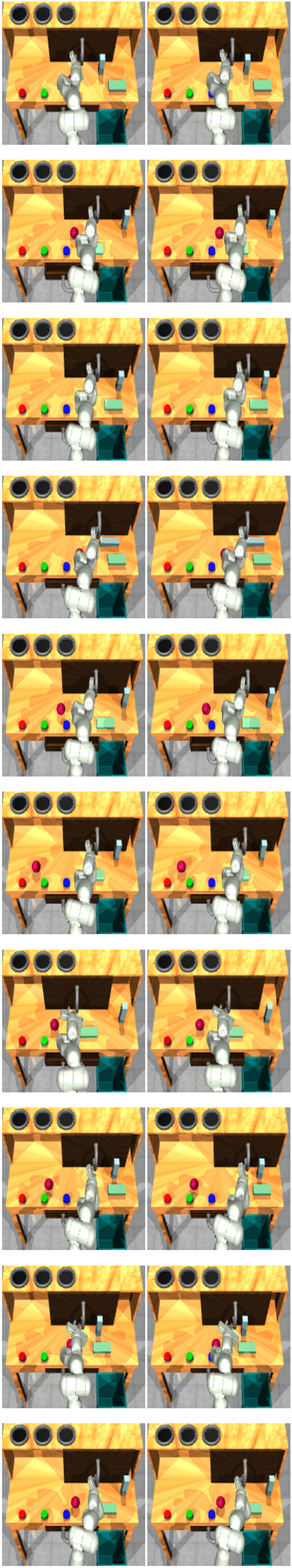}
        \caption{LAPO}
    \end{subfigure}
    \hspace{0.02\textwidth}
    \begin{subfigure}{0.2\textwidth}
        \centering
        \includegraphics[width=\linewidth]{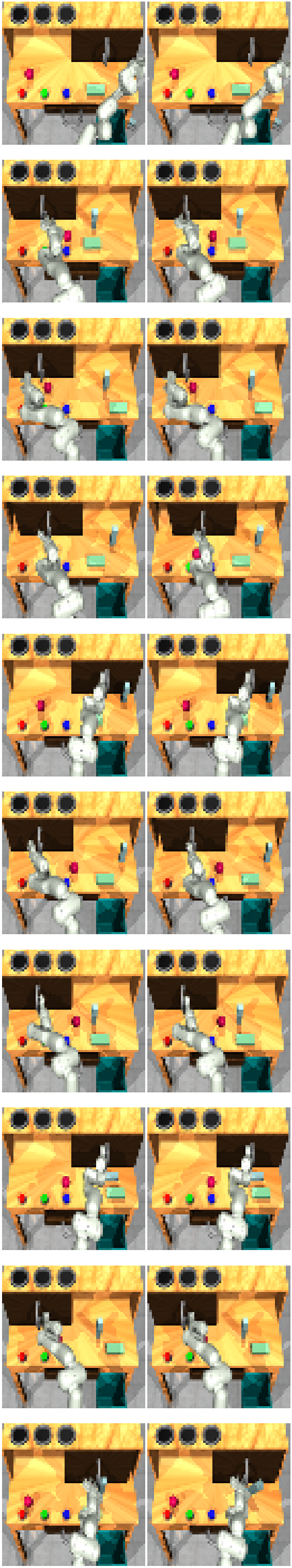}
        \caption{CoMo}
    \end{subfigure}
    \caption{
        \textbf{Closest Latent Action retrieval via $L_2$ nearest-neighbor:}
        (Top) query transition.
        (Bottom) top-10 neighbors retrieved via $L_2$ search in each method's latent space (\alg, AdaWorld, LAPO, CoMo). Retrieval is independent per encoder.
    }
    \label{fig:action_retrieval_robosuite_3_supp}
\end{figure}

Additional 10 closest action retrieval examples for different query transitions, along with a broader set of transitions, are shown in Figures~\ref{fig:action_retrieval_robosuite_supp}, ~\ref{fig:action_retrieval_robosuite_2_supp}, and~\ref{fig:action_retrieval_robosuite_3_supp}. Overall, these results demonstrate that \alg’s latent action representations are robust and transferable, enabling more reliable planning and control in downstream visual tasks.

\begin{figure}
    \centering
    \includegraphics[width=0.3\linewidth]{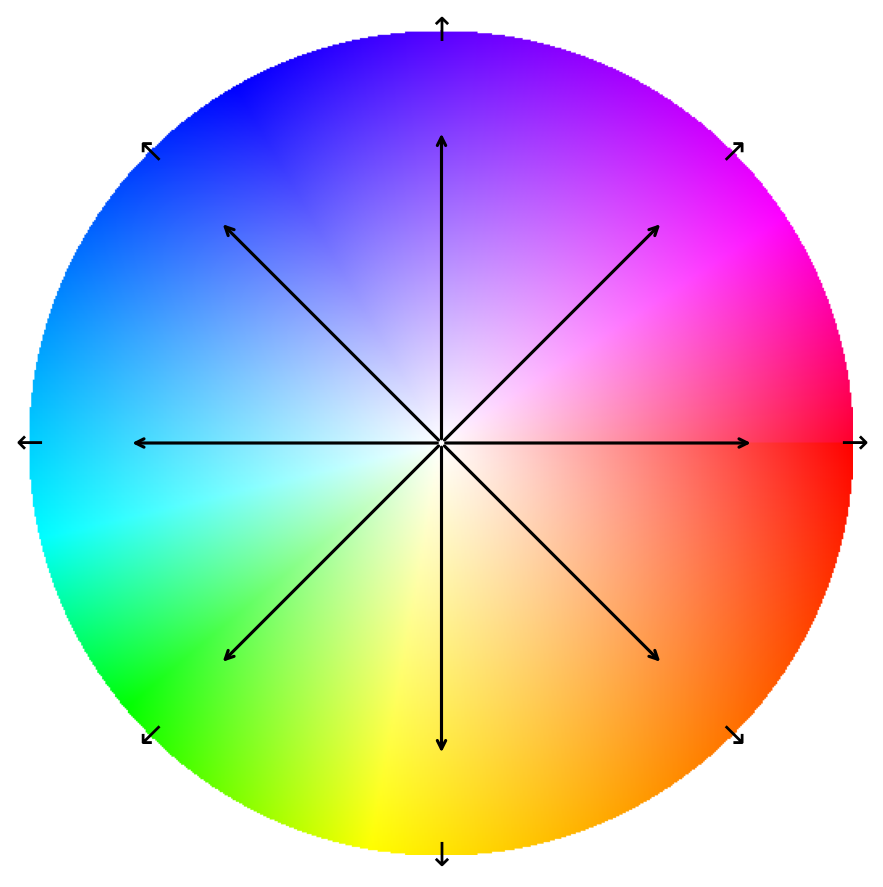}
    \caption{Middlebury Color Wheel.}
    \label{fig:middleburycolorwheel}
\end{figure}

\subsection{Analysis of Learned World Model}
\subsubsection{Bounded Latent Action Leakage Metric}
\label{app:leakage_metric}

We evaluate whether a learned latent action captures the underlying action
independently of the visual context from which it was inferred. To control for
action identity, we construct groups of transitions that share the same
ground-truth semantic action:
\begin{equation}
    \mathcal{G}
    =
    \left\{
    (I_i^j,a,I_{i+1}^j)
    \right\}_{j=1}^{n}.
\end{equation}
In our experiments, we randomly sample 200 same-action groups per setting, each containing $n=8$ transitions.

For each transition $j$, the latent action encoder extracts
\begin{equation}
    z^j = E(I_i^j,I_{i+1}^j).
\end{equation}
Because all transitions in $\mathcal{G}$ correspond to the same semantic
action, an ideal action representation should remain effective when transferred
between contexts in the group.

We therefore apply each source latent $z^j$ to every target context $k$ using
the learned world model:
\begin{equation}
    \hat I_{i+1}^{j\rightarrow k}
    =
    W(I_i^k,z^j).
\end{equation}
The case $j=k$ corresponds to matched-context prediction, while $j\neq k$
tests cross-context transfer.

We define the image-level prediction error
\begin{equation}
    M_{j,k}
    =
    \operatorname{MSE}
    \left(
    W(I_i^k,z^j),
    I_{i+1}^k
    \right).
\end{equation}
For each cross-context transfer, we then define the bounded pairwise leakage
score
\begin{equation}
\label{eq:pairwise_leakage}
    D_{j,k}
    =
    \frac{
    M_{j,k}-M_{j,j}
    }{
    M_{j,k}+M_{j,j}+\epsilon
    },
    \qquad j\neq k.
\end{equation}

This comparison keeps the latent $z^j$ fixed and changes only the context.
Thus, $D_{j,k}$ measures whether a latent performs preferentially in the
context from which it was extracted. Since the prediction errors are
nonnegative, $D_{j,k}\in[-1,1]$. A score near zero indicates that the latent
transfers with similar accuracy across same-action contexts, while a positive
score indicates increased error after transfer and therefore greater context
dependence. Negative values indicate cases in which the transferred latent
performs better than in its source context.

We aggregate across all off-diagonal transfers in a group:
\begin{equation}
    D_{\mathcal G}
    =
    \frac{1}{n(n-1)}
    \sum_{j=1}^{n}
    \sum_{k\neq j}
    D_{j,k}.
\end{equation}

We report the mean and standard deviation of $D_{\mathcal G}$ across evaluation groups. Lower values near zero indicate less context-specific information in the latent action representation, while systematically positive values indicate stronger leakage from the source transition.

\paragraph{Reference anchors.} To calibrate the scale we include two idealized references. \textit{Oracle} directly predicts each target context's true next frame, so $M_{j,k} = M_{j,j}$ for all pairs and $D_{j,k} = 0$. \textit{Future-copy} represents complete future-observation leakage: instead of encoding the underlying action, the ``latent'' from source transition $j$ is taken to be the full future frame $I^j_{t+1}$. Copying this representation across target contexts gives near-zero error in its source context but large cross-context error, so $D_{j,k}$ approaches $1$.

\paragraph{Why $\Delta_t$PSNR diverges for the reference anchors.}
Both anchors calibrate the leakage scale but not controllability. Since $\text{PSNR} = 10\log_{10}(\text{MAX}^2/\text{MSE})$, a prediction that matches its target exactly gives zero MSE and infinite PSNR, and both anchors are exact in the matched condition by construction: \textit{Oracle} predicts the true next frame for whichever action it is given, and \textit{Future-copy} reproduces the latent action source future frame it was handed. Under a random latent, drawn from the empirical latent distribution, each produces a different frame and therefore finite PSNR, so the difference between the two terms diverges. What that wrong prediction is differs between them. For Oracle, a random latent is a different action, which it executes correctly in the current scene, so the prediction is a valid transition that does not match the target. For Future-copy, the latent is itself a frame, so a random latent recovers an unrelated observation and a rollout becomes a sequence of disconnected frames rather than a trajectory. Oracle's random prediction therefore remains scene-consistent while Future-copy's does not, and this is the distinction the leakage score captures. We report the anchors only for leakage in Table~\ref{tab:controllability}, where they bound it on $[0,1]$.

\begin{figure*}[t]
    \centering
    \includegraphics[width=\textwidth]{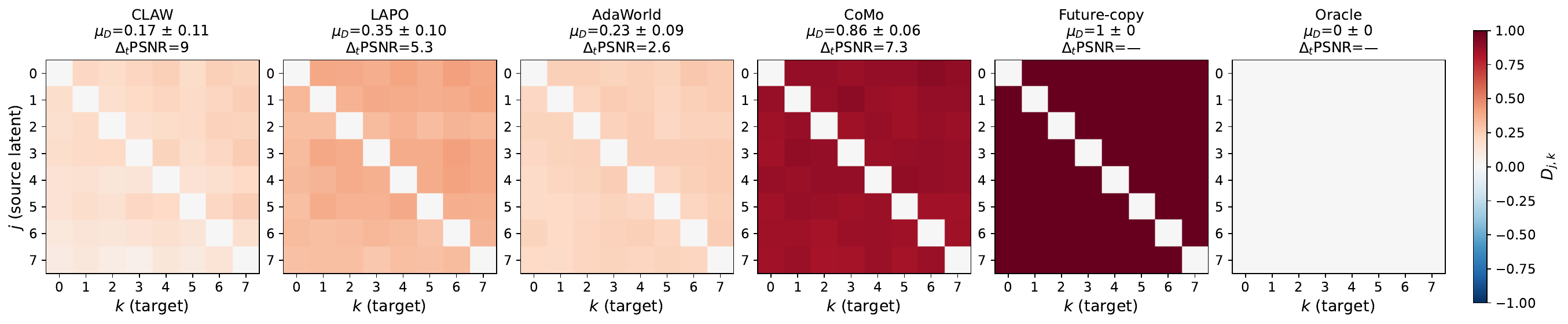}
    \caption{\textbf{Mean pairwise leakage matrices $D_{j,k}$ on a dataset taken from Procgen Ninja}, averaged over 200 same-action groups of size $n{=}8$. Rows index the source latent $z^j$ and columns the target context $k$. The diagonal is zero by construction, so only the off-diagonal entries carry information. A large positive value there means the latent predicts better in the context it came from than in other contexts with the same ground-truth action, i.e.\ the context-specific leakage defined above. The two rightmost panels bound the scale. Future-copy uses the source future frame $I^j_{t+1}$ in place of the latent, which is exact at home and useless anywhere else, so its off-diagonal entries approach 1. Oracle sees each target context's true next frame, so transfer costs it nothing and the matrix is zero throughout. Neither anchor conditions on a learned latent, so $\Delta_t$PSNR does not apply. Read the two metrics together: CLAW and AdaWorld have similarly pale matrices, but AdaWorld's $\Delta_t$PSNR of 2.6 against CLAW's 9 shows that its low leakage reflects weak latent conditioning rather than a cleaner action representation.}
    
    \label{fig:leakage_D_panels}
\end{figure*}

\paragraph{Interpreting low leakage.} A score near zero has two distinct causes, which the metric alone cannot separate. It arises when the latent genuinely transfers across contexts, but also when the world model makes little use of $z$: in that case $W(I^k_i, z^j)$ depends only weakly on $z^j$, so $M_{j,k}$ and $M_{j,j}$ are similar by construction and $D_{j,k} \to 0$ regardless of what the latent encodes. Leakage must therefore be interpreted alongside $\Delta_t$PSNR, which measures how strongly predictions depend on the latent in the first place. Figure~\ref{fig:leakage_D_panels} reports both for each method.

\paragraph{Spatial leakage decomposition.}
The scalar leakage score measures the magnitude of context dependence but does
not indicate where the resulting prediction degradation occurs. To visualize
its spatial origin, we define the per-pixel squared prediction error
\begin{equation}
    e_{j,k}(p)
    =
    \left\|
    \hat I_{i+1}^{j\rightarrow k}(p)
    -
    I_{i+1}^{k}(p)
    \right\|^2
\end{equation}
and construct the normalized spatial difference map
\begin{equation}
    \mathbf{D}_{j,k}(p)
    =
    \frac{
    e_{j,k}(p)-e_{j,j}(p)
    }{
    M_{j,k}+M_{j,j}+\epsilon
    }.
\end{equation}

Positive regions identify image locations where transferring the latent increases prediction error relative to its matched context, while negative regions indicate improved prediction after transfer. The denominator is shared across all pixels in a source-target pair, avoiding instability from pixel-specific normalization. Importantly, spatially averaging the map recovers the scalar leakage score:
\begin{equation}
    \frac{1}{|\mathcal P|}
    \sum_{p\in\mathcal P}
    \mathbf{D}_{j,k}(p)
    =
    D_{j,k}.
\end{equation}
Thus, $D_{j,k}$ measures how much context-specific leakage is present, while
$\mathbf{D}_{j,k}(p)$ visualizes where its effect appears in the predicted
frame.

Figure~\ref{fig:leakage_D_panels} visualizes the group-averaged leakage matrices $D_{j,k}$ on Ninja. Because each group shares a fixed ground-truth action, the off-diagonal entries measure context dependence of the latent: values near zero indicate that $z^j$ transfers about as well as in its source context, whereas systematically positive values indicate that the world model preferentially uses the latent in the context from which it was inferred. The two outer panels bound this scale. \textit{Oracle} has access to each target context's true next frame, so transfer costs nothing and the matrix is uniformly zero. \textit{Future-copy} replaces the latent with the source future frame $I^j_{t+1}$, which is exact in its source context and uninformative everywhere else, saturating the off-diagonal entries near $1$. The learned methods fall between these extremes, and their position is interpretable only alongside $\Delta_t$PSNR: high $\Delta_t$PSNR with high $\mu_D$ indicates a latent that is informative but context-entangled, whereas lower $\mu_D$ at comparable controllability indicates greater cross-context consistency of the action representation.

\subsubsection{Action Transfer}
\label{section:action_transfer_supp}

In Figure~\ref{fig:action_swap} and Figure~\ref{fig:action_retrieval_robosuite_supp}, \alg's end-effector motion corresponds to rightward and slightly downward displacement much like the source (yellow in the Middlebury color wheel~\citep{baker2011database}, key in Figure~\ref{fig:middleburycolorwheel}), even when object identity and scene layout differ substantially from the query. This consistency is reflected in the optical flow visualizations, where retrieved neighbors share a coherent dominant flow orientation, in contrast to retrievals from baseline methods that exhibit heterogeneous or weakly aligned flow patterns. 

\begin{figure}[h]
    \centering
    \includegraphics[width=\linewidth]{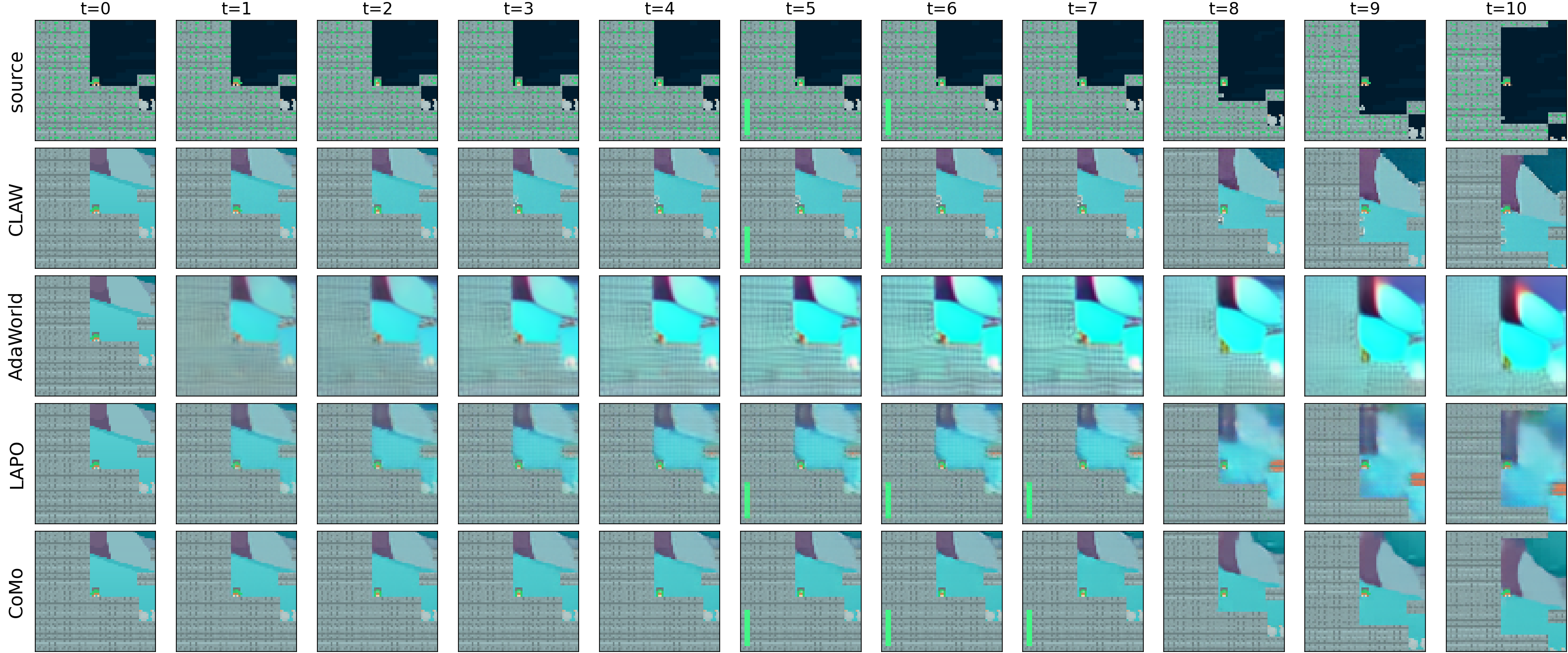}
    \caption{Action transfer on Procgen Ninja. Top: source trajectory. Below: rollouts on the target initial frame using CLAW, AdaWorld, LAPO and CoMo.}
    \label{fig:action_swap_supp_ninja}
\end{figure}

\begin{figure}[!htbp]
    \centering
    \includegraphics[width=\linewidth]{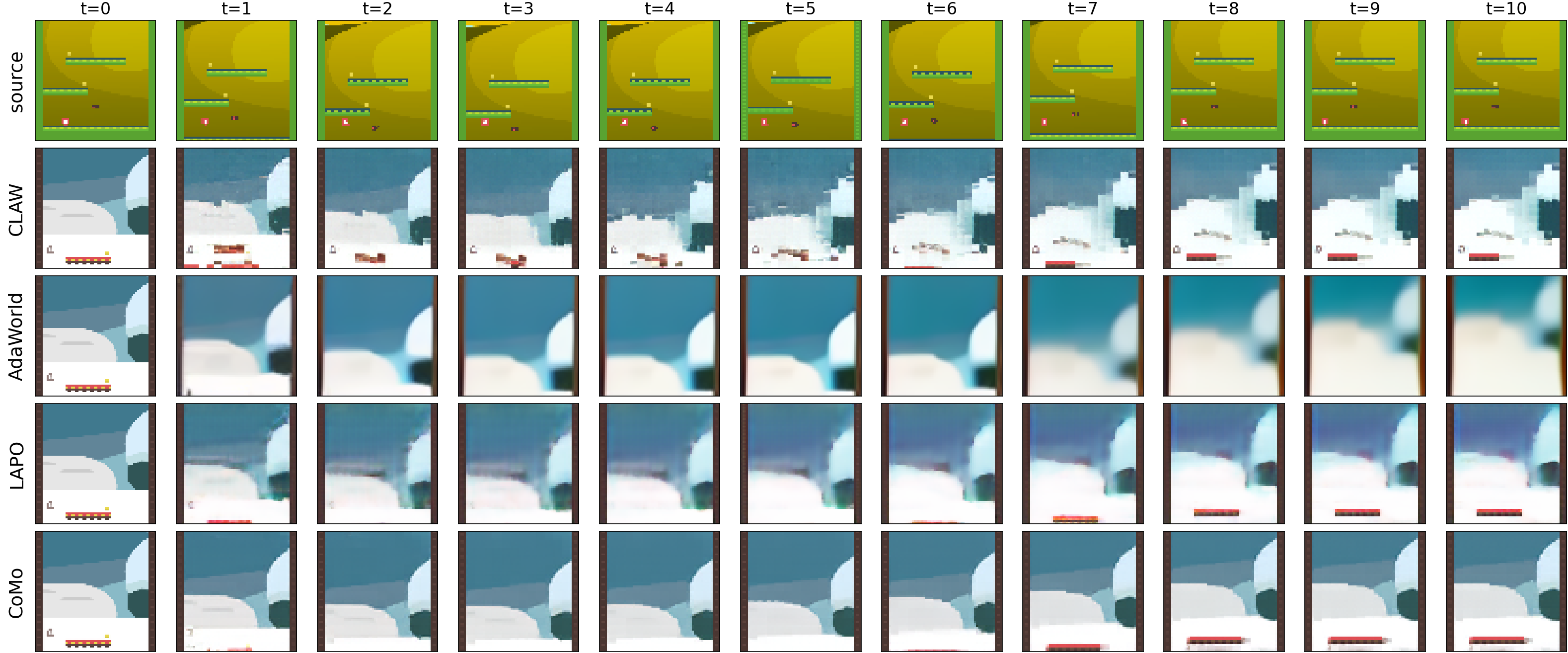}
    \caption{Action transfer on Procgen Climber. Top: source trajectory. Below: rollouts on the target initial frame using CLAW, AdaWorld, LAPO and CoMo.}
    \label{fig:action_swap_supp_climber}
\end{figure}

\begin{figure}[!htbp]
    \centering
    \includegraphics[width=\linewidth]{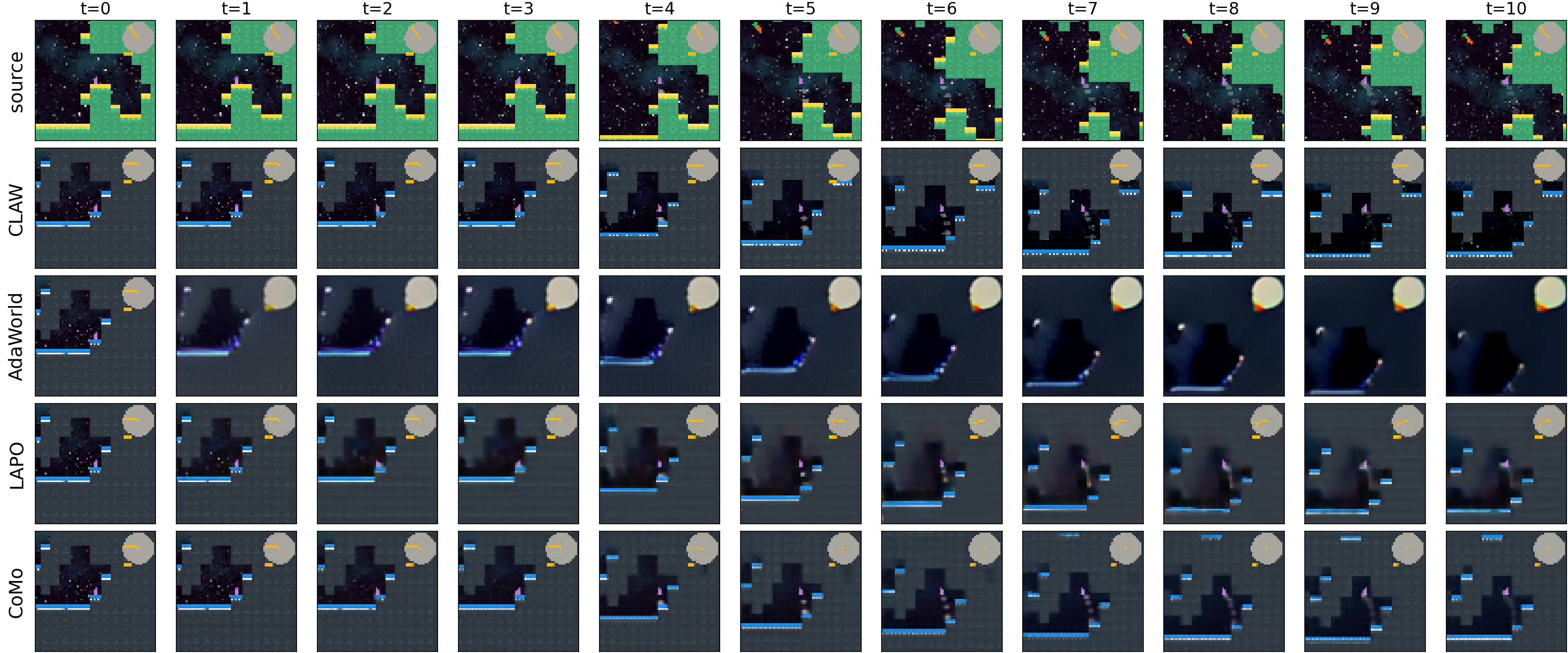}
    \caption{Action transfer on Procgen Jumper. Top: source trajectory. Below: rollouts on the target initial frame using CLAW, AdaWorld, LAPO and CoMo.}
    \label{fig:action_swap_supp_jumper}
\end{figure}

Additional Action Transfer experiments in the Procgen Ninja, Climber, and Jumper environments are shown in Figure~\ref{fig:action_swap_supp_ninja}, Figure~\ref{fig:action_swap_supp_climber}, and Figure~\ref{fig:action_swap_supp_jumper}, respectively. We compare \alg, AdaWorld, CoMo, and LAPO in each environment. Across all environments, as predictions progress further away from the initial observation, AdaWorld and LAPO show greater degradation than \alg.

\subsubsection{Visual Planning Task}
\label{section:visual_planning_supp}

\begin{figure}[t]
    \centering

    \begin{subfigure}[b]{0.9\linewidth}
        \centering

        \begin{subfigure}[b]{0.25\linewidth}
            \centering
            \begin{subfigure}[b]{0.48\linewidth}
                \centering
                \includegraphics[width=\linewidth]{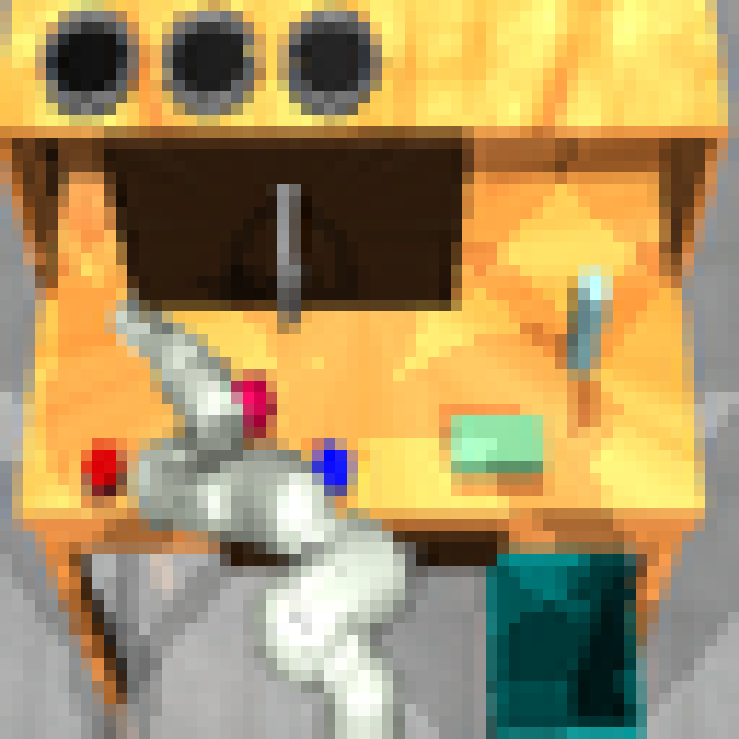}
                \caption{Initial}
            \end{subfigure}
            \hfill
            \begin{subfigure}[b]{0.48\linewidth}
                \centering
                \includegraphics[width=\linewidth]{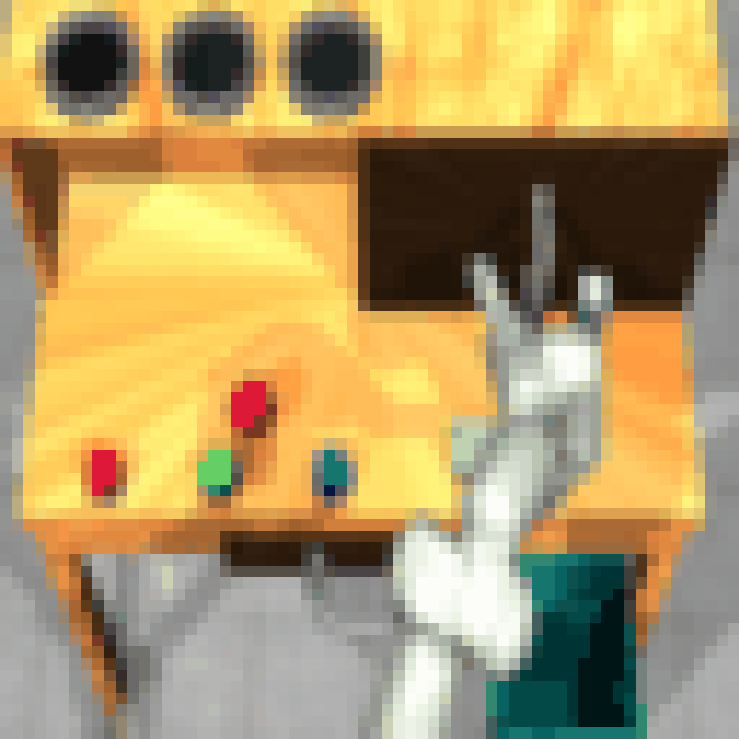}
                \caption{Goal}
            \end{subfigure}
        \end{subfigure}

        \vspace{0.3em}

        \begin{subfigure}[b]{\linewidth}
            \centering
            \includegraphics[width=\linewidth]{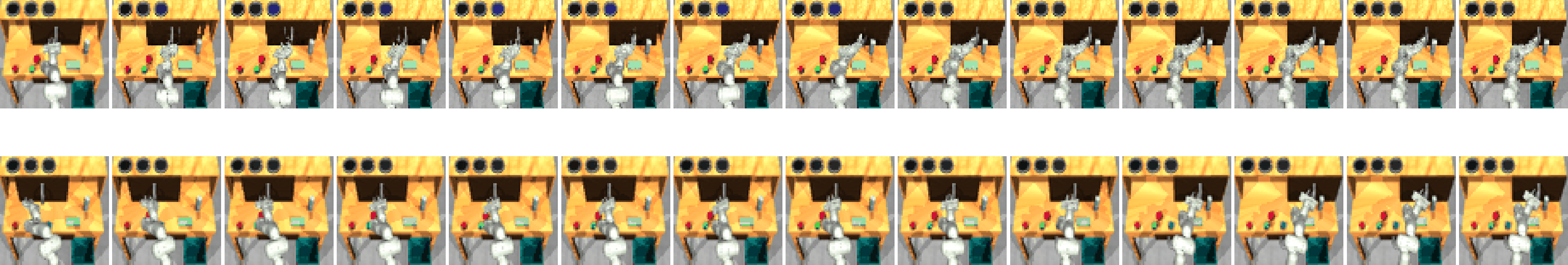}
            \caption{Successful trajectory for \texttt{open slide}}
        \end{subfigure}

    \end{subfigure}

    \vspace{0.6em}

    \begin{subfigure}[b]{0.9\linewidth}
        \centering

        \begin{subfigure}[b]{0.25\linewidth}
            \centering
            \begin{subfigure}[b]{0.48\linewidth}
                \centering
                \includegraphics[width=\linewidth]{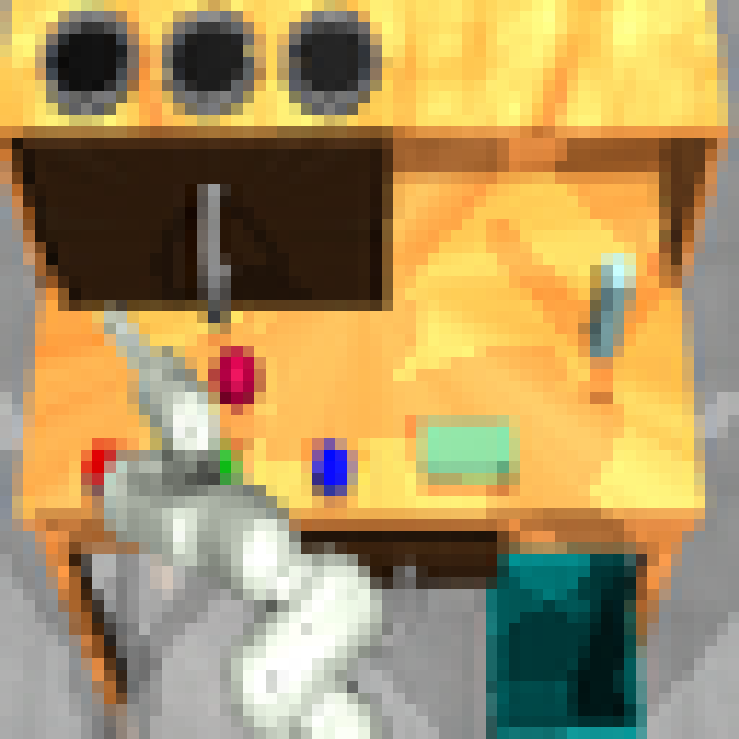}
                \caption{Initial}
            \end{subfigure}
            \hfill
            \begin{subfigure}[b]{0.48\linewidth}
                \centering
                \includegraphics[width=\linewidth]{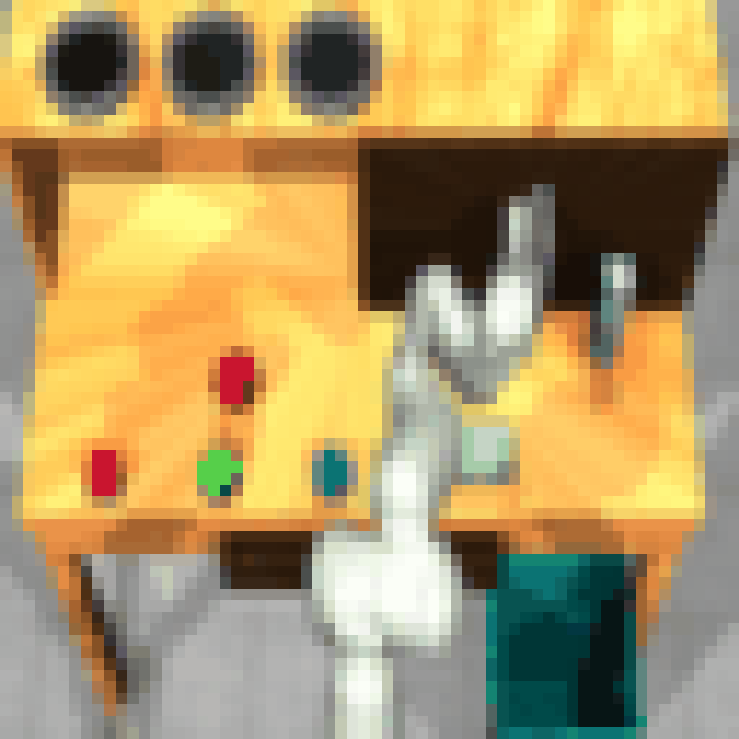}
                \caption{Goal}
            \end{subfigure}
        \end{subfigure}

        \vspace{0.3em}

        \begin{subfigure}[b]{\linewidth}
            \centering
            \includegraphics[width=\linewidth]{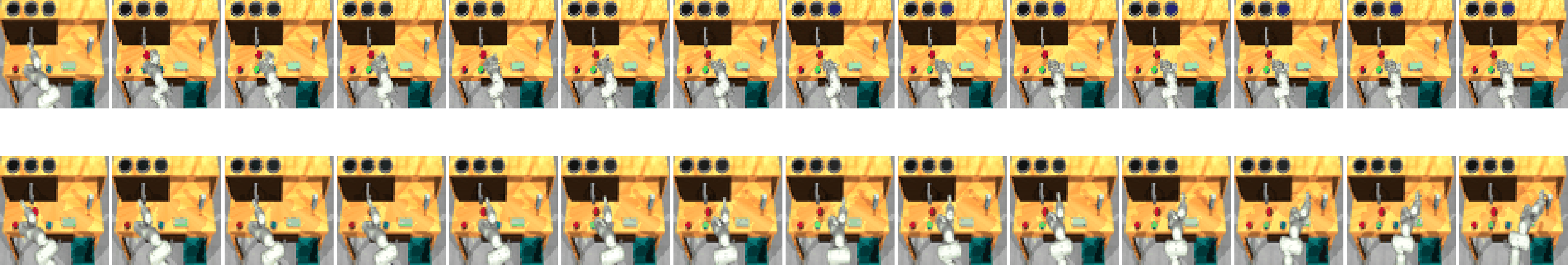}
            \caption{Failure trajectory for \texttt{open slide}}
        \end{subfigure}

    \end{subfigure}

    \caption{
    Visual planning results in the \textbf{VP2} benchmark using the \textbf{robodesk} environment for the task \texttt{open slide}. We show both successful and failed trajectories generated by the planner.
    }
    \label{fig:robodesk_open slide_planning}
\end{figure}

\begin{figure}[t]
    \centering

    \begin{subfigure}[b]{0.9\linewidth}
        \centering

        \begin{subfigure}[b]{0.25\linewidth}
            \centering
            \begin{subfigure}[b]{0.48\linewidth}
                \centering
                \includegraphics[width=\linewidth]{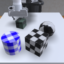}
                \caption{Initial}
            \end{subfigure}
            \hfill
            \begin{subfigure}[b]{0.48\linewidth}
                \centering
                \includegraphics[width=\linewidth]{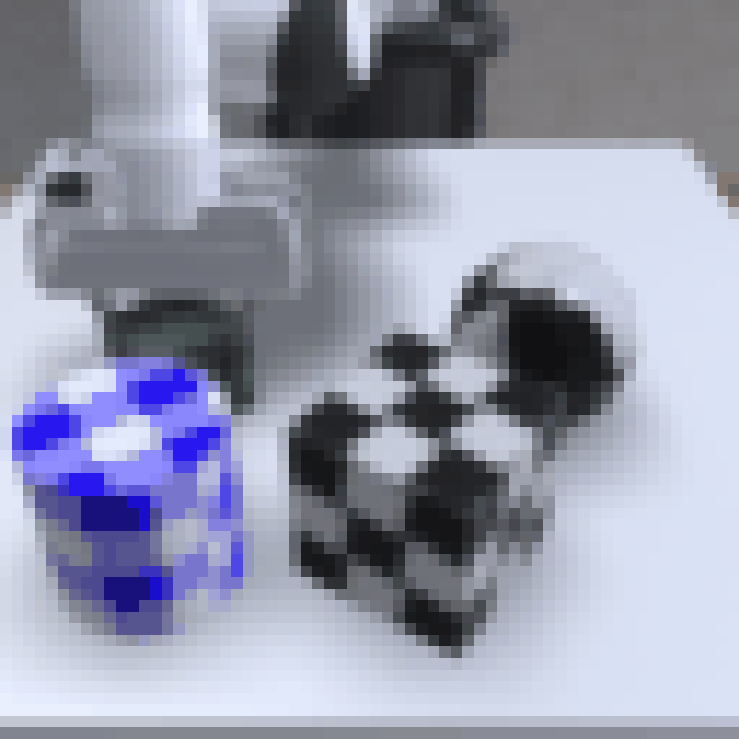}
                \caption{Goal}
            \end{subfigure}
        \end{subfigure}

        \vspace{0.3em}

        \begin{subfigure}[b]{\linewidth}
            \centering
            \includegraphics[width=\linewidth]{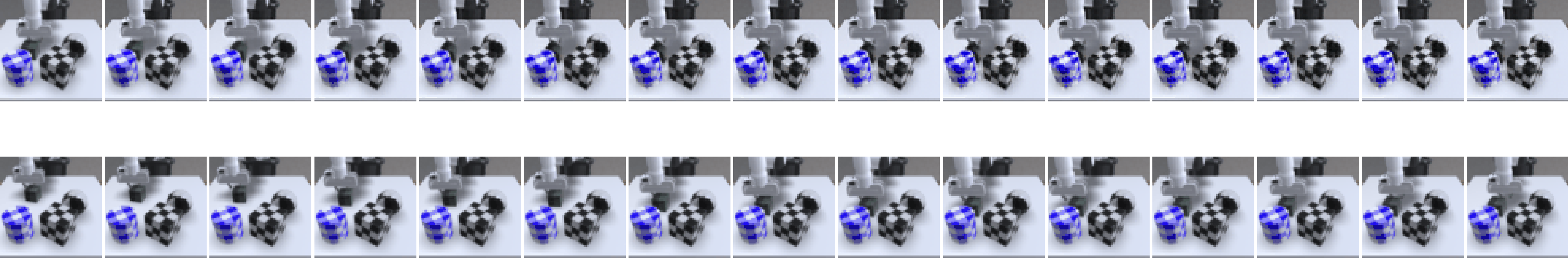}
            \caption{Successful trajectory for \texttt{push}}
        \end{subfigure}

    \end{subfigure}

    \vspace{0.6em}

    \begin{subfigure}[b]{0.9\linewidth}
        \centering

        \begin{subfigure}[b]{0.25\linewidth}
            \centering
            \begin{subfigure}[b]{0.48\linewidth}
                \centering
                \includegraphics[width=\linewidth]{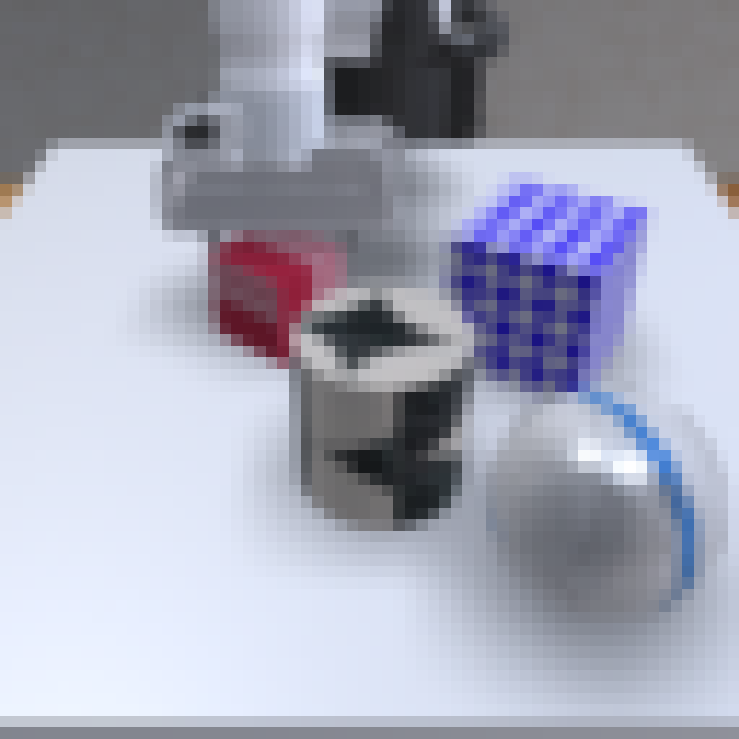}
                \caption{Initial}
            \end{subfigure}
            \hfill
            \begin{subfigure}[b]{0.48\linewidth}
                \centering
                \includegraphics[width=\linewidth]{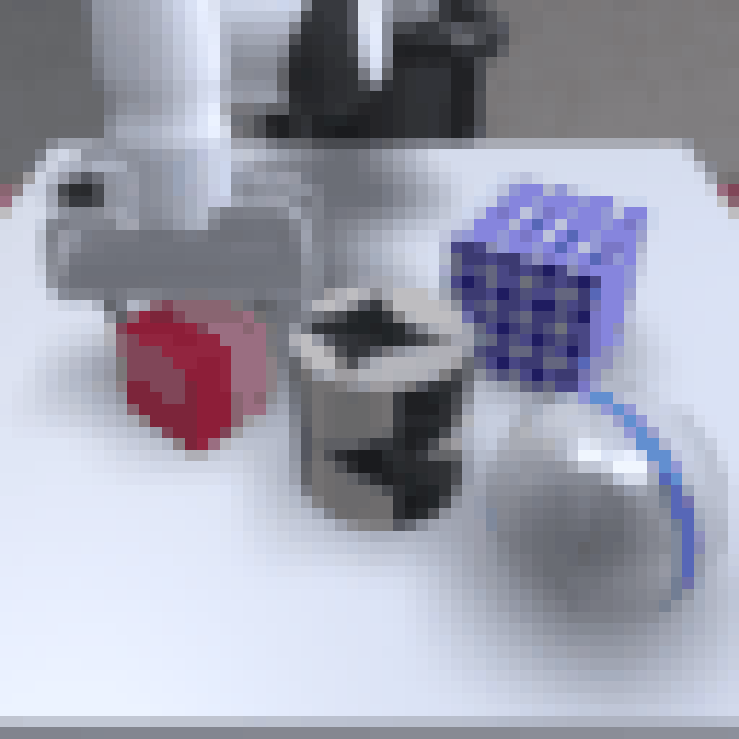}
                \caption{Goal}
            \end{subfigure}
        \end{subfigure}

        \vspace{0.3em}

        \begin{subfigure}[b]{\linewidth}
            \centering
            \includegraphics[width=\linewidth]{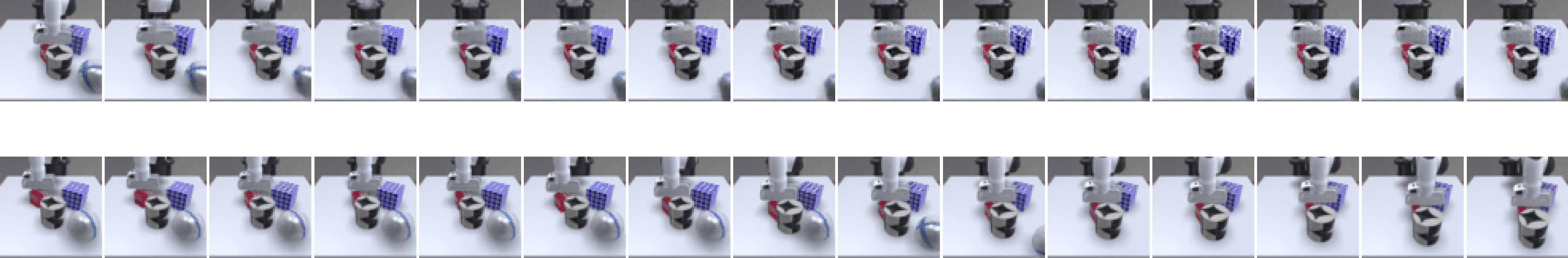}
            \caption{Failure trajectory for \texttt{push}}
        \end{subfigure}

    \end{subfigure}

    \caption{
    Visual planning results in the \textbf{Robosuite} benchmark using the \textbf{robosuite} environment for the task \texttt{push}. We show both successful and failed trajectories generated by the planner.
    }
    \label{fig:robosuite_push_planning}
\end{figure}

\begin{figure}[t]
    \centering

    \begin{subfigure}[b]{0.9\linewidth}
        \centering

        \begin{subfigure}[b]{0.25\linewidth}
            \centering
            \begin{subfigure}[b]{0.48\linewidth}
                \centering
                \includegraphics[width=\linewidth]{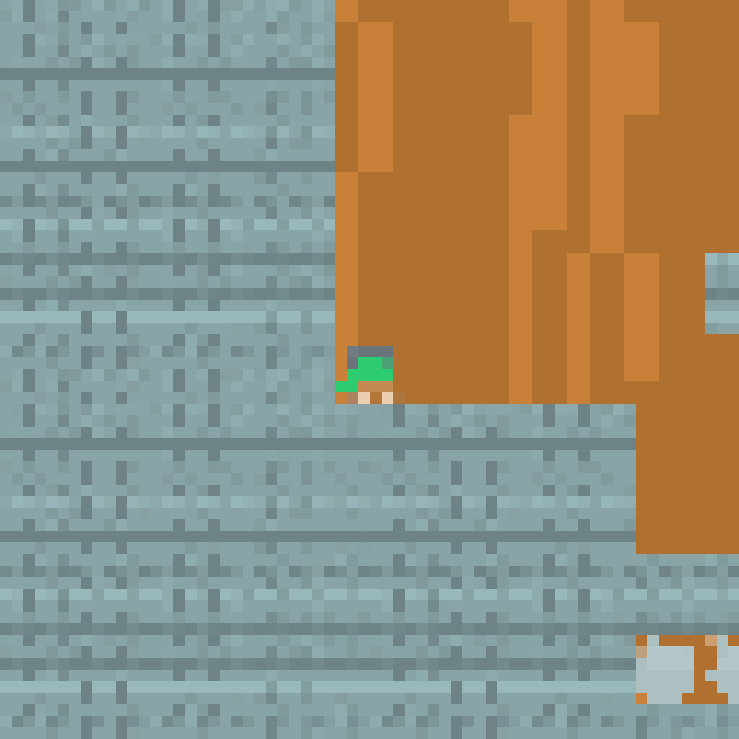}
                \caption{Initial}
            \end{subfigure}
            \hfill
            \begin{subfigure}[b]{0.48\linewidth}
                \centering
                \includegraphics[width=\linewidth]{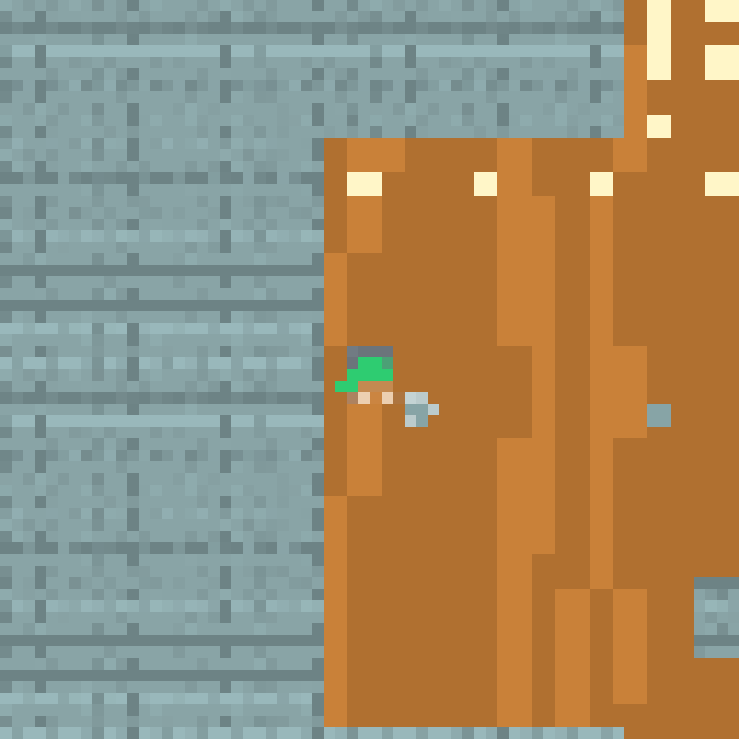}
                \caption{Goal}
            \end{subfigure}
        \end{subfigure}

        \vspace{0.3em}

        \begin{subfigure}[b]{\linewidth}
            \centering
            \includegraphics[width=\linewidth]{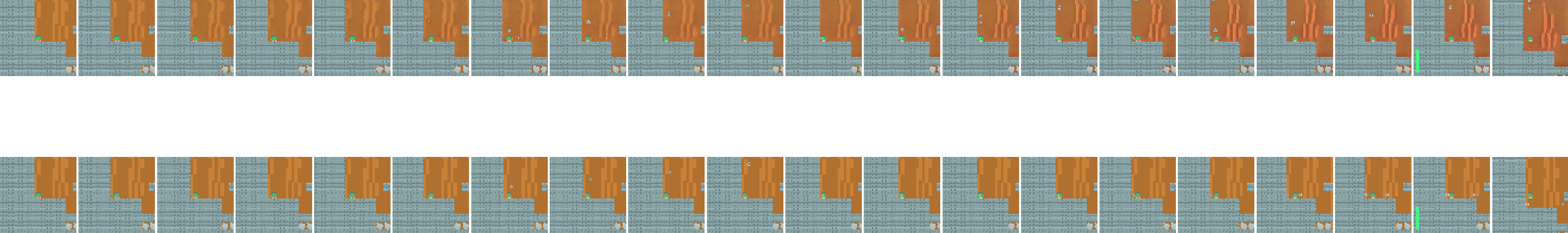}
            \caption{Successful trajectory for \texttt{Ninja}}
        \end{subfigure}

    \end{subfigure}

    \vspace{0.6em}

    \begin{subfigure}[b]{0.9\linewidth}
        \centering

        \begin{subfigure}[b]{0.25\linewidth}
            \centering
            \begin{subfigure}[b]{0.48\linewidth}
                \centering
                \includegraphics[width=\linewidth]{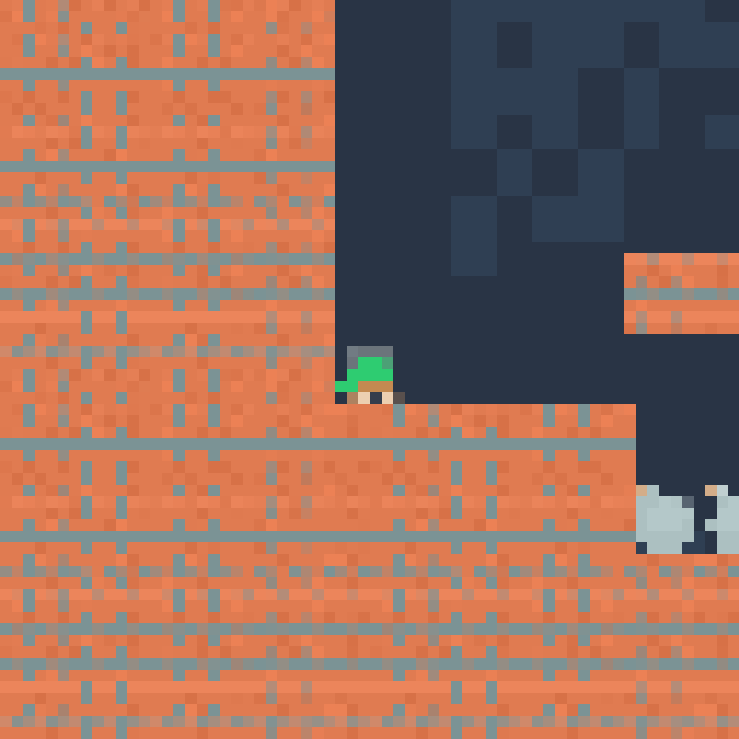}
                \caption{Initial}
            \end{subfigure}
            \hfill
            \begin{subfigure}[b]{0.48\linewidth}
                \centering
                \includegraphics[width=\linewidth]{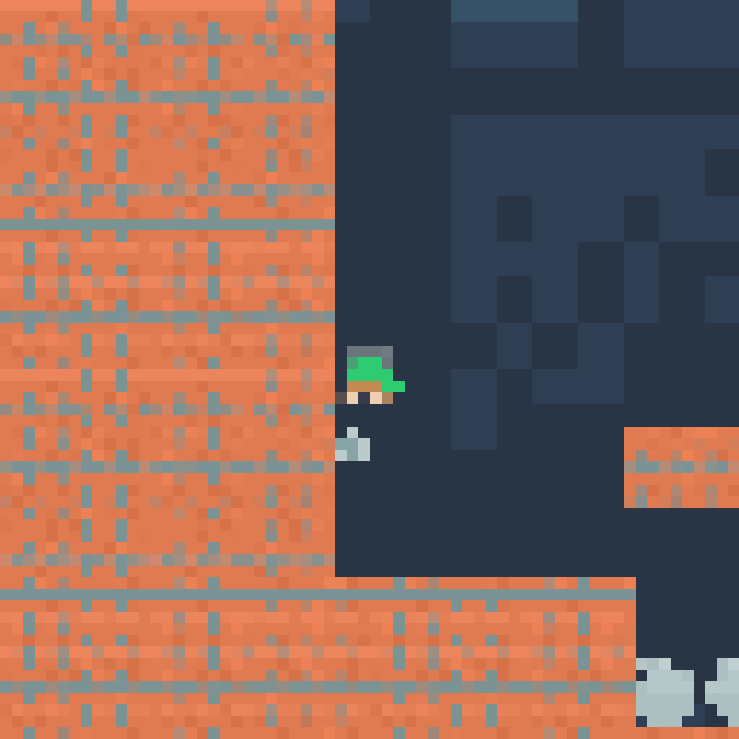}
                \caption{Goal}
            \end{subfigure}
        \end{subfigure}

        \vspace{0.3em}

        \begin{subfigure}[b]{\linewidth}
            \centering
            \includegraphics[width=\linewidth]{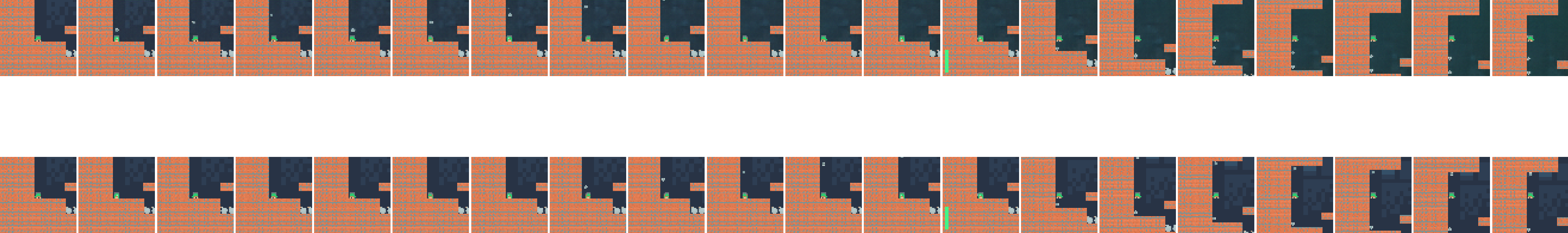}
            \caption{Failure trajectory for \texttt{Ninja}}
        \end{subfigure}

    \end{subfigure}

    \caption{
    Visual planning results in the \textbf{ProcGen} benchmark using the \textbf{ninja} environment for the task \texttt{Ninja}. We show both successful and failed trajectories generated by the planner.
    }
    \label{fig:ninja_Ninja_planning}
\end{figure}

\begin{figure}[t]
    \centering

    \begin{subfigure}[b]{0.9\linewidth}
        \centering

        \begin{subfigure}[b]{0.25\linewidth}
            \centering
            \begin{subfigure}[b]{0.48\linewidth}
                \centering
                \includegraphics[width=\linewidth]{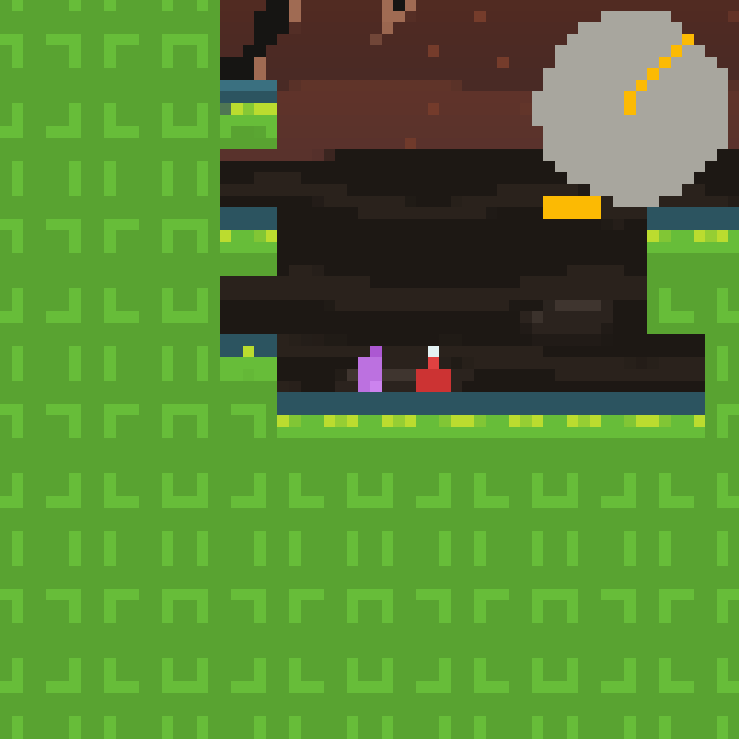}
                \caption{Initial}
            \end{subfigure}
            \hfill
            \begin{subfigure}[b]{0.48\linewidth}
                \centering
                \includegraphics[width=\linewidth]{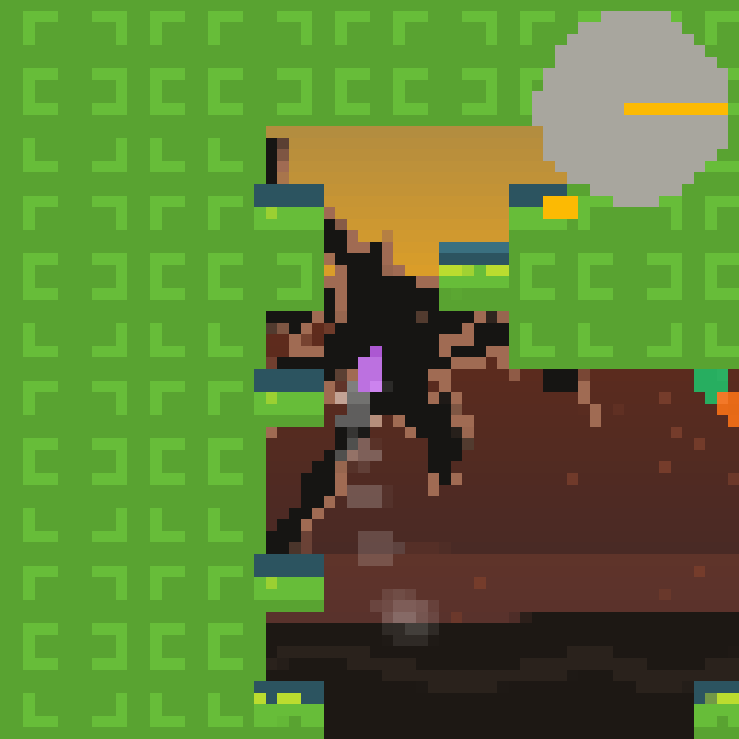}
                \caption{Goal}
            \end{subfigure}
        \end{subfigure}

        \vspace{0.3em}

        \begin{subfigure}[b]{\linewidth}
            \centering
            \includegraphics[width=\linewidth]{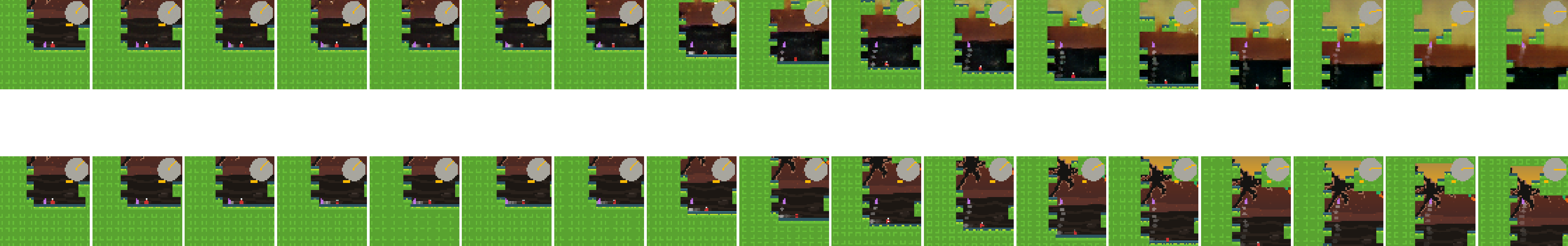}
            \caption{Successful trajectory for \texttt{Jumper}}
        \end{subfigure}

    \end{subfigure}

    \vspace{0.6em}

    \begin{subfigure}[b]{0.9\linewidth}
        \centering

        \begin{subfigure}[b]{0.25\linewidth}
            \centering
            \begin{subfigure}[b]{0.48\linewidth}
                \centering
                \includegraphics[width=\linewidth]{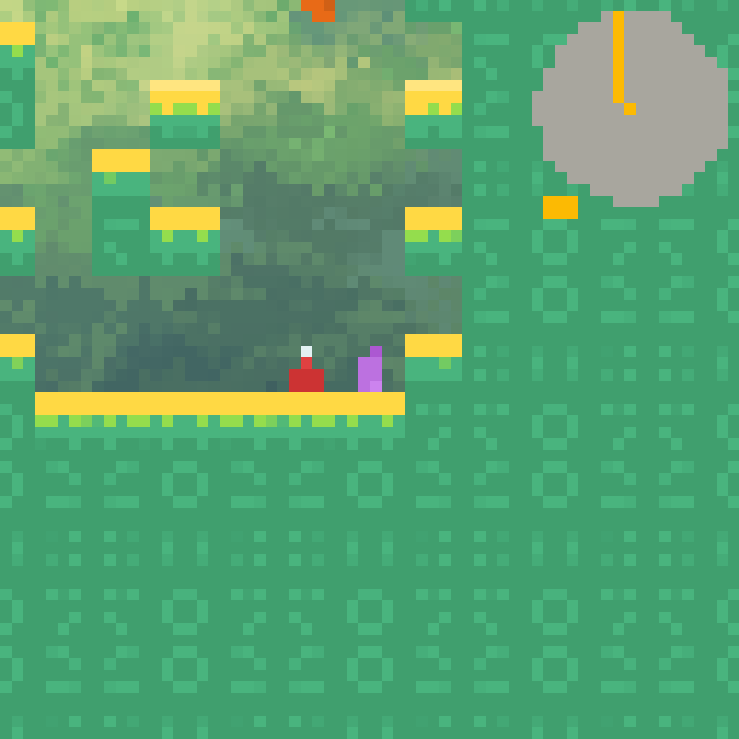}
                \caption{Initial}
            \end{subfigure}
            \hfill
            \begin{subfigure}[b]{0.48\linewidth}
                \centering
                \includegraphics[width=\linewidth]{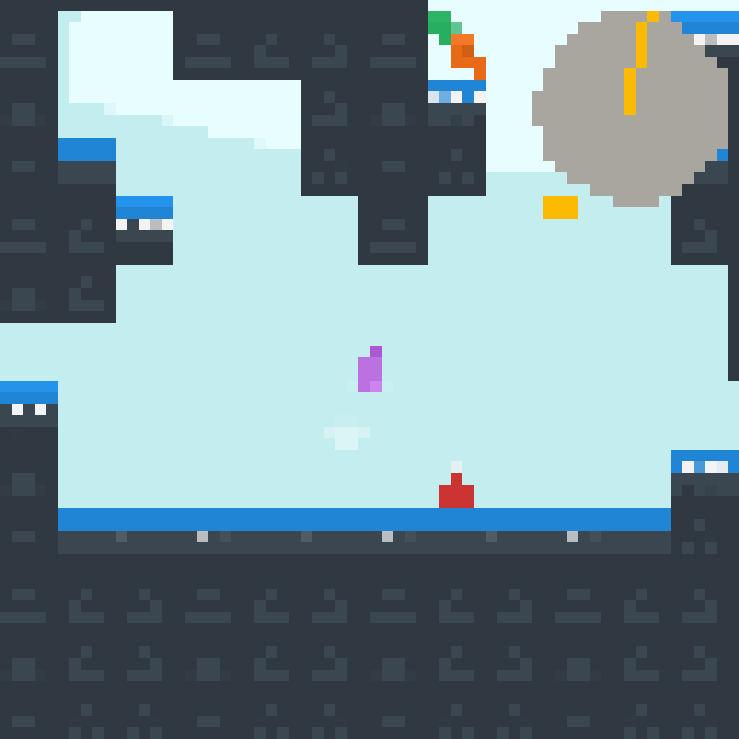}
                \caption{Goal}
            \end{subfigure}
        \end{subfigure}

        \vspace{0.3em}

        \begin{subfigure}[b]{\linewidth}
            \centering
            \includegraphics[width=\linewidth]{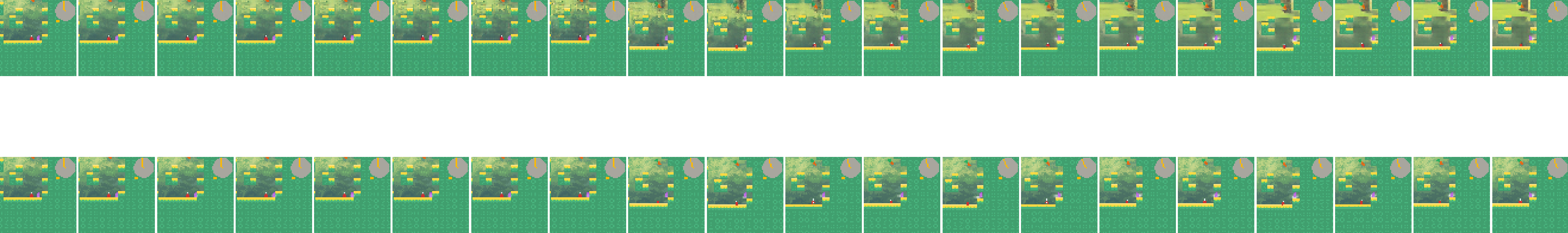}
            \caption{Failure trajectory for \texttt{Jumper}}
        \end{subfigure}

    \end{subfigure}

    \caption{
    Visual planning results in the \textbf{ProcGen} benchmark using the \textbf{jumper} environment for the task \texttt{Jumper}. We show both successful and failed trajectories generated by the planner.
    }
    \label{fig:jumper_Jumper_planning}
\end{figure}

We display successful and failed visual planning trajectories for the VP2 Robodesk and Robosuite benchmarks in Figure~\ref{fig:robodesk_open slide_planning} and Figure~\ref{fig:robosuite_push_planning}. Visual planning trajectory successes and failures for the Procgen Ninja and Jumper environments are shown in Figure~\ref{fig:ninja_Ninja_planning} and Figure~\ref{fig:jumper_Jumper_planning}.

%% file: extended_sup.tex
\clearpage

\subsubsection{World Model Controllability and Transfer.}
\label{app:controllability_discussion}
Beyond the latent action representation, our results also characterize how the
learned world model uses these latents to model environment dynamics.
Sec.~\ref{sec:controllability} shows that CLAW's predictions are strongly
conditioned on the latent action while remaining comparatively robust to
cross-context transfer of latents representing the same semantic action. The
high $\Delta_t$PSNR indicates that different latent inputs induce meaningfully
different predicted futures, while the bounded leakage analysis shows that this
conditioning is not explained solely by source-context-specific information.
Together, these results suggest that the world model learns dynamics that are meaningfully conditioned on the latent action, with different latent inputs inducing distinct and transferable future transitions.

This interpretation is further supported by the planning experiments in
Sec.~\ref{sec:planning}, where optimizing latent action sequences through the
world model produces goal-directed behavior. Thus, the learned forward dynamics
are not only predictive, but sufficiently action-sensitive and transferable to
support downstream planning.

\subsubsection{What the Adversarial Penalty Removes}
\label{sec:adversarial_scope}
The adversarial pathway penalizes prediction of the future observation from $z_i$ alone, which raises a question about action spaces in which the action is highly predictive of the next frame. If a robot is controlled in absolute joint coordinates, the action determines the end-effector configuration outright, and a latent that faithfully encodes it will carry real information about the future. Penalizing that would remove action content rather than leakage.

The distinction is context specific. An absolute-coordinate action predicts where the end-effector will be in any scene; it does not predict which objects are present, where they sit, or what the background looks like. Because the adversarial pathway receives only $z_i$, its loss decreases only through information that predicts this particular future without the current observation, so the gradient pushes against scene-specific content rather than against action content that generalizes across contexts. Whether the optimization separates the two cleanly is an empirical question, and one the leakage metric (Appendix~\ref{app:leakage_metric}) is designed to probe.

The action transfer results in Sec.~\ref{sec:action_transfer} are consistent with this account. We observe that the learned latents specify end-effector motion directly: transferred to a different initial frame, they reproduce the source end-effector trajectory while the objects respond according to the new scene. This suggests the absolute component of the action survives the regularizer, though we have not isolated it directly.

The penalty is applied uniformly, with a schedule rather than a learned per-dimension weighting, so in domains where the action determines almost the entire observation the pressure may suppress action content along with context. We did not observe this in our settings, but we have not tested action spaces in which the agent controls the full scene.

\subsubsection{Sampling Diversity Under a Fixed Latent Action}
\label{sec:sampling_diversity}
\begin{figure}[t]
    \centering
    \includegraphics[width=\linewidth]{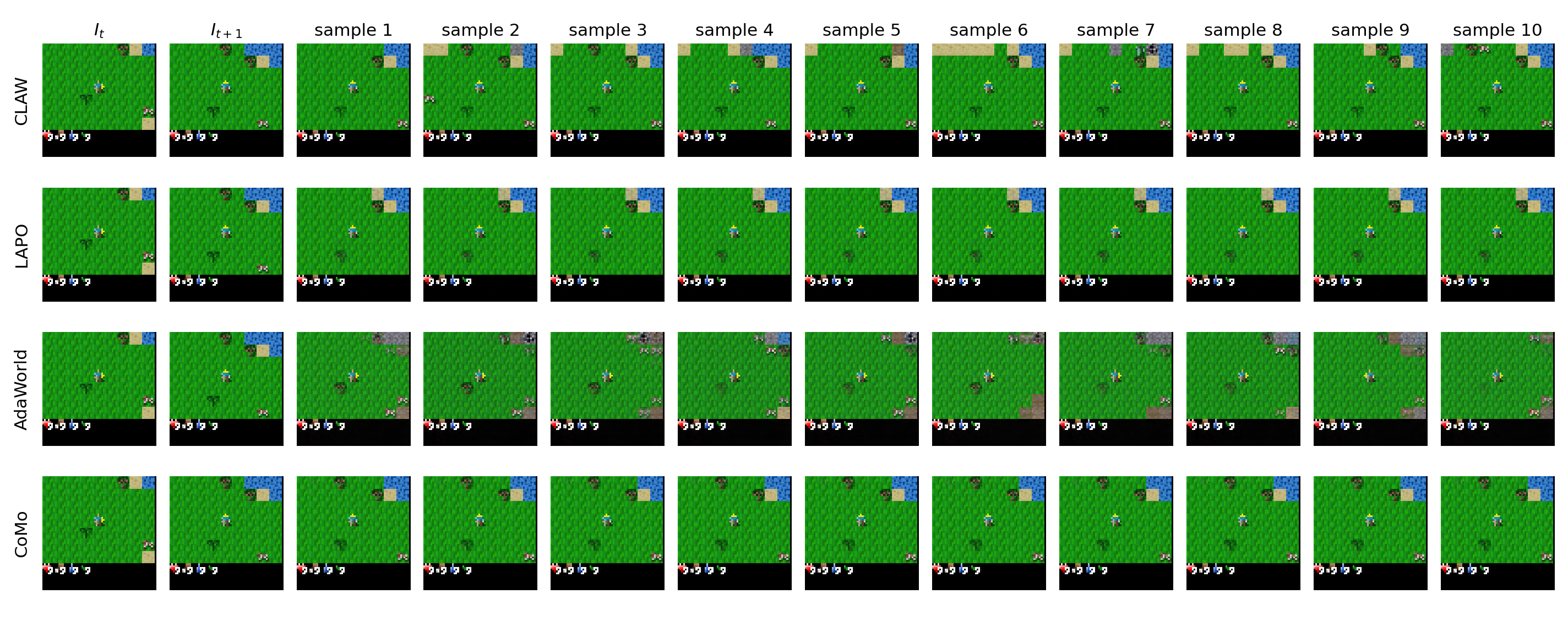}
    \caption{\textbf{Sampling diversity on a shared \texttt{up} transition in Crafter.} The first two columns show the frames the latent $z$ is extracted from. The remaining columns are ten independent samples of $\hat{I} = W(I_t, z)$ under the same latent, varying only the sampling seed. The motion is the same in every sample; only the newly visible terrain along the top edge changes. LAPO and CoMo are deterministic, so each seed returns the same frame.}
    \label{fig:sampling_diversity}
\end{figure}

The Crafter view is centered on the agent, so moving up exposes a previously off-screen row whose contents are not recoverable from $I_t$. Figure~\ref{fig:sampling_diversity} shows ten samples from each world model conditioned on the same initial frame and the same latent action. \alg reproduces the same scrolling motion in every sample while varying the newly revealed terrain, so the samples agree on what the action does and differ only where the outcome is genuinely unknown. LAPO and CoMo decode deterministically and return an identical frame for every seed, committing to one completion of a region they cannot observe.

\subsection{Further Details For the Real-World Experiments}
\label{app:realworld_details}
\begin{figure}
    \centering
    \includegraphics[width=0.9\linewidth]{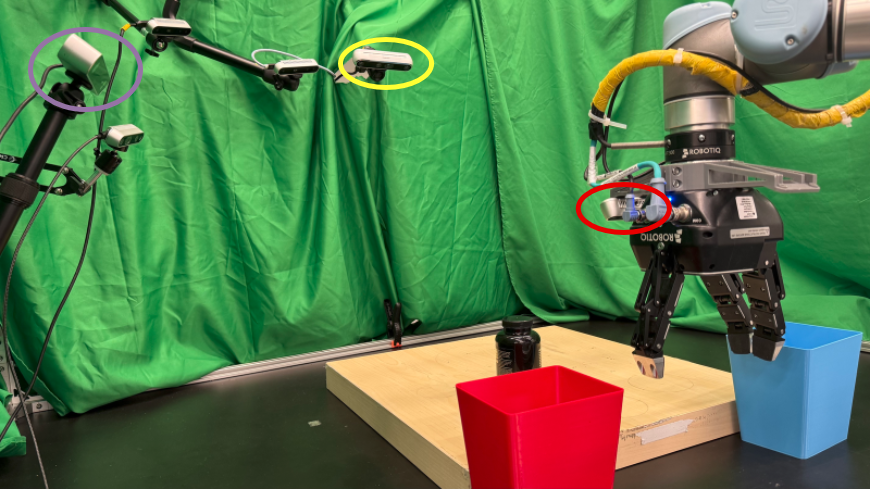}
    \caption{Workspace setup for the real robot experiments in Sec.~\ref{sec:realworld_pretraining}. Three camera views are used for observation: a wrist-mounted camera (\texttt{cam2}, \textcolor{red}{red}) and two third-person cameras (\texttt{cam5}, \textcolor{yellow}{yellow}, and \texttt{cam6}, \textcolor{violet}{purple}).}
    \label{fig:robotic_setup}
\end{figure}

As shown in Figure~\ref{fig:robotic_setup}, a UR5 robotic arm is mounted on a fixed base in a tabletop manipulation setting. The workspace is captured from three cameras: a wrist-mounted camera (\texttt{cam2}, \textcolor{red}{red}) that travels with the end-effector for a close-up egocentric view, and two static third-person cameras (\texttt{cam5}, \textcolor{yellow}{yellow}, and \texttt{cam6}, \textcolor{violet}{purple}) that provide complementary perspectives of the full workspace. Demonstrations are collected via teleoperation, with each rollout recorded as a synchronized multi-view RGB trajectory.

\subsubsection{UR5 Latent Action Extraction.}
We extract latent actions offline from frozen pretrained models on UR5 teleoperation demonstrations. Each successful rollout provides synchronized RGB observations from all three camera views (\texttt{cam2}, \texttt{cam5}, and \texttt{cam6}). Demonstrations are processed episode-by-episode, with latents indexed by the timestep of the first frame in each encoded transition (our method and AdaWorld) or by the observation timestep itself (LAPA). All models are run in evaluation mode with gradients disabled, and RGB inputs are resized to $224\times224$ with bilinear interpolation where required.

\paragraph{\alg.}
We use a frozen LAM trained on Something-Something V2 Dataset. For each camera view, the encoder receives a two-frame clip $(o_j,\, o_{j+\Delta})$ with $\Delta=3$, so each latent summarizes the visual change over three control steps. 

\paragraph{AdaWorld.}
We apply the publicly released AdaWorld LAM with the same $32$-d latent dimension. On UR5, the LAM is fed consecutive frame pairs $(o_{j-1},\, o_j)$, with latents subsampled every $\Delta{-}1=2$ timesteps to reduce redundancy at the demonstration control frequency. Extraction is performed separately for all three camera views.

\paragraph{LAPA.}
LAPA uses a qualitatively different interface: a frozen $7$B \texttt{DeltaSampler} that predicts discrete delta tokens rather than a continuous transition embedding. For every frame, the image is paired with a fixed task-specific natural-language instruction (``Pick up the bottle and place it in the recycling bin'' and ``Push the ramen to the box''), and the \texttt{DeltaSampler} autoregressively generates four delta token IDs from a vocabulary of eight. These token IDs serve as the downstream latent representation and are distinct from the continuous pre-quantization vectors in the LAQ submodule.

\paragraph{CoMo.} We use a frozen latent-motion model pretrained on Something-Something v2. As with \alg{}, the encoder takes a two-frame clip $(o_j, o_{j+\Delta})$ with $\Delta=3$, so each latent covers three control steps. A frozen ViT-MAE embeds both frames, and eight learned motion queries read out the difference between their patch tokens, each projected to $16$ dimensions for a continuous $128$-d code. No vector quantization is applied.

\section{Limitations}
\label{sec:limitations}
The LAM encodes whatever changed between two frames, and it has no way to tell an agent action from an environment-induced transition. Crafter's day-night cycle is one case: ground-truth \texttt{Noop} transitions that coincide with a lighting change sit farther from the learned no-op centroid than idle transitions do (Figure~\ref{fig:pseudo-noop-examples}), because the LAM registers the change even though the agent caused none of it. Procgen Ninja shows the same effect during rollout, where lights appearing in the scene are treated as part of the action (Figure~\ref{fig:action_swap_supp_ninja}). This is not specific to \alg: every method that generates visually plausible future observation reproduces the lights, since each infers its latent from the full visual transition. Factoring the latent into agent-driven and environment-driven components is a natural next step, which we leave to future work.

\begin{figure}[h]
    \centering
    \includegraphics[width=.5\linewidth]{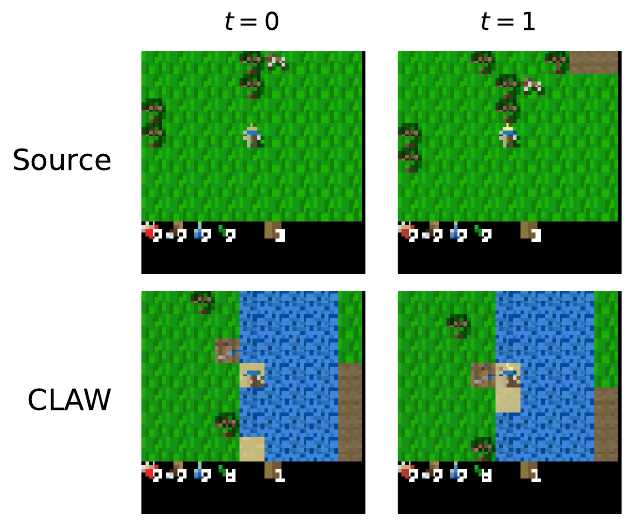}
    \caption{\textbf{Physically implausible rollout in Crafter.} Given a source transition encoding upward movement, the world model moves the agent onto a water tile by replacing it with sand. The environment would return a \texttt{no-op}.}
    \label{fig:implausible_rollout}
\end{figure}

The world model is also free to roll out transitions the environment would never permit. Because it is trained to generate plausible continuations rather than to respect the environment's rules, it can satisfy a latent action by changing the scene instead of refusing the action. Figure~\ref{fig:implausible_rollout} shows this in Crafter: given a latent corresponding to upward movement while the tile above the agent is water, the environment would block the move and return a no-op transition, but the world model instead renders the agent moving up with the water tile replaced by sand. The rollout is visually coherent and locally consistent with the action, yet it does not correspond to any transition the environment can produce. This matters for planning, where a plan optimized against such rollouts may be unexecutable. Constraints of this kind are hard to recover from passive video: a blocked command produces no visual change, so the LAM encodes it as a near-idle latent rather than as evidence that the action was refused. Learning which actions an environment rejects would likely require interaction with it rather than observation alone.

The world model also predicts from the most recent frame and latent action with no explicit memory. In partially observable settings this costs accuracy and makes long rollouts drift, since anything that left the frame is gone.

Finally, latent actions are not executable on their own. Grounding them to a particular embodiment still takes a small amount of action-labeled data, so \alg reduces the annotation burden rather than removing it.